 \documentclass[final,3p,times,twocolumn]{elsarticle}

\usepackage{graphicx}

\usepackage{amssymb}
\usepackage{bm}
\usepackage{amsthm}
\usepackage{mathtools}
\usepackage{multirow}
\usepackage{caption}
\usepackage{subcaption}

\begin{document}

\begin{frontmatter}



\title{A Resilient Image Matching Method with an Affine Invariant Feature Detector and Descriptor}


\author{Biao Zhao, Shigang Yue}

\address{University of Lincoln}

\begin{abstract}

Image feature matching is to seek, localize and identify the similarities across the images. The matched local features between different images can indicate the similarities of their content. Resilience of image feature matching to large view point changes is challenging for a lot of applications such as 3D object reconstruction, object recognition and navigation, etc, which need accurate and robust feature matching from quite different view points. In this paper we propose a novel image feature matching algorithm, integrating our previous proposed Affine Invariant Feature Detector (AIFD) and new proposed Affine Invariant Feature Descriptor (AIFDd). Both stages of this new proposed algorithm can provide sufficient resilience to view point changes. With systematic experiments, we can prove that the proposed method of feature detector and descriptor outperforms other state-of-the-art feature matching algorithms especially on view points robustness. It also performs well under other conditions such as the change of illumination, rotation and compression, etc. 
\end{abstract}

\begin{keyword}
Feature detector, Feature descriptor, Image feature matching, Histogram 
\end{keyword}

\end{frontmatter}

\section{Introduction}

Feature detection and matching is critical to applications such as wide baseline matching, object recognition, texture recognition, image retrieval, robot localization, panorama creation, etc \cite{mikolajczyk2005performance}. It is based on content analysis rather than the manually tags \cite{gudivada1995content}, and is also the most promising method to indicate the similarities between different images. These matched features can either be used to identify the content of the images or to figure out the relative geometry positions of the snapshots. A good matching algorithm is not only capable of extracting as many as possible matched correspondence but also sufficiently stable to the fluctuation of the scenes, such as illumination, exposure, partial occlusion, scale translations, rotation and affine transformations \cite{mikolajczyk2005performance} \cite{mikolajczyk2004scale}. The resilience to affine transformations (e.g. different view points), which can greatly widen the scope of the utilization, becomes quite desirable based on the practical applications \cite{zhao2017affine}, since the source of visual content can be quite different. It is natural to take a snapshot without careful calibrations and settings, if it is just to be shared with friends or uploaded to social media. Without sufficient scene stabilities, the range to apply the feature matching technique can become quite limited. The object of a feature matching algorithm is to be applicable to the images of all categories, rather than a certain type and some applications, such as self-driving, 3D reconstruction, etc are quite sensitive to scene fluctuations, especially to the affine transformations.

Some matching algorithms have shown their success on the resilience to some scene fluctuations, including Scale Invariant Feature Transform (SIFT) \cite{lowe1999object} \cite{lowe2004distinctive}, Harris-Affine detector \cite{gueguen2011multi}, Hessian-Affine detector \cite{mikolajczyk2004scale}, Speeded-Up Robust Features (SURF) \cite{bay2006surf} \cite{bay2008speeded}, A Low-dimensional Polynomial detector (ALP) \cite{TM12cdvs} \cite{cdvs_white}, maximally stable extremal region (MSER) \cite{matas2004robust}, gradient location and orientation histogram (GLOH) \cite{mikolajczyk2005performance}, shape context \cite{mikolajczyk2003shape} and steerable filters \cite{zhao2015affine} on the respect(s) of feature detector, or feature descriptor. All of these algorithms are translation invariant. Harris-Laplace detector, determinant of the Hessian operator, and the DoG (SIFT detector), GLOH, SURF are also invariant to scale and rotation. Meanwhile, the moment-based blob detectors like Harris-Affine and Hessian-Affine, and MSER are partially resilient to affine transformations. In particular, MSER \cite{matas2004robust} has a better performance than others in terms of affine invariance. However, it is not robust to scale and vulnerable to the drastic changes of the level line geometry. SIFT is a quite powerful scale invariant feature detector and descriptor, but its performance on the affine resilience is not satisfying \cite{lowe2004distinctive}. SURF can be deemed as a speeded up image matching implementation based on the wavelet. It has a similar performance with SIFT with much less computational complexity \cite{bay2008speeded}. It has also achieved it success on image rotation, scale invariance and it is also less powerful on affine resilience. GLOH is a new descriptor which extends SIFT by changing the location grid and using PCA to reduce the descriptor size. It has a better performance then SIFT on matching precision but it is also more vulnerable to the scene fluctuations \cite{mikolajczyk2005performance}. Among all these brilliant designs, however, none of them performs adequately well at affine fluctuations.  

Affine invariance is to describe the image matching algorithm which is capable of matching images from any perspectives. It is critical to a lot of view point sensitive applications of computer vision and becomes quite appealing especially when all the other scene fluctuations have been well solved. 

A typical image feature matching pipeline can be simplified as 3 steps: feature detector, feature descriptor and feature matching \cite{cdvs_white}. All of these steps account equally for its performance on affine invariance. A good affine invariant feature matching algorithm should perform well on all of these steps.

With regard to the feature detector, we have proposed Affine Invariant Feature Detector (AIFD) in our previous work, for dealing with the fluctuations especially the affine transformations \cite{zhao2017affine} within the framework of affine scale space. This algorithm provides a more general accessible structure, allowing the features to be detected under variety of affine transformations. To approach a fully affine invariance, the local extrema detection, has been reshaped by comparing the local Harris and Hessian matrices to adapt for the introduced affine scale space \cite{zhao2017affine}. Additionally, we have applied the polynomial expression to approximate the affine scale space and utilize the pyramid structure to speed up the generalization processing. It outperforms the most general applied feature detectors especially for the cases with large different view points.

As an important step of image feature matching, feature descriptor also plays an important role in the resilience to affine fluctuations. An adequately affine robust feature descriptor can guarantee that the features detected from different view points can be well identified and matched. Following the idea of AIFD, we propose the Affine Invariant Feature Descriptor (AIFDd), with the purpose to provide an affine invariant image feature descriptor, compatible and integrable with our previous proposed AIFD. Feature descriptor, as it is defined, is a set of data used to tag the local extracted features. A local feature descriptor generally is formed by a set of data regarding the local image properties, such as gradient, colours, textures, etc and can be measured by a similarity score between different descriptors. A pair of descriptors with a similarity score above the threshold can be selected as the matched features. Our proposed descriptor AIFDd introduces the affine scale space and polynomial expressions to approach the local gradient of the specified scale as the local image property. After reshaping the local patches of gradient, the effect of affine transformations can be eliminated. By steering the patches to the main directions, subdividing them to small cells and accumulating the gradient from each cell, the histograms filled with the local image gradient can be created. This histogram can be well used to tag the features from different perspectives. After quantizing the accumulated gradient values, the histogram can be formed as our proposed local feature descriptor AIFDd. AIFDd has a similar structure as SIFT descriptor, and can be used to replace SIFT in some cases of a higher affine robustness requirement. 

With systematic experiments, we can prove that our proposed resilient image matching method has a better performance than other image matching algorithms especially for the cases with a large view point angles. 

\section{Affine invariant feature detector}

This affine invariant image matching method is based on our previous proposed feature detector, a more strong and stable feature detector to tackle with the images with extremely tilted transformations. It has achieved its success by introducing the affine scale space and adapting the local extrema detection to suit the affine transformed environment. By analysing the results, the feature detector has an overwhelming out-performance against the widely applied feature detectors, including SIFT, SURF and ALP \cite{zhao2017affine} \cite{6571989}. In this section, we presents a brief introduction of AIFD.

\subsection{Affine scale space and its implementation}

Scale space is the fundamental of scale invariant feature detector. It provides a multi-scale image representation, allowing the content to be analysed from different scales. However, as has been pointed out  \cite{zhao2015affine}~\cite{zhao2017affine}, Gaussian convolution, the basis to generate the scale space, is not linear deformable. That is the reason why the scale invariant feature detectors are sensitive to view point changes. Our previous proposed Affine scale space~\cite{zhao2017affine}, which provides an approach to linear deformable multi-scale image representation, is more applicable to affine transformations. In practice, some feasible implementations to generate this structure~\cite{zhao2017affine} are also available. 

For a given image $I(x,y)$, its scale space is given by a family of pre-smoothed images $L(x,y,\sigma)$, where the scale parameter is pre-defined according to its kernel size:

\begin{equation}
\begin{split}
g(x,y;\sigma)&=\frac{1}{2\pi\sigma^{2}}e^{-\frac{x^{2}+y^{2}}{2\sigma^{2}}} \\
\text{such as,} \quad L(x,y;\sigma)&=g(x,y;\sigma)*I(x,y)
\end{split}
\end{equation}

\begin{figure}[!h]
\begin{minipage}{0.49\linewidth}
\centering
\includegraphics[width=0.80\linewidth]{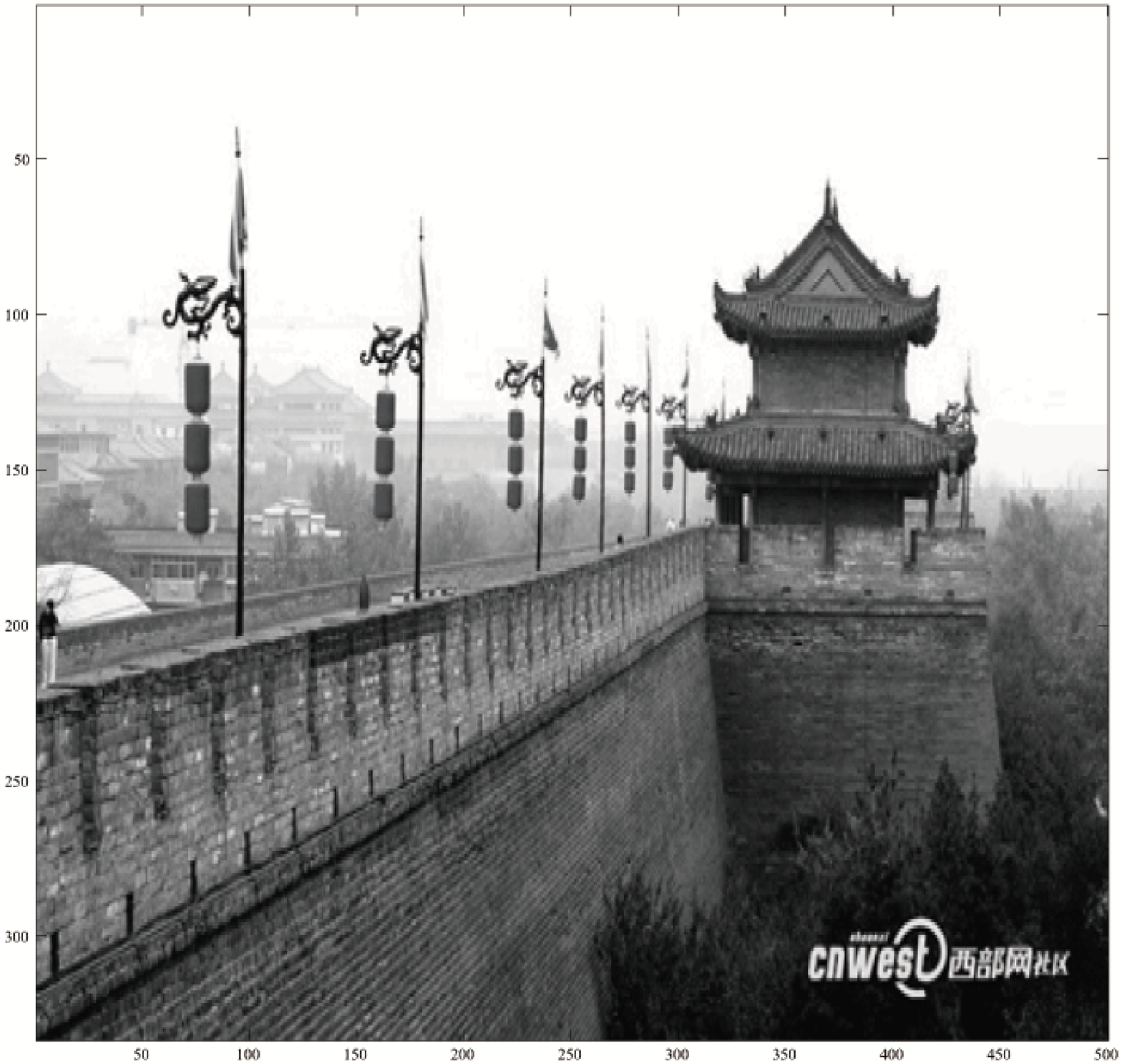}

\small\centerline{(a) $\sigma=0$, original image.}\medskip
\end{minipage}
\begin{minipage}{0.49\linewidth}
\centering
\includegraphics[width=0.80\linewidth]{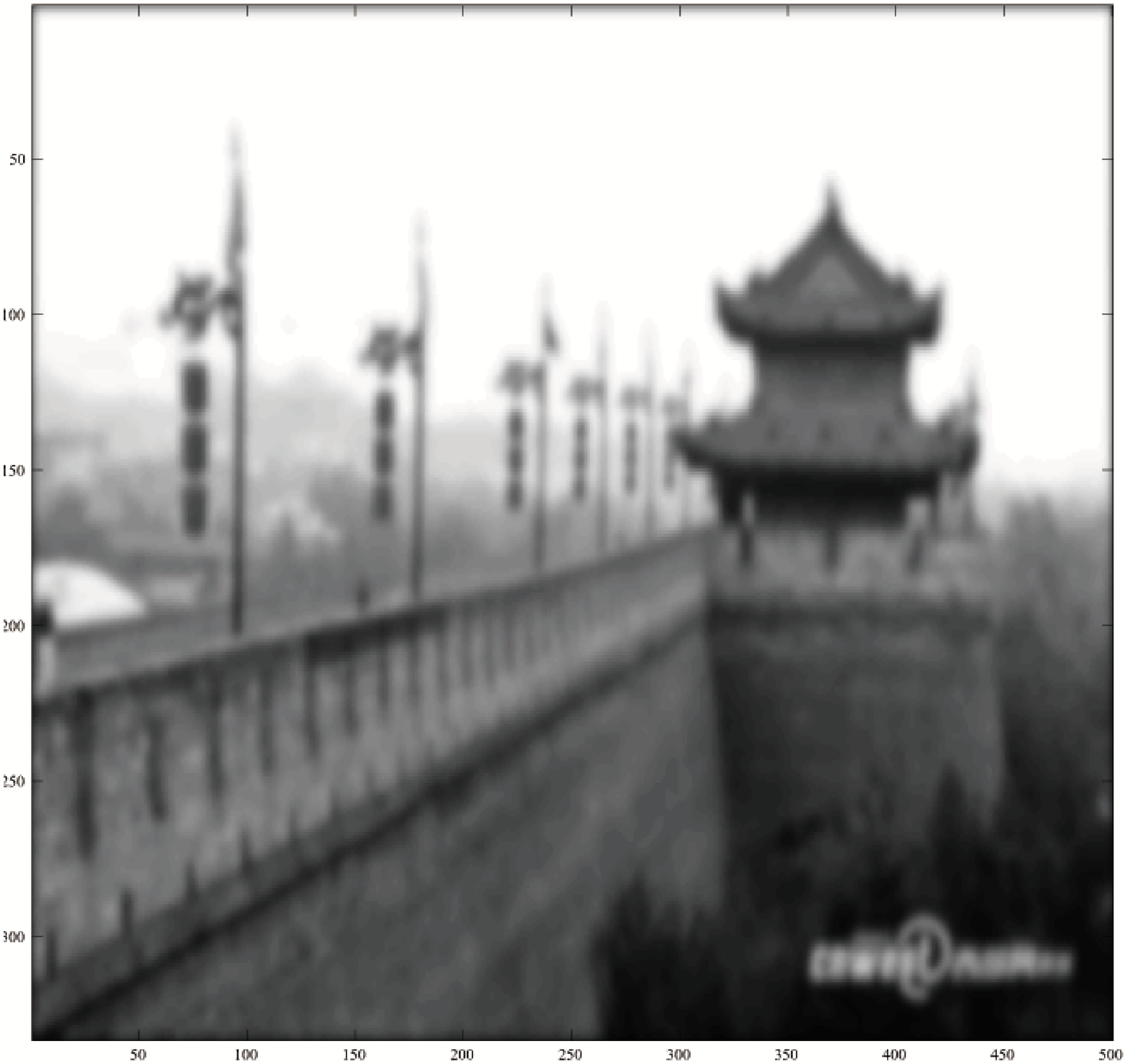}

\small\centerline{(b) $\sigma=1.6$.}\medskip
\end{minipage}
\begin{minipage}{0.49\linewidth}
\centering
\includegraphics[width=0.80\linewidth]{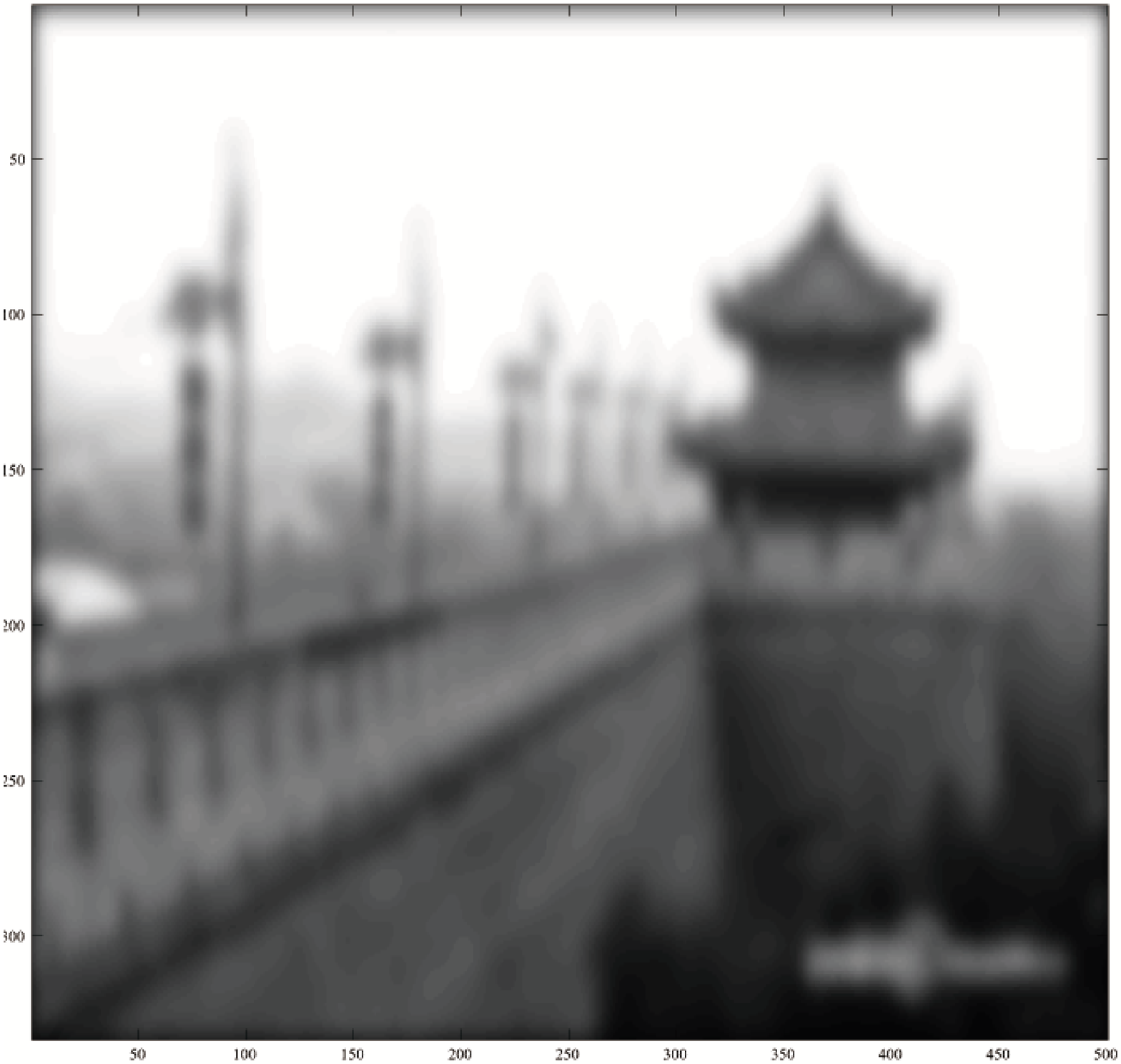}

\small\centerline{(c) $\sigma=1.6 \times 2$. }\medskip
\end{minipage}
\begin{minipage}{0.49\linewidth}
\centering
\includegraphics[width=0.80\linewidth]{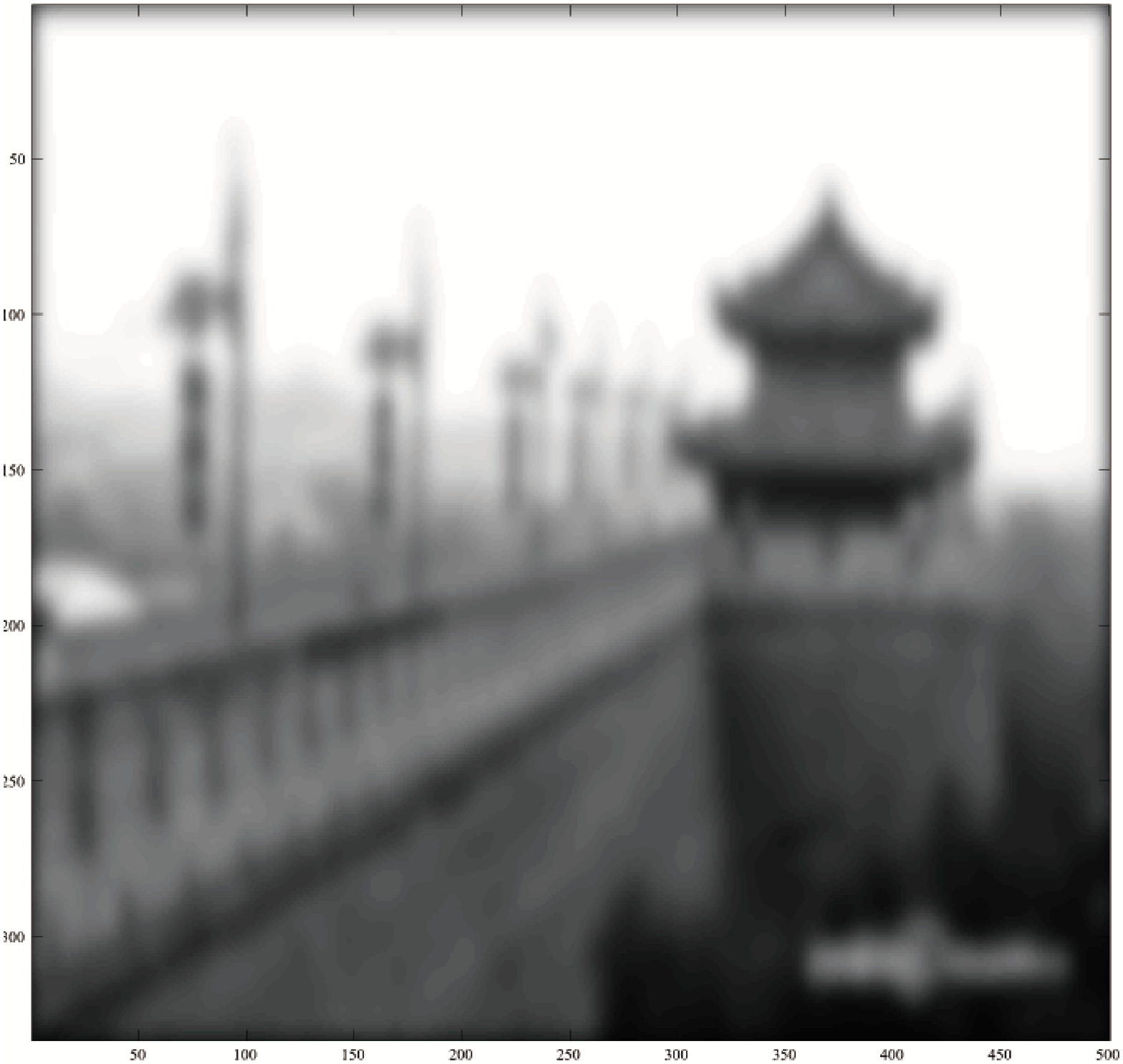}

\small\centerline{(d) $\sigma=1.6 \times 2\sqrt{2}$. }\medskip
\end{minipage}
\caption{A typical scale space, containing the images pre-smoothed by different kernel size of Gaussian filters. When the Gaussian scale equals to $0$, the pre-smoothed image is equivalent to the original image.}
\label{fig:scal_spa}
\end{figure}
 
Based on this definition, a structure more adaptable to the affine transformation is defined below:

\begin{equation}
\begin{split}
g(\eta,\Sigma_{s})_{A}=&\tfrac{1}{2 \pi \sqrt{det \Sigma_{s}}} e^{- \tfrac{\eta^{T}\Sigma_{s}^{-1}\eta}{2}}. \\
\text{where,} \quad  \Sigma_{s}=&A\sigma^{2}A^{T}
\end{split}
\end{equation}

In this formula, $A$ represents the affine transformation, a  $2 \times 2$ matrix. $\sigma$ is the scale. This deformed Gaussian kernel is specialized to generate affine scale space which can maintain linear relationship regardless the change of view point. Based on this structure, the images from any view points can be well represented from multi-scales \cite{lindeberg2013generalized}. From the definition of affine scale space, conventional isotropic scale space can be deemed as a spacial case, whose affine transformation equals to the $2 \times 2$ identity matrix.

Besides a multi-scale image representation from a specified view point. Some more robust and representable points will then be selected as the feature candidates. The similarity of two visual content is largely depends on the matched features detected from the scale space.   
To the conventional scale space, several approaches to detect the local maximum or minimum from derivatives have been proposed \cite{lindeberg1992scale}, and local LoG extrema detection outperform all others, concerning the accuracy and efficiency of a method in practice \cite{tuytelaars2008local}. 

The Laplacian of Gaussian (LoG ) scale space can be mathematically expressed as:

\begin{equation}
\nabla^{2}L=L_{xx}+L_{yy} 
\end{equation} 
In this formula, $L$ represents the scale space. The local maximum or minimum over the Laplacian can then be selected as the feature candidates \cite{Lindeberg:2012}.  

Borrowing the idea of LoG, we have also proposed an affine LoG, with the purpose to promote the feature candidates detection over affine scale space \cite{zhao2017affine} \cite{zhao2015affine}. Different to the conventional LoG, affine LoG can not be directly generated from affine Gaussian scale space. Instead of a direct laplacian operation, we have proposed a feasible implementation based on our proposed pyramid structure to efficiently generate affine LoG. By this implementation, the affine Gaussian and LoG scale space can be simultaneously generated. More information about affine scale space and affine LoG can be found \cite{zhao2017affine} \cite{zhao2015affine}. With this framework of affine scale space \cite{zhao2017affine} \cite{zhao2015affine}, we have proposed the Affine Invariant Feature detector (AIFD), as described in the following sub section.

\subsection{Affine invariant feature detector}

Based on the affine scale space, we have proposed an affine invariant feature detector (AIFD) \cite{zhao2017affine}, which is more capable of dealing with the images of different view points accounting for the view points difference. The pipeline of this feature detector is demonstrated below.

\begin{figure*}[!htp]
\centering
\includegraphics[width=0.9\linewidth]{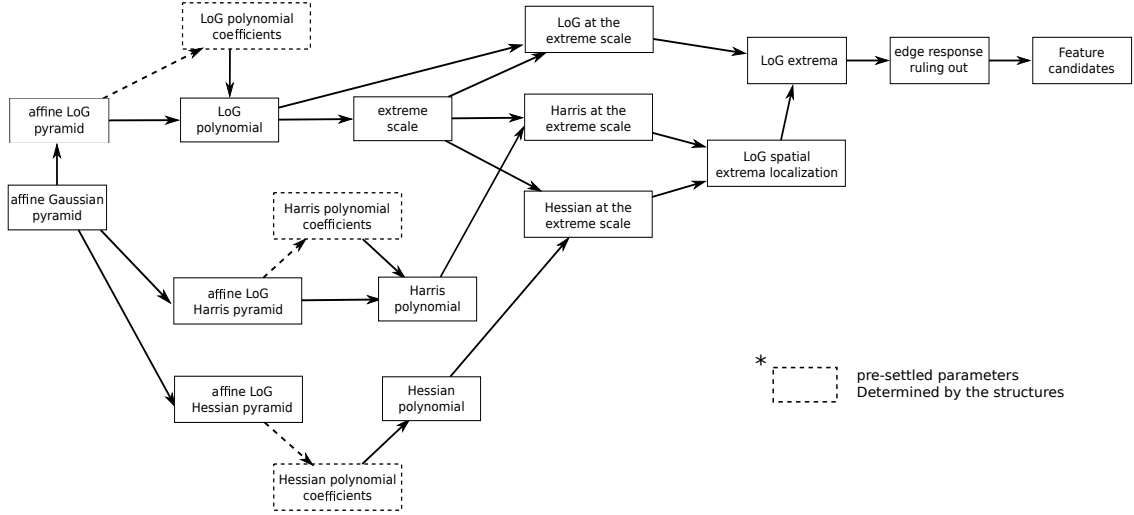}
\caption{The pipeline of AIFD. The parameters in the dashed box is pre-settled and only determined by the pyramid structure. }
\label{fig:pipe}
\end{figure*}    

As the Fig~\ref{fig:pipe} illustrated, the fundamental of AIFD are, affine Gaussian scale space, affine Laplacian of Gaussian (LoG) scale space, affine LoG Harris scale space and affine LoG Hessian scale space. Establishing these 4 scale spaces aided by the affine Gaussian pyramid structure, we can then seek a polynomial expression to maximumly approximate the affine LoG scale space as a continual function, presenting the scale space\rq{} change over the scale variable. Granted with this maths equation, we can then calculate the scale values at which the affine LoG scale space approach its maximum or minimum values. 
General speaking, a stable and robust feature candidates, in the environment of content-based visual search, are the local extremes from both the geometry dimension and the scale dimension. Thanks to the polynomial equation, we can then calculate the maximum or the minimum values that each pixel can approach by the change of scale and select it as the feature candidate if it is the maximum or the minimum point around its local geometry neighbourhoods. Different to the traditional isotropic feature detection method which finds the local geometry extremes by comparing the maximum or the minimum point to its 8 neighbourhood, we have proposed a different local geometry extreme detection, accounting for the affine invariance, by  comparing the eigenvalues of the local Harris and Hessian matrices.

If we denote the eigenvalues of Hessian matrix as $\psi_{1},\psi_{2}$ and the eigenvalues of Harris matrix as $\nu_{1},\nu_{2}$, the local maximum(minimum) point can be determined if,
\begin{equation}
\frac{1}{4}\min\{\psi_{1},\psi_{2}\}^{2}>\max\{\nu_{1},\nu_{2}\}
\label{eq:locextem}
\end{equation}
In advance, a maximum point can be defined if:
\begin{equation}
\max\{\psi_{1},\psi_{2}\}<0
\label{eq:locmax}
\end{equation}
a minimum point can be defined if:
\begin{equation}
\max\{\psi_{1},\psi_{2}\}>0
\label{eq:locmin}
\end{equation}

We have also formulated the LoG derivative filters to form the local Harris and Hessian matrices. The first can be presented in the form:
\begin{equation}
l_{1}=A^{-1}\eta\frac{1}{\pi\sigma^{6}}e^{-\frac{\eta^{T}(AA^{T})^{-1}\eta}{2\sigma^{2}}}\left(2-\frac{\eta^{T}(AA^{T})^{-1}\eta}{2\sigma^{2}}\right) 
\end{equation}
If we suppose the affine transformation is $A$, $(x,y)^{T}=\xi$, $\eta=A\xi$, then $\xi=A^{-1}\eta$. The first order of affine LoG derivatives can then be obtained to form the Harris matrix. In the same way, we can also formulate the second order of affine LoG derivatives in the form of:
\begin{equation}
\begin{split}
l^A_{xx}=&\frac{1}{\pi\sigma^{6}}\left[\left(\frac{\eta^{T}(AA^{T})\eta}{2\sigma^{2}}-3\right)(\frac{M_{a}(1,1)}{\sigma^{2}}-1)-1\right] \\
l^A_{yy}=&\frac{1}{\pi\sigma^{6}}\left[\left(\frac{\eta^{T}(AA^{T})\eta}{2\sigma^{2}}-3\right)(\frac{M_{a}(2,2)}{\sigma^{2}}-1)-1\right] \\
l^A_{xy}=&\frac{1}{\pi\sigma^{6}}\left(\frac{\eta^{T}(AA^{T})\eta}{2\sigma^{2}}-3\right)\frac{M_{a}(1,2)}{\sigma^{2}} 
\end{split}
\label{eq:sederivative}
\end{equation}
$M_{a}=A^{-1}\eta\eta^{T}(A^{-1})^{T}$, where the affine transformation is $A$, $(x,y)^{T}=\xi$, $\eta=A\xi$. In the Eq.\ref{eq:sederivative}, $M_{a}(i,j)$ represents the corresponding elements of $M_{a}$ ($i,j$). 
 
Thanks to the properties semi-group and sub-sampling, also satisfied by the affine LoG derivatives, we can then apply the LoG pyramid structure to generate the LoG derivative scale spaces to form the corresponding Harris and Hessian matrices of the specified scale and positions.   

Semi-group of affine LoG derivatives can then be expressed as,
\begin{equation}
D(l_{\Delta}(A^{-1}\eta;\sigma_{1}+\sigma_{2}) )= D(l_{\Delta}(A^{-1}\eta;\sigma_{1}))*g(A^{-1}\eta;\sigma_{2})
\end{equation}
 
\begin{figure}[!htp]
\centering
\includegraphics[width=0.9\linewidth]{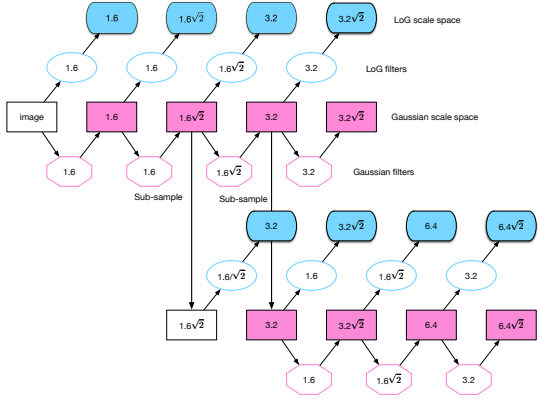}
\caption{Proposed pyramid structure to speed up the generation of Gaussian, LoG and also LoG derivatives scale space. The pink Octagons depict the Gaussian filters and the azure ellipses depict LoG or LoG derivatives filters. This structure can be divided into Gaussian part and LoG part, meaning the LoG/LoG derivatives can be later generated based on the same Gaussian scale space. }
\label{fig:pry}
\end{figure}    

The generation of affine LoG derivative operation can be divided into a smaller scaled LoG derivative operation convolving with a smaller scaled Gaussian blurring operation. An affine LoG derivative scale space can be generated by convolving a small scaled LoG derivative operations with the corresponding affine Gaussian pyramid. The established affine Gaussian scale space can then be re-utilized as the Gaussian blurred images part to speed up the generation of LoG and LoG derivative scale space.

Thanks to the pyramid structure, the affine LoG and the LoG derivatives can then be generated without a heavy computation. However, in the framework of the pyramid structure, each Octave is just constituted by 4 blurred images with different Gaussian scales. Specified to a certain feature candidate, hardly can it be of any of these 4 scales. Instead, in most cases, a feature can be of any scale value. Alternatively, the expressions of scale space in the scale domain shall be continual to present the whole scale space, rather than a few discrete scale samples. Thus, we have designed a polynomial expression, combined 4 discrete Gaussian blurred images with different assigned parameters, to formulate the whole scale space by a continual function. The polynomial expression can be formulated in such a way:
 \begin{equation}
f(\sigma)=\bar{a}(x,y)\cdot \sigma^3+\bar{b}(x,y)\cdot \sigma^2 + \bar{c}(x,y)\cdot \sigma +\bar{d}(x,y)
\label{eq:polynom}
\end{equation}
where, $\bar{a}(x,y)$, $\bar{b}(x,y)$, $\bar{c}(x,y)$ and $\bar{d}(x,y)$ are different combinations of 4 discrete blurred images. The parameters of these 4 combinations can be calculated in such a way:
 \begin{equation}
 M =\begin{bmatrix}\sigma^{3}&\sigma^{2}&\sigma&1 \end{bmatrix}^{-1} L_{t}(\sigma) \begin{bmatrix}L(\sigma_{1}) \\ L(\sigma_{2}) \\ L(\sigma_{3}) \\ L(\sigma_{4}) \end{bmatrix}^{-1}
\end{equation}
$M$ is the parameter defining the combination of $\bar{a}(x,y)$, $\bar{b}(x,y)$, $\bar{c}(x,y)$ and $\bar{d}(x,y)$. It is constant, regardless of the input images and affine transformations and can be pre-settled according to the implemented pyramid structures. Then, the polynomial expression can be obtained in such a way:
 \begin{equation}
 L_{t}(A,\sigma) =\begin{bmatrix}\sigma^{3}&\sigma^{2}&\sigma&1 \end{bmatrix} M \begin{bmatrix}L(A,\sigma_{1}) \\ L(A, \sigma_{2}) \\ L(A,\sigma_{3}) \\ L(A,\sigma_{4}) \end{bmatrix}
\end{equation}
$A$ is the corresponding affine transformation. The series polynomial expressions of LoG first and second derivatives can also be generated in the same way. We can denote them respectively as $L_{tx}(A,\sigma)$, $L_{ty}(A,\sigma)$, $L_{txx}(A,\sigma)$, $L_{txy}(A,\sigma)$ and $L_{tyy}(A,\sigma)$.

The scale value to approach the LoG\rq{}s maximum or the minimum can be calculated by taking:
 \begin{equation}
3\bar{a}(x,y)\cdot \sigma^2+2\bar{b}(x,y)\cdot \sigma + \bar{c}(x,y) = 0
\label{eq:derpoly}
\end{equation}
Since $\bar{a}(x,y)$, $\bar{b}(x,y)$,$\bar{c}(x,y)$ and $\bar{d}(x,y)$ are the images of Octave\rq{}s combinations, the result of $\sigma$ is of the same size with the input images. For each point, $2$ results can be obtained to satisfy the Eq.\ref{eq:derpoly}, meaning each LoG point will be assigned with two scale values, one approach to its maximum and another approach to its minimum. We can denote the scale approach to the maximum as $\sigma_{max}$ and the approach to the minimum as $\sigma_{min}$. If the $\sigma_{max}$ and $\sigma_{min}$ of one LoG point are within the scale range of the Octave, usually between $1.6$ and $3.2\sqrt{2}$, we can further form its Hessian and Harris matrices to check its geometry extreme situation. Calculating the values of  $L_{tx}(A,\sigma_{max})$, $L_{ty}(A,\sigma_{max})$, $L_{txx}(A,\sigma_{max})$, $L_{txy}(A,\sigma_{max})$, $L_{tyy}(A,\sigma_{max})$ and $L_{tx}(A,\sigma_{min})$, $L_{ty}(A,\sigma_{min})$, $L_{txx}(A,\sigma_{min})$, $L_{txy}(A,\sigma_{min})$, $L_{tyy}(A,\sigma_{min})$, forming the Harris and Hessian matrices and applying the method we proposed (indicated in Eq.\ref{eq:locextem}, Eq.\ref{eq:locmax}, Eq.\ref{eq:locmin}), we can then check if the point at its $\sigma_{max}$ is the local geometry maximum or at $\sigma_{min}$ is the local geometry minimum. Such points at such scales can be selected as the feature candidates.

Following the algorithm we proposed, the feature candidates will generally exert a very strong edge response, rendering the features more vulnerable to the change of the image properties, including the direction, blurring and titling. In order to rule out the features sensitive to the image changes, we will only preserve the features of less edge response, which can be measured by $R=Tr(H)^2/Det(H)$. In the equation, $H$ is the Hessian matrix, already generated to select the local geometry extreme points.

\section{Affine invariant feature descriptor}

\subsection{A more accurate feature location}
The feature detector we previously proposed can localize the detection up to pixel\rq{}s precision. However, for a more accurate feature descriptor, localizing the features to sub-pixel\rq{}s level becomes quite necessary \cite{brown2002invariant}. 

By multivariate Taylor theorem, a detected feature on LoG can be expressed in the form:
\begin{equation}
\begin{aligned}
&L({\boldsymbol {x+\Delta x}})=\sum _{|\alpha |\leq k}{\frac {D^{\alpha }L({\boldsymbol {x}})}{\alpha !}}({\boldsymbol {\Delta x}})^{\alpha }+\sum _{|\alpha |=k}h_{\alpha }({\boldsymbol {x+\Delta x}})({\boldsymbol {\Delta x}})^{\alpha },\\
&{\mbox{and}}\quad \lim _{{\boldsymbol {\Delta x}}\to {\boldsymbol {0}}}h_{\alpha }({\boldsymbol {x}})=0.
\end{aligned}
\end{equation}
where ${\boldsymbol {x}}=(x,y)^{T}$ is the feature position to integral precision, $\Delta x$ is the offset up to a higher accuracy. We can approach this offset by a second order of Taylor expansion, giving:
\begin{equation}
L({\boldsymbol {x+\Delta x}}) = L(\boldsymbol {x})+\Delta \boldsymbol {x}^{T}\{DL(\boldsymbol {x})\}+ \frac{1}{2}\Delta \boldsymbol {x}^{T}\{D^{2}L(\boldsymbol {x})\}\Delta \boldsymbol {x}
\end{equation}
where $DL(\boldsymbol {x})$ is the gradient of $L$  and $D^{2}L(\boldsymbol {x})$ is its Hessian matrix.

Since $L(\boldsymbol{x+\Delta x})$ is the local extreme, its derivative equals $\boldsymbol{0}$. By taking the derivative on both sides of the equation, we can have:
\begin{equation}
\Delta \boldsymbol{x}=-\{D^{2}L(\boldsymbol{x})\}^{-1}DL(\boldsymbol{x})
\label{eq:offsetadd}
\end{equation} 
If the offset is larger than $0.5$ in any dimension, it implies the real extreme location is more closer to a different pixel sample. Considering the local extreme scale for each pixel sample can be quite diverse, the feature located at the specified integral position becomes no more adequate.

The offset $\hat{x}$̂ will be summed up with the detected integral position to approach the local extreme location to sub-pixel\rq{}s precision according to the formula Eq.\ref{eq:offsetadd}. The Hessian matrix and local gradient of the pixel sample can be obtained by the corresponding LoG derivative polynomial expressions.

\subsection{Affine image gradient}

To cope with affine gradient based feature descriptor, we will introduce the affine gradient filter and the related scale space in this section. 
At the very beginning, the definition of image gradient can be given by:
\begin{equation}
\nabla I=\left(  \frac{\partial I}{\partial x} , \frac{\partial I}{\partial y} \right)
\end{equation}
It is equivalent to the first order of image derivatives. The traditional method to calculate the image gradient is by taking the subtraction from two image neighbouring pixels in the form:
\begin{equation}
\nabla I=\frac{1}{2}\left(I(x-1,y)-I(x+1,y), I(x,y-1)-I(x,y+1)\right)
\end{equation}

For every point of image gradient, its direction and magnitude can be given by:
\begin{equation}
\begin{split}
m(x,y)= & \sqrt{\left(\frac{\partial I}{\partial x}\right)^2+ \left(\frac{\partial I}{\partial y}\right)^2}, \\
\theta= & \arctan \left(\frac{\cfrac{\partial I}{\partial x}}{\cfrac{\partial I}{\partial y}}\right).
\end{split}
\end{equation}
To identify the scale value associated with the detected features the descriptor can be generated based on the gradient of the Gaussian blurred image of the corresponding scale space, which can be formulated as:
\begin{equation}
\begin{split}
\nabla L=&\left(  \frac{\partial L}{\partial x} , \frac{\partial L}{\partial y} \right) \\ 
= &\left(  I*\frac{\partial g}{\partial x} , I*\frac{\partial g}{\partial y} \right) 
\end{split}
\end{equation}  
where $I$ is the original image, $*$ is convolution operation, $g$ is Gaussian filter. Thus the derivative of the Gaussian blurred image of the corresponding scale space can be equivalently obtained though filtering the image with the Gaussian derivative filters, which can be derived by: 
\begin{equation}
\begin{bmatrix} g_{x}\\ g_{y} \end{bmatrix}=\begin{bmatrix}\cfrac{\partial g}{\partial x}\\ \cfrac{\partial g}{\partial y} \end{bmatrix}=\begin{bmatrix} \cfrac{x}{2\pi \sigma^{4}}e^{-\frac{x^{2}+y^{2}}{2\sigma^{2}}} \\ \cfrac{y}{2\pi \sigma^{4}}e^{-\frac{x^{2}+y^{2}}{2\sigma^{2}}}\end{bmatrix}=\frac{1}{2\pi \sigma^{2}}e^{-\frac{x^{2}+y^{2}}{2\sigma^{2}}}\begin{bmatrix}\cfrac{x}{\sigma^{2}} \\ \cfrac{y}{\sigma^{2}}\end{bmatrix}
\end{equation}
Thus the Gaussian derivatives scale space can easily be obtained by multiplying the corresponding Gaussian scale space with $x/\sigma^2$ and $y/\sigma^2$.

Borrowing the idea of affine LoG, we can have the affine Gaussian derivative filters as:
\begin{equation}
\begin{split}
\begin{bmatrix}g^A_{x} \\ g^A_{y}\end{bmatrix}=g^{A}_{\eta}=& \frac{A^{-1}\eta}{2\pi \sigma^{4}}e^{-\frac{\eta^{T}(AA^{T})^{-1}\eta}{2\sigma^{2}}}
\end{split}
\end{equation}
where $A$ is a $2\times2$ matrix, indicating the affine transformation, $\eta=A\begin{bmatrix}x \\ y \end{bmatrix}$. Similar to the isotropic Gaussian derivative scale space, the affine Gaussian derivative scale space can also be generated by multiplying the corresponding Gaussian scale space with $A^{-1}\eta/\sigma^2$.

In the same way of the affine Gaussian scale space, the polynomial expressions of the affine Gaussian gradient can also be generated with assigned parameters of the affine Gaussian scale space in the form as indicated in Eq.\ref{eq:polynom}.

\subsection{Affine gradient relocation}

Thanks to the efforts made in the previous section, the Gaussian gradient can be guaranteed to be constant regardless of the affine transformations except the geometry difference. To compensate the view point difference with the snapshot and the geometry difference brought by the affine Gaussian derivatives, the gradient values can be relocated according to the affine transformations. Then the relocated gradient value can then be utilized to compensate the gradient localization distortions. After relocating the required image gradient can it then be used to generate the approximation of the feature descriptors taking into account the view point difference.

Gradients from affine transformed images are restrained by the affine matrices between different view points. Around each features, the relocated gradients, according to the affine transformation, can then form a histogram as the feature descriptor. 

\begin{figure}
\centering
\includegraphics[width=0.99\linewidth]{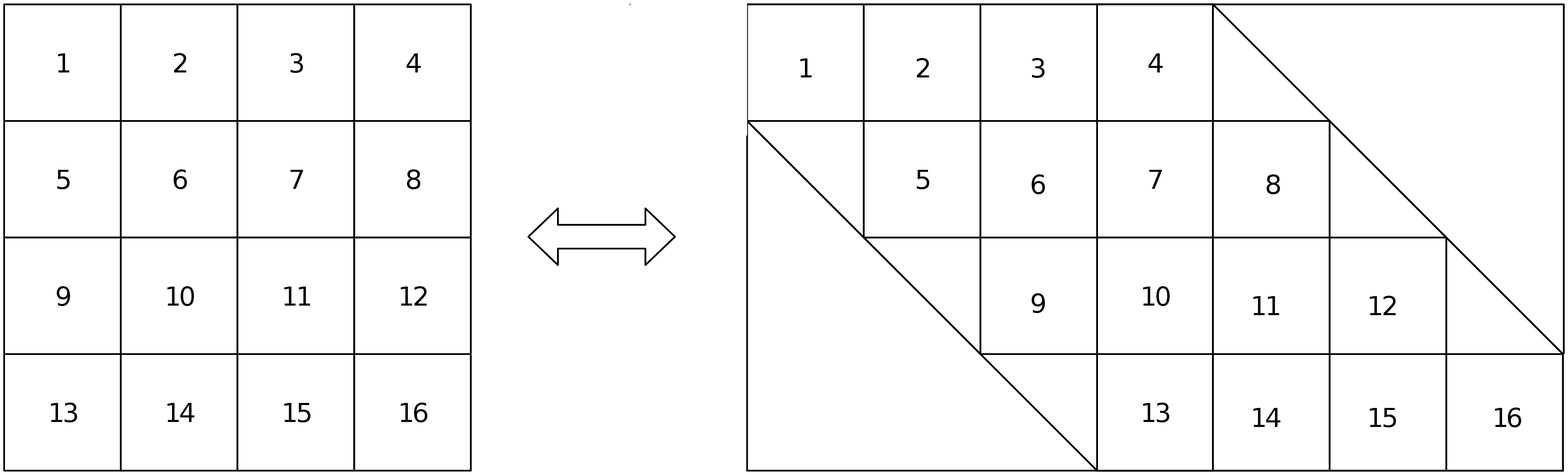}
\caption{Gradient relocated from the specified area around the detected features. The index of the gradient can be calculated according to the affine transformations.}
\label{fig:gradcollect}
\end{figure}

The gradient relocation can be done in the form of:
\begin{equation}
\begin{bmatrix} x^{\prime} \\ y^{\prime}\end{bmatrix}=A\begin{bmatrix} x \\ y  \end{bmatrix}
\end{equation}
where $x,y$ are the index of the gradient around the detected features from $1$ to the descriptor , pre-defined by the scale of the features. $x^{\prime}$ and $y^{\prime}$ are the index accounting for the affine transformation. The relocation of the gradient can then be collected according to the new calculated index. The interpolation of the gradients may also be applied if the new calculated index are not integers.

\subsection{Orientation assignment}
Assigning an orientation to each feature, the feature descriptor can be represented relative to this orientation and achieve its invariance to image rotation. To calibrate the orientation of a feature, an area of scale space gradient around the feature will first be formed after proceeding our proposed gradient relocation eliminating the effect of the affine distortion. The area of gradient to be collect is in a square shape with its size equal to $3$ time of the feature scale. Then, the orientation of each sample of scale space gradient can be added to the orientation histogram weighted by its gradient magnitude and by a Gaussian-weighted circular window with $1.5$ times of the scale \cite{lowe2004distinctive}.

Then the orientation histogram will be subdivided into $36$ bins covering the $360^{\circ}$ range of orientations and filled the corresponding accumulated magnitude<. The peak of the histogram points to its main direction. Any other local peak that is within $80\%$ of the highest peak and higher than the average of its two neighbours will be assigned with different orientations. The features with multiple peaks, will be respectively created at the same location with same scale but of different orientations \cite{lowe2004distinctive}.
\begin{figure}
\centering
\includegraphics[width=0.75\linewidth]{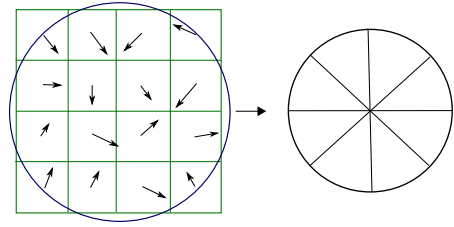}
\caption{The gradient around the feature will be added to the gradient histogram weighted by its magnitude and a Gaussian-weighted circular window.}
\label{fig:siftdir}
\end{figure}

\begin{figure}
\centering
\includegraphics[width=0.99\linewidth]{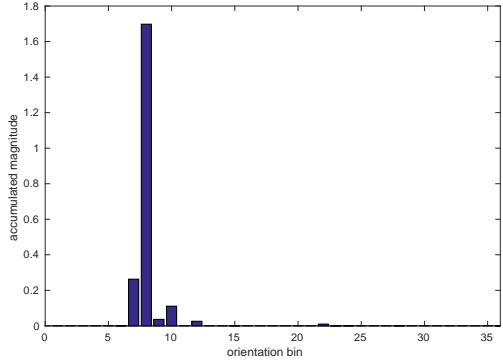}
\caption{Orientation histogram, created by subdividing the surrounding gradient into $36$ bins according to the gradient orientations and accumulating weighted magnitude. The largest bin will be selected as the main direction of the descriptor.}
\label{fig:gradcollect}
\end{figure}

Borrowing the idea of achieving the location accuracy, the main directions of the feature can also promote its accuracy by utilizing the second order of Taylor expansion in the form of:
\begin{equation}
\Delta \boldsymbol{x}=-\{D^{2}L(\boldsymbol{x})\}^{-1}DL(\boldsymbol{x})
\label{eq:offsetadd}
\end{equation} 
Supposing the detected main direction bin is $M$ and its two neighbour direction bins are $M_{-}$ and $M_{+}$, the above equation can be implemented in the form of:
\begin{equation}
\Delta M = - \cfrac{\frac{1}{2}(M_{+}-M_{-})}{\frac{1}{2}(M_{+}+M_{-})-M}
\end{equation} 
Then, the main direction is:
\begin{equation}
\theta=(M+\Delta M) \cdot \frac{360}{36}
\end{equation}
The offset larger than $0.5$ will be autocratically rejected.

\subsection{Affine descriptor}

\begin{figure}[!h]
\centering
\includegraphics[width=0.85\linewidth]{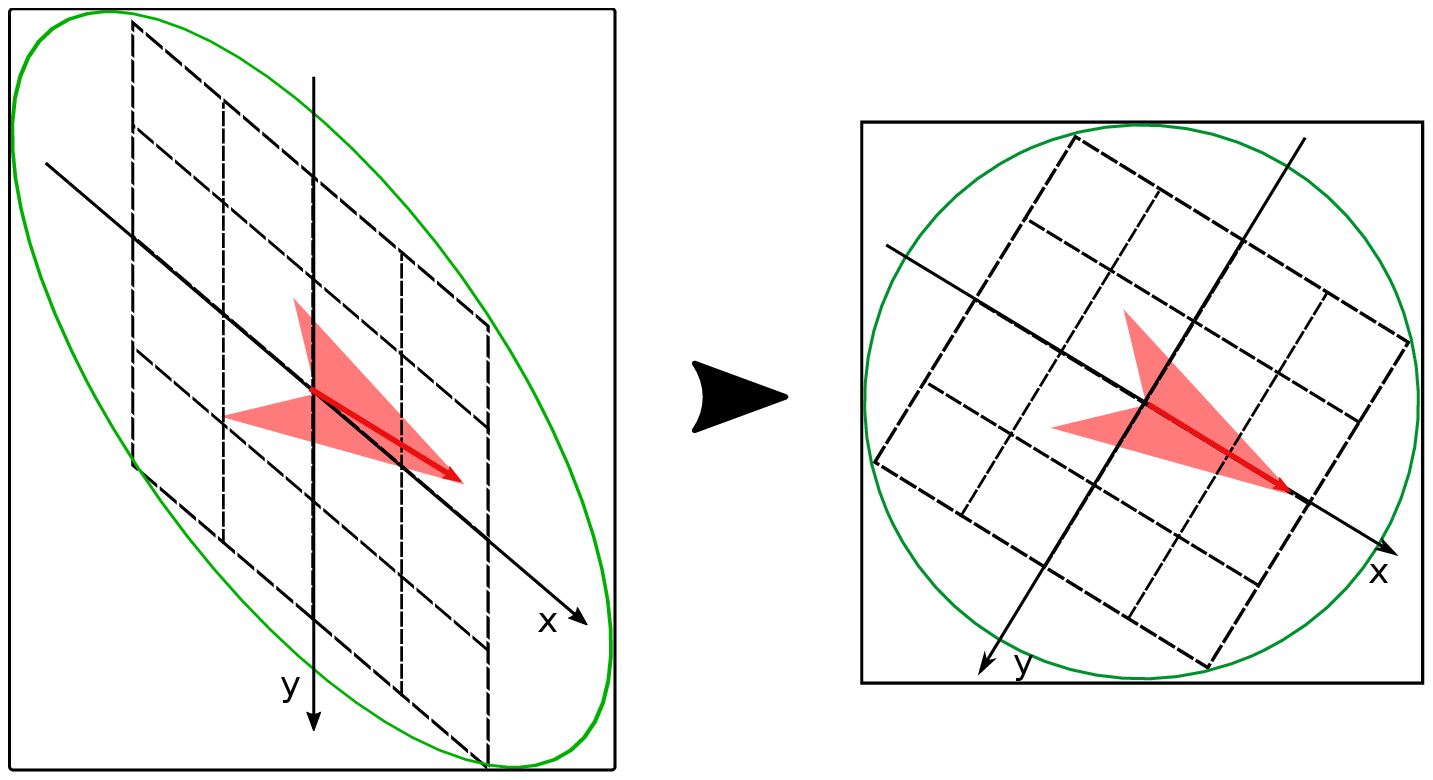}
\caption{Gradient relocation according to the synthesized transformation matrix. After the relocation, we will obtain a set of gradients eliminating the affine transformation and pointing to the main direction.}
\label{fig：aff_des}
\end{figure}

Assigning an orientation to each feature descriptor is to gain the invariance to image rotation by clarifying the relations between the represents and the orientation. Gradient is introduced with the purpose to avoid the effects of  illumination changes. The patch to collect the gradient will then steer to the main direction to generate the traditional SIFT descriptor \cite{lowe2004distinctive} \cite{lowe1999object}. The image rotation can also be equalized as a special case of affine transformation in the form of:
\begin{equation}
R(\theta )={\begin{bmatrix}\cos \theta &-\sin \theta \\\sin \theta &\cos \theta \\\end{bmatrix}}
\end{equation}
Combining the affine transformation, the total transformation can be synthesized as:
\begin{equation}
A^{\prime}={\begin{bmatrix}\cos \theta &-\sin \theta \\\sin \theta &\cos \theta \\\end{bmatrix}}\cdot A
\end{equation}
By applying the gradient relocation depicted in Fig.\ref{fig:gradcollect}, a square patch of gradient can we obtained.

\begin{figure}[!h]
\centering
\includegraphics[width=0.85\linewidth]{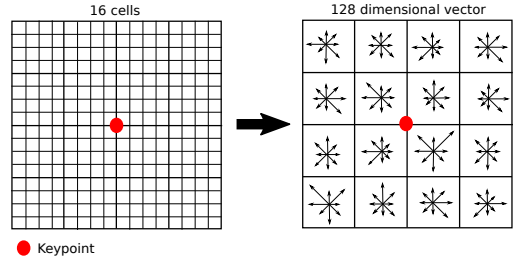}
\caption{Borrowing the idea of SIFT descriptor, affine descriptor can be generated from the relocated gradient.}
\label{fig:siftdes}
\end{figure}

As depicted in Fig \ref{fig:siftdes}, a SIFT like affine descriptor can then be generated. By dividing the affine-reformed gradient patch into $4 \times 4$ cells, we can then form a histogram, accumulating the gradient within each cell. Different to what it illustrated in Fig \ref{fig:gradcollect}, the accumulated gradient within each cell can be divided into $8$ degrees and the accumulating value will also be weighted by its magnitude. After normalizing such degrees, we can then form a $4 \times 4 \times 8 = 128$ elements histogram as the affine descriptor. This descriptor has a similar structure as SIFT, and can be utilized in some SIFT based applications for a better performance on affine invariance \cite{lowe1999object}. 

\section{Geometrical consistency check}

The feature matching at extreme view points may incur some incorrect matches. Utilization of affine descriptor can promote the correct matchings, but cannot prevent incorrect matches especially at extreme view points. Random sample consensus (RANSAC) \cite{huber2011robust} is available to rule out some potential outliers but requiring a higher computational cost. To provide a more precise and efficient image matching scheme, we will apply DISTRAT \cite{lepsoy2011statistical} (distance ratio coherence) to guarantee a coherent and consistent geometrical pairwise matching.

Supposing a set of pairwise matches are 
\begin{equation}
(\boldsymbol{X}_{1},\boldsymbol{Y}_{1}),(\boldsymbol{X}_{2},\boldsymbol{Y}_{2}) \ldots (\boldsymbol{X}_{N},\boldsymbol{Y}_{N})
\end{equation}
where $\boldsymbol{X}_{N}$ contains the coordinates of feature $N$ from the first image and $\boldsymbol{Y}_{N}$ contains the coordinates of the pairwise matched feature from the second image. The pairwise, meeting the geometrical consistency can be deemed as inlier and other pairwise can be termed as outliers.  

The logarithmic distance ratios (LDR) of these matched pairs are:
\begin{equation}
ldr(\boldsymbol{X_{i}},\boldsymbol{X_{j}},\boldsymbol{Y_{i}},\boldsymbol{Y_{j}}) = \ln \left(\frac{\|\boldsymbol{X_{i}}-\boldsymbol{X_{j}}\|}{\|\boldsymbol{Y_{i}}-\boldsymbol{Y_{j}}\|}\right), i \neq j
\end{equation}
It is quite obvious that the histogram for pairs of inliers is usually narrower than the histogram for pairs of outliers, which is to say that it is zero. Suppose that two distinct points in one image are never extremely close to each other, at least not in only one of the images. Then there exist numbers $\alpha$ and $\beta$, neither very large nor very small, such that
\begin{equation}
\alpha\|\boldsymbol{X_{i}}-\boldsymbol{X_{j}}\| \leq \| \boldsymbol{Y_{i}}- \boldsymbol{Y_{j}}\| \leq \beta\|\boldsymbol{X_{i}}-\boldsymbol{X_{j}}\|
\end{equation}
for the sets of points in the two images. The LDR is then restricted to an interval
\begin{equation}
ldr(\boldsymbol{X_{i}},\boldsymbol{X_{j}},\boldsymbol{Y_{i}},\boldsymbol{Y_{j}}) \in [-\ln\beta, -\ln\alpha]
\end{equation}

Generally speaking, the features for an image are independent and identically distributed with a normal distribution of variance $1/2 \cdot I$. For the outliers, we can assume that the coordinates have been suitably scaled so that the features are distributed over the whole image. Then, the difference between two outliers also has a normal distribution:
\begin{equation}
\boldsymbol{X_{i}}-\boldsymbol{X_{j}} \sim \mathcal{N}(0,I);i \neq j.
\end{equation}
Then,
\begin{equation}
R^{2}_{ij}=S=\frac{\|\boldsymbol{X_{i}}-\boldsymbol{X_{j}}\|^2}{\|\boldsymbol{Y_{i}}-\boldsymbol{Y_{j}}\|^2} \sim \mathcal{F}(2,2)
\end{equation}
Since the distribution of the inliers and the outliers behaviours quite differently, we can apply the Pearson\rq{}s chi-square test to rule out the potential outliers. The method proceeds in two stages: one is a hypothesis test used to exclude rapidly most non-matching pairs of images, the second estimates the number of inliers – a number used for ranking and decision. 

The outlier behavior is expressed through a discrete probability density:
\begin{equation}
f(k),k=1,2\ldots K
\end{equation}
The observation of LDR behaviour can be studied by forming a histogram, accounting the occurrences over each bin in the form $h(k)$ with $K$ bins over the adjacent intervals subdividing the numbers between for example between $-2.5$ to $2.5$ \cite{lepsoy2011statistical}.

The number of inliers can be estimated by solving the eigenvalue problem.
\begin{equation}
\beta=\cfrac{\sum\limits_{k=1}^{K}h(k)f(k)}{\sum\limits_{k=1}^{K}(f(k))^{2}}
\end{equation}
With the $\beta$, the outlier normal of the histogram can be created in the form,
\begin{equation}
d(k)=h(k)-\beta f(k)
\end{equation}
This outlier normal is orthogonal to the outlier pdf $f$.
Let $q$ be the quantizer that assigns a bin to any LDR value, the $ldr$ of interval $k$ can be expressed as:
\begin{equation}
ldr \in \zeta_{k} \rightarrow \boldsymbol{z} qk.
\end{equation}
The inlier matrix $D$ can be established as:
\begin{equation}
D_{ij}=\left\{\begin{aligned}  d_{q}(z_{ij})   \qquad i \neq j \\   0  \qquad  i=j \end{aligned} \right.
\end{equation}
where $z$ is $ldr$, $d_{q}=d\circ q$.
The dominate eigenvector $r$ of $D$ with eigenvalue $\mu$ can be found in the form of:
\begin{equation}
Dr=\mu r
\end{equation}
The number of inliers can be estimated by:
\begin{equation}
\hat{m}=1+\cfrac{\mu}{\max\limits_{k=1,\ldots,K} d(K)}
\end{equation}
The inlier correspond to the $\hat{m}$ largest elements in the eigenvector $r$. This geometric check can be applied after the AIFD matching to further rule out the potential false matched pairs.
\section{Experiments and results}

In the above chapters, we have proposed a resilient image matching method with an affine invariant feature detector and descriptor. Based on a more general image representation structure, this feature detector and descriptor is more capable of representing the features at extreme view points and can be matched under variable geometry and photometric transformations. In this section, we present several systematic experiment results, evolving with our previous proposed feature detector, to have a better evaluation on its resilience to affine and some other geometry and photometric transformations.

\begin{figure*}[!t]
\begin{minipage}{0.49\linewidth}
\centering
\includegraphics[width=0.89\linewidth]{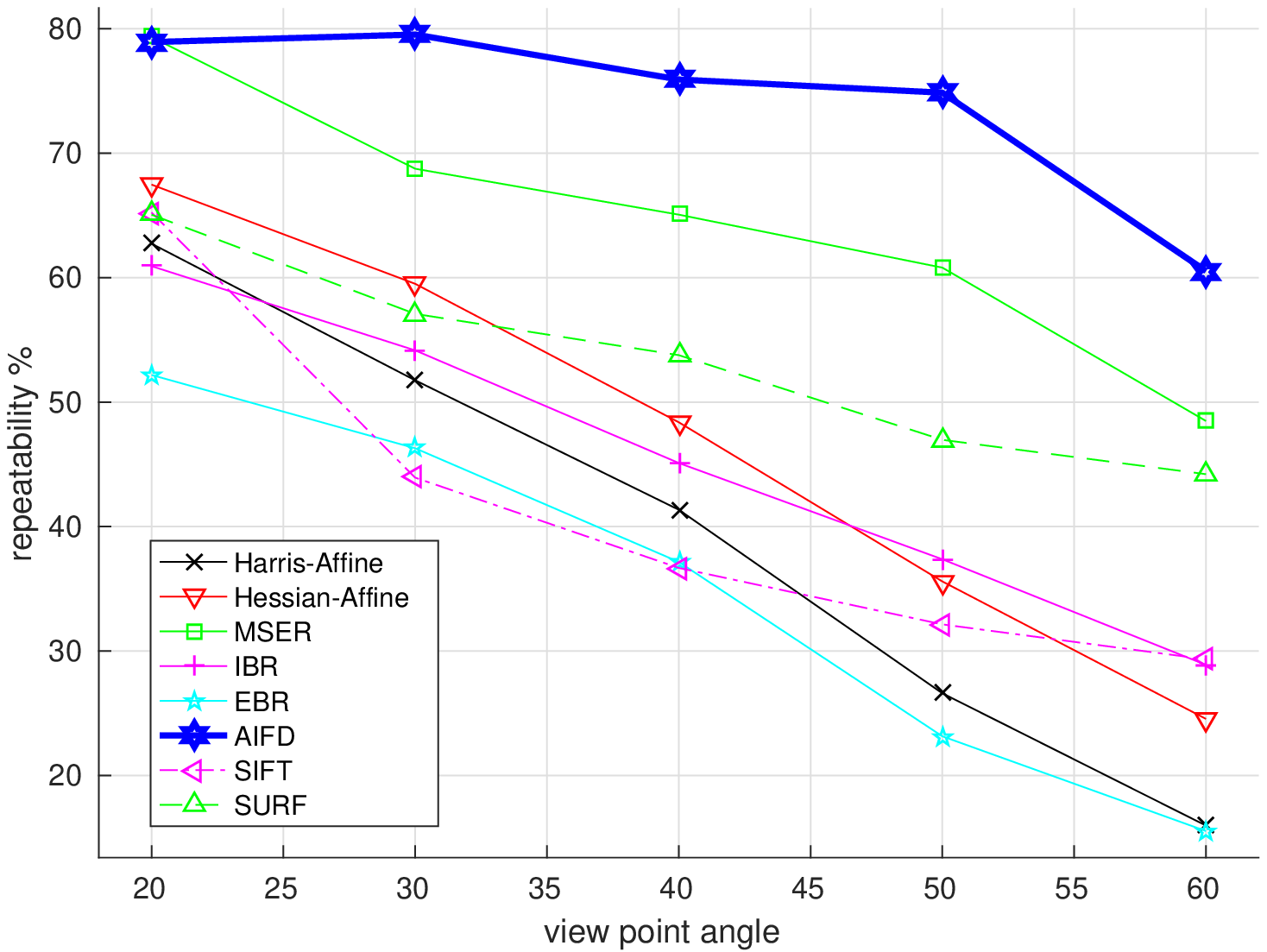}
\small\centerline{(a)}\medskip
\end{minipage}
\begin{minipage}{0.49\linewidth}
\centering
\includegraphics[width=0.89\linewidth]{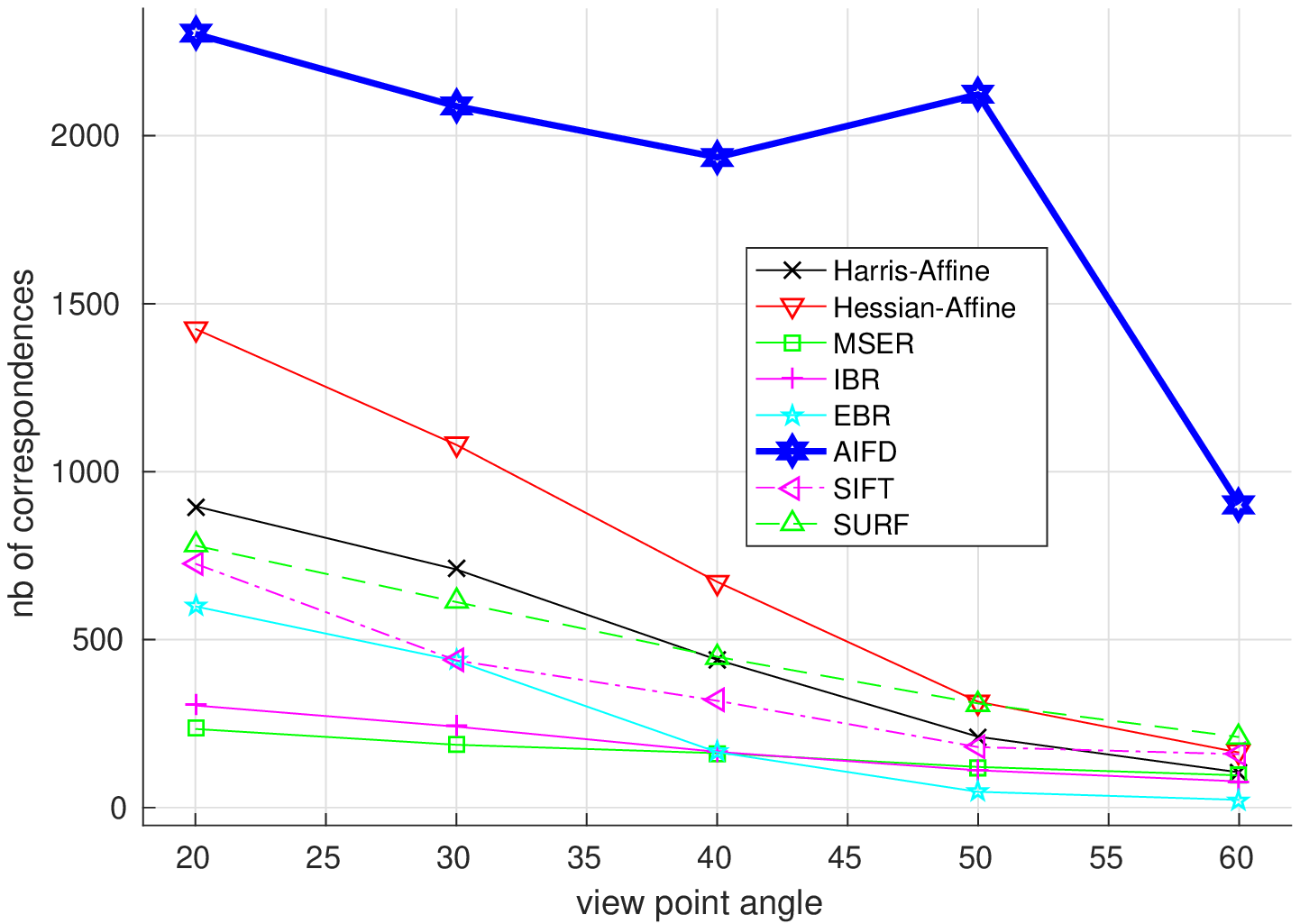}
\small\centerline{(b)}\medskip
\end{minipage}
\begin{minipage}{0.49\linewidth}
\centering
\includegraphics[width=0.89\linewidth]{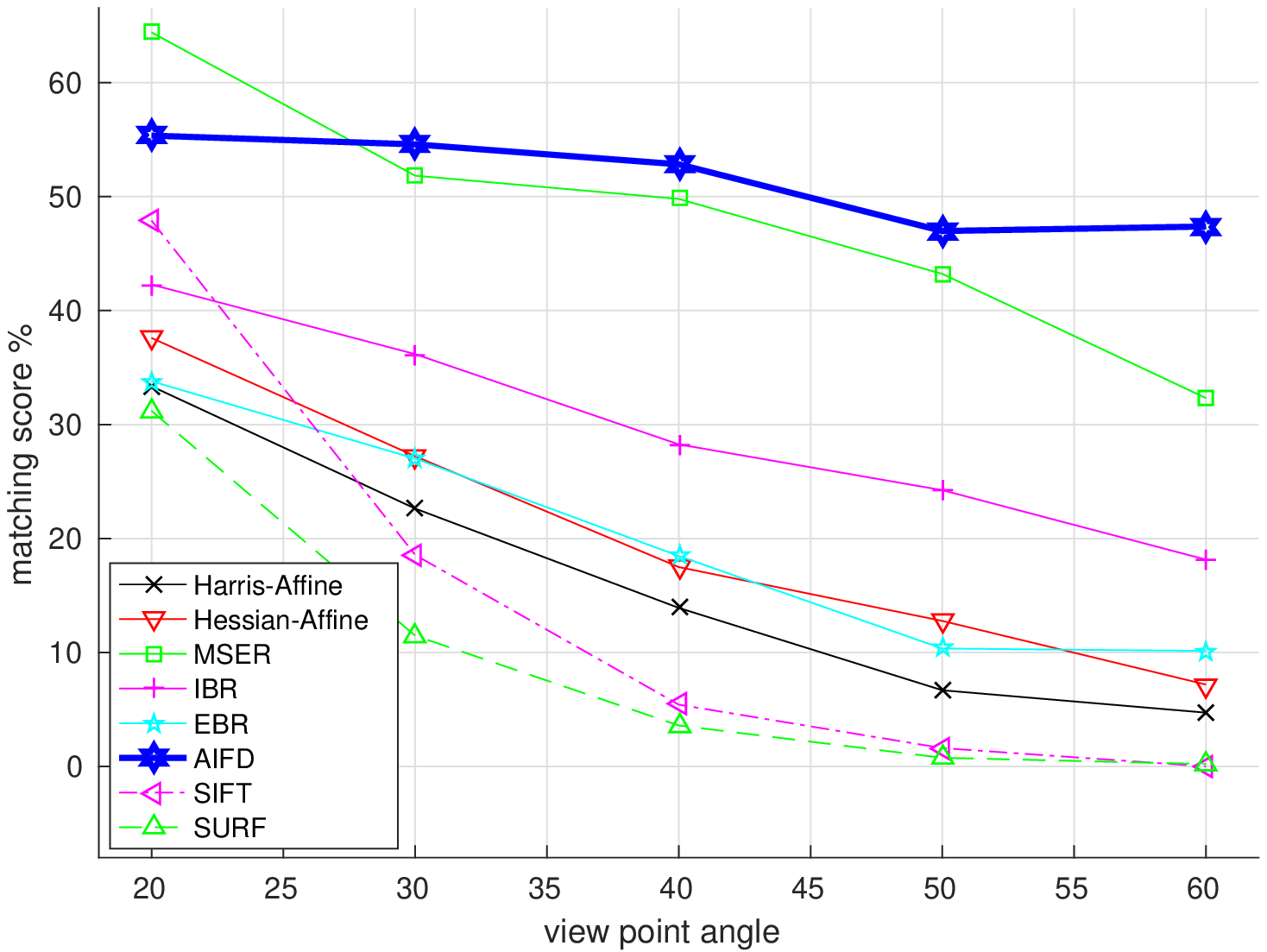}
\small\centerline{(c)}\medskip
\end{minipage}
\begin{minipage}{0.49\linewidth}\centering
\includegraphics[width=0.89\linewidth]{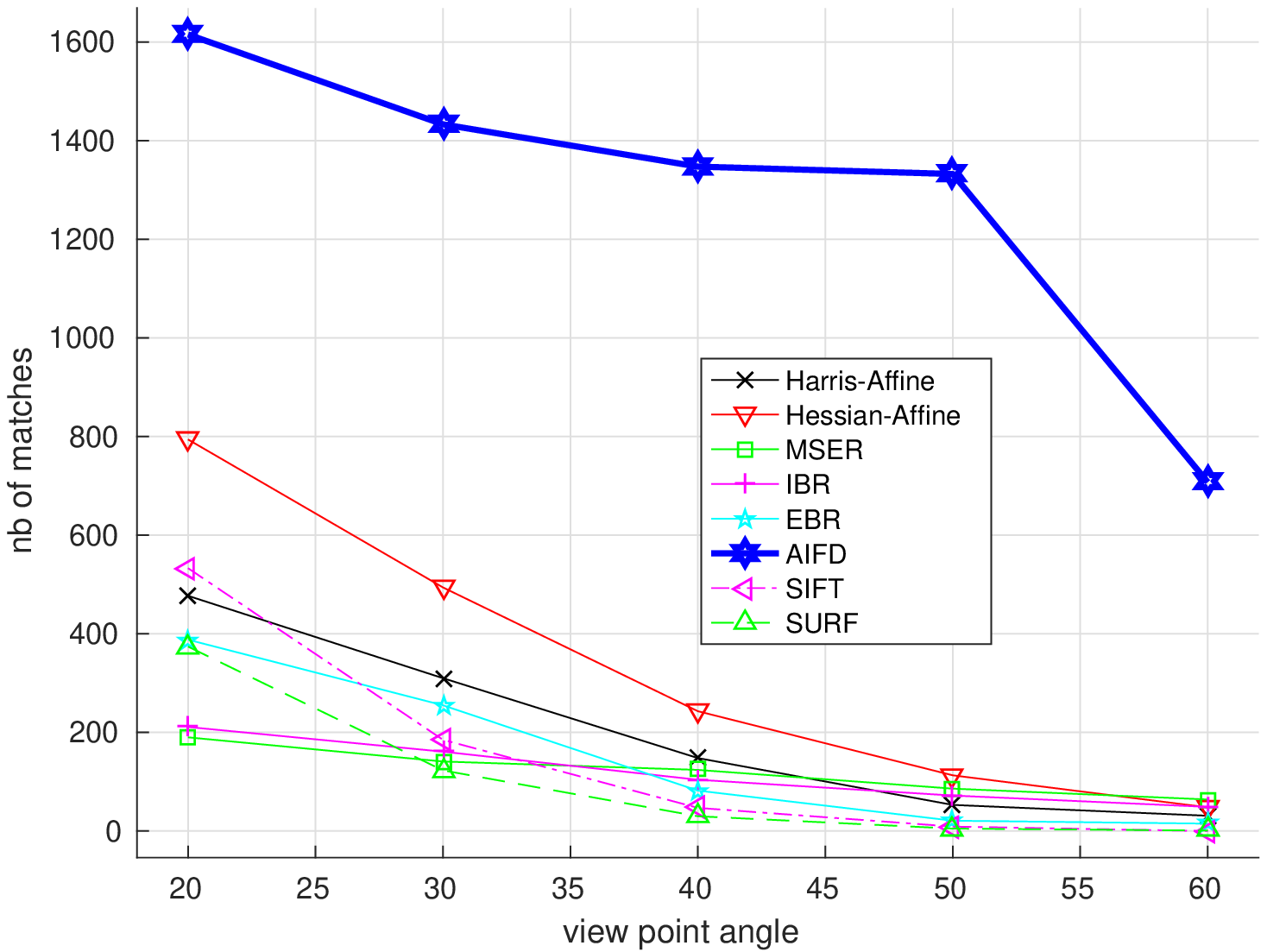}
\small\centerline{(d)}\medskip
\end{minipage}
\caption{The detector and descriptor performance on affine transformed images (graf sequence). (a) \emph{repeatability score}. (b) \emph{number of correspondences}. (c) \emph{matching scores}. (d) \emph{number of matches}. (a), (b) are used to evaluate the performance of the detector; (c), (d) are used to evaluate the performance of the descriptor. Compared to the original Mikolajczyk\rq{}s design, we added the performance of SIFT, SURF and our proposed AIFD. Generally speaking, our proposed AIFD has a much better performance on the affine transformed image sequence. \vspace{18pt}}
\label{fig:affine1_test}
\end{figure*}

\begin{figure*}[!h]
\begin{minipage}{0.49\linewidth}
\centering
\includegraphics[width=0.89\linewidth]{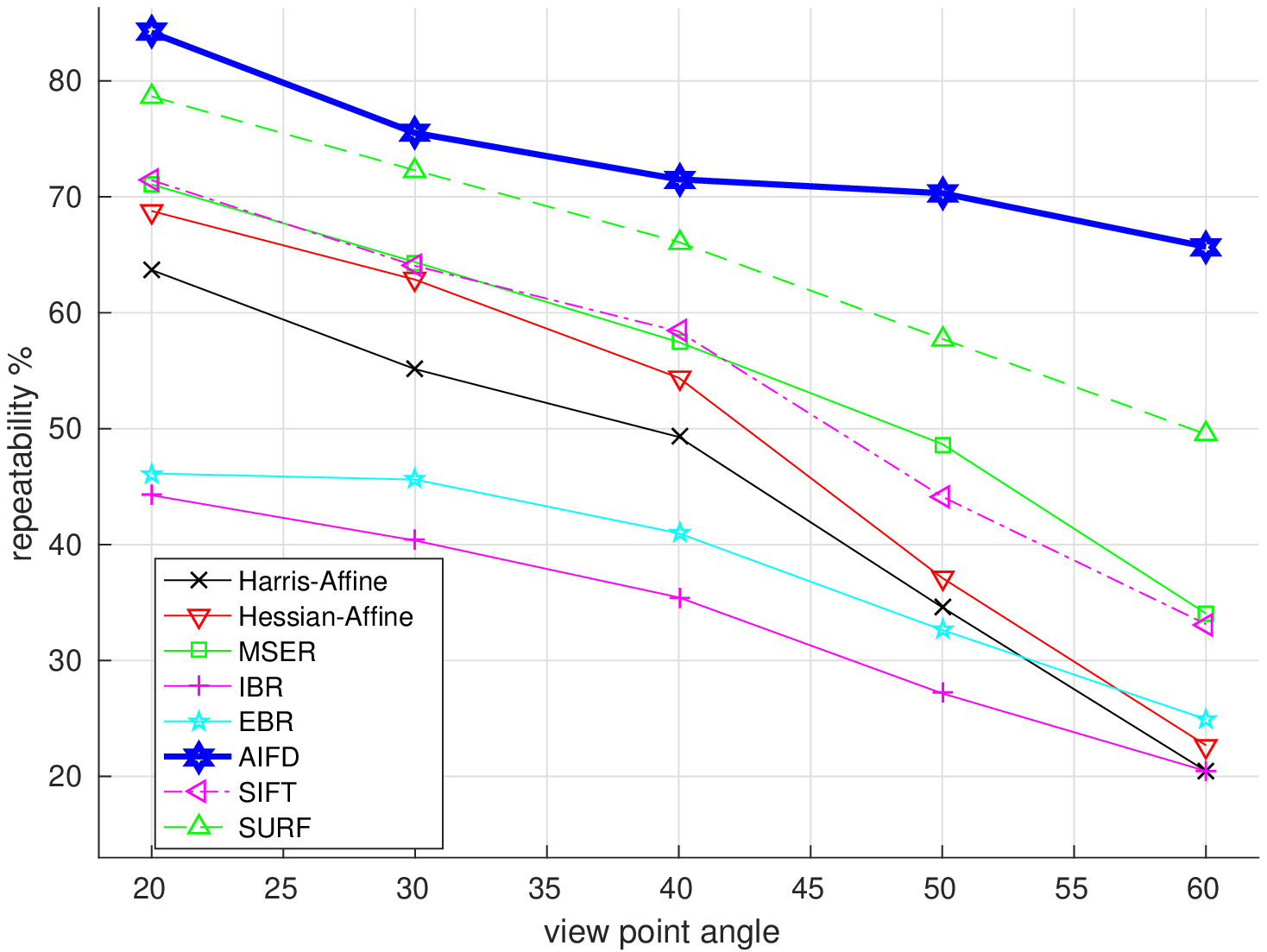}
\small\centerline{(a)}\medskip
\end{minipage}
\begin{minipage}{0.49\linewidth}
\centering
\includegraphics[width=0.89\linewidth]{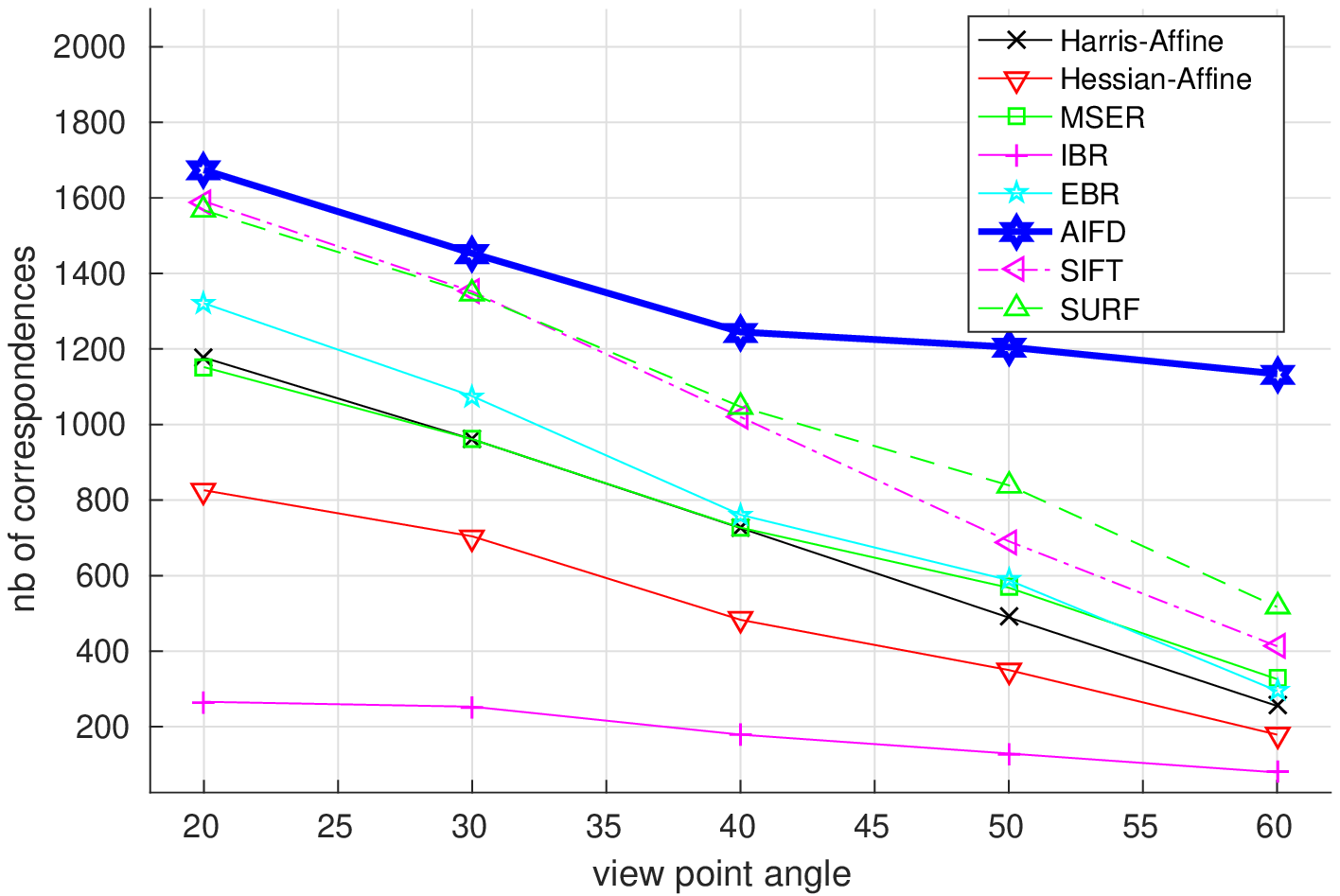}
\small\centerline{(b)}\medskip
\end{minipage}
\end{figure*}

\begin{figure*}[!t] \ContinuedFloat
\begin{minipage}{0.49\linewidth}
\centering
\includegraphics[width=0.89\linewidth]{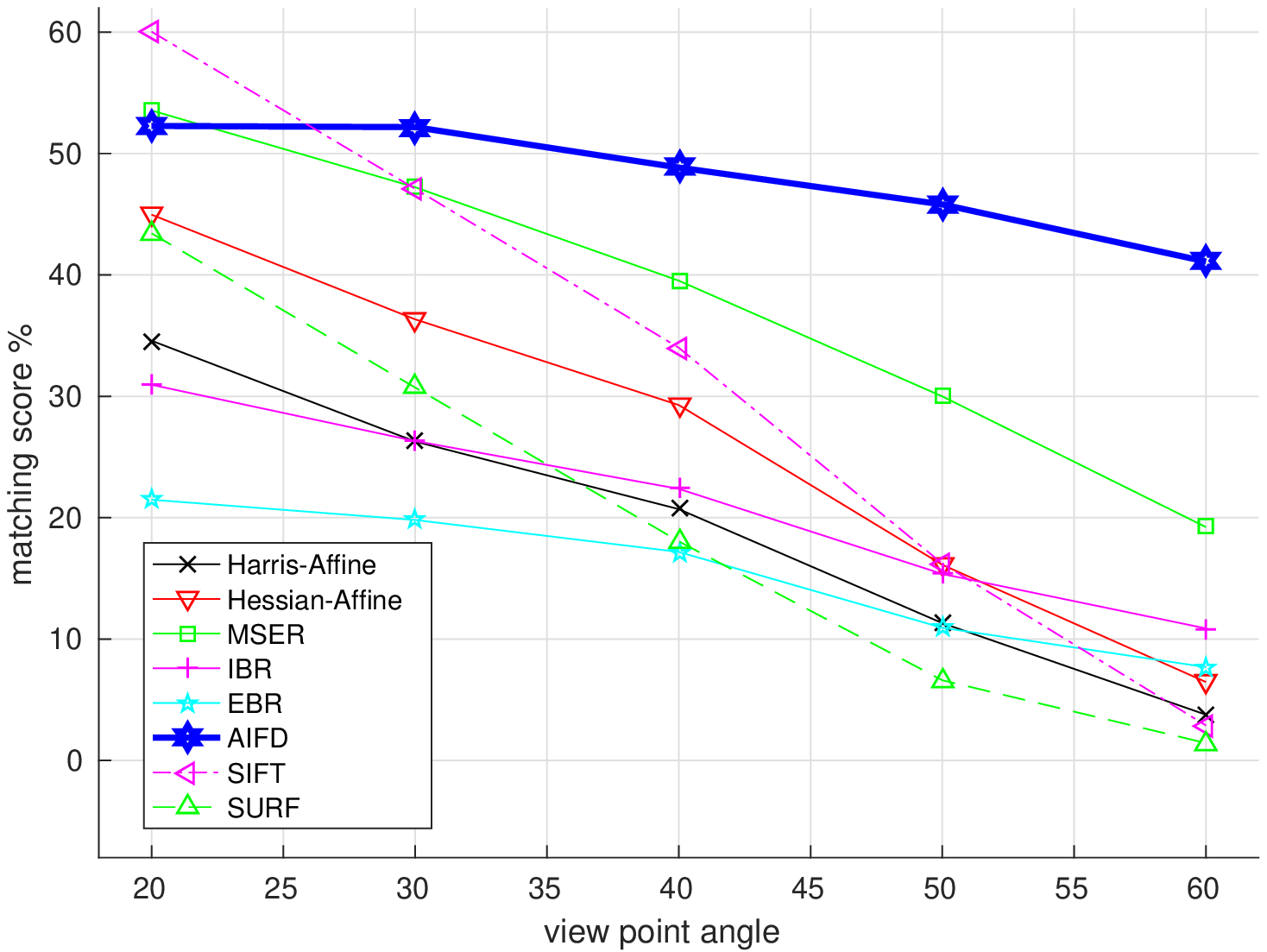}
\small\centerline{(c)}\medskip
\end{minipage}
\begin{minipage}{0.49\linewidth}\centering
\includegraphics[width=0.89\linewidth]{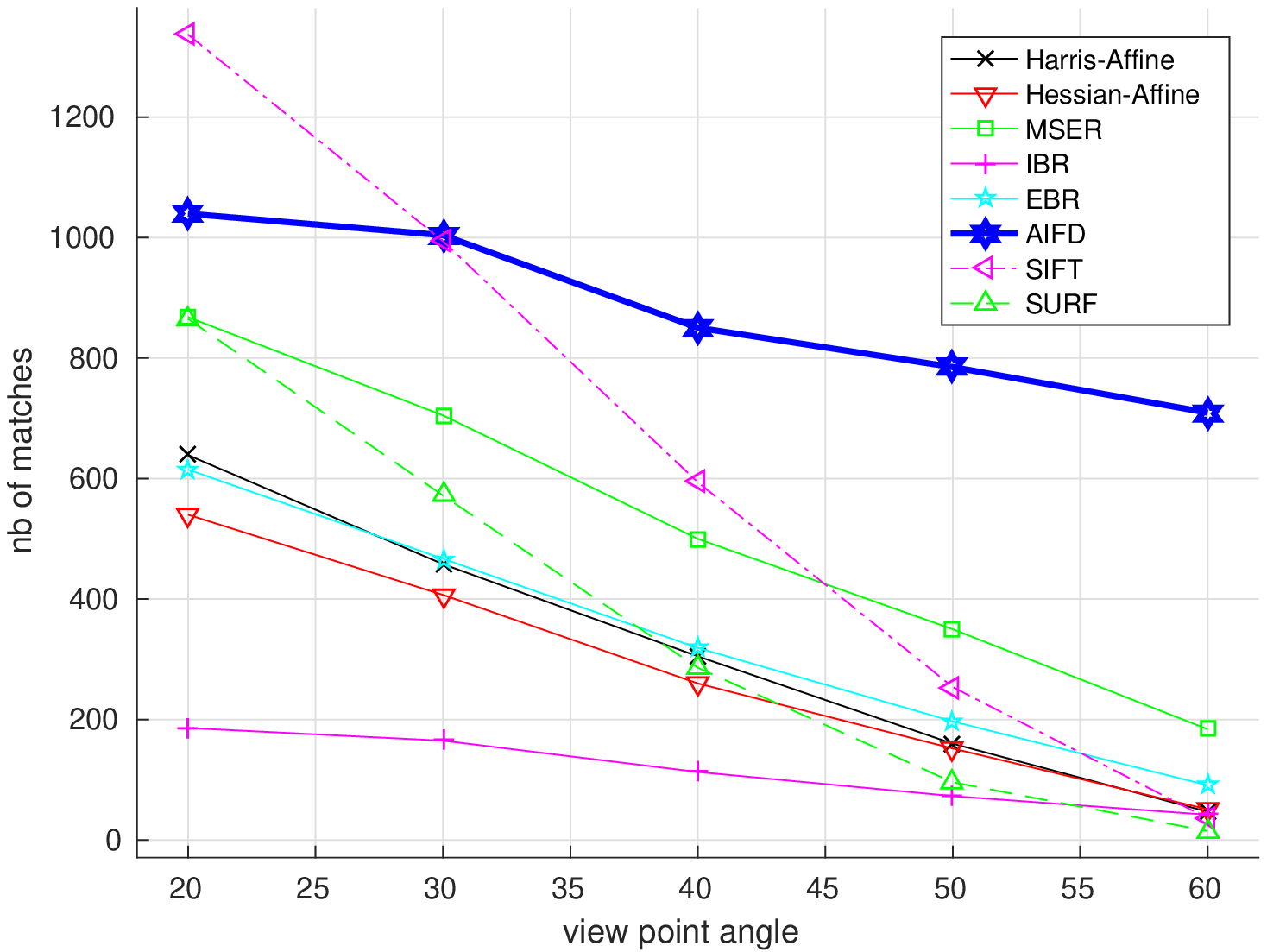}
\small\centerline{(d)}\medskip
\end{minipage}
\caption{The detector and descriptor performance on affine transformed images (wall sequence). (a) \emph{repeatability score}. (b) \emph{number of correspondences}. (c) \emph{matching scores}. (d) \emph{number of matches}. Generally speaking, our proposed AIFD has a much better performance on the affine transformations.}
\label{fig:affine2_test}
\end{figure*}

\begin{figure*}[!t]
\begin{minipage}{0.24\linewidth}
\centering
\includegraphics[width=0.99\linewidth]{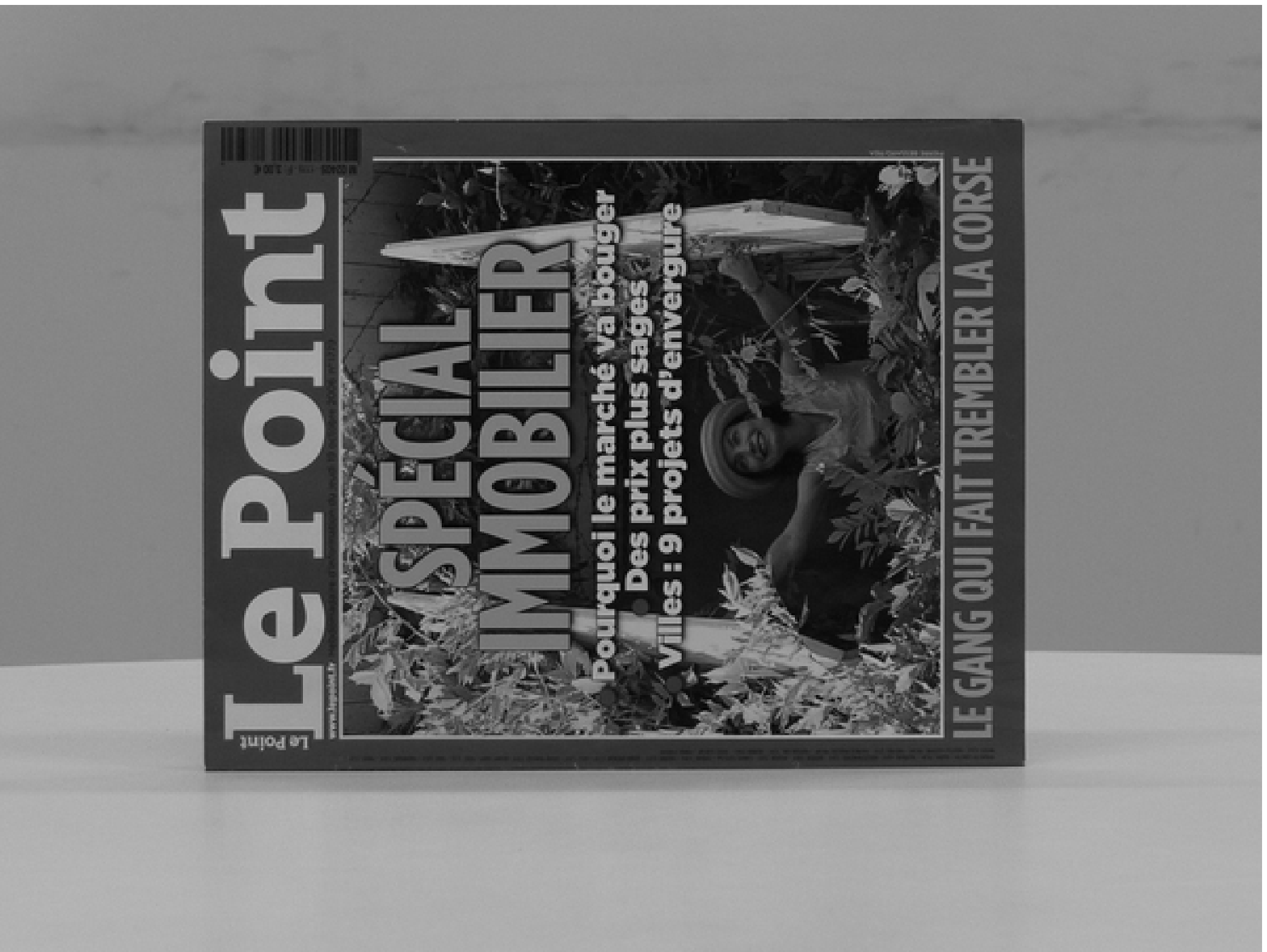}

\end{minipage}
\begin{minipage}{0.24\linewidth}
\centering
\includegraphics[width=0.99\linewidth]{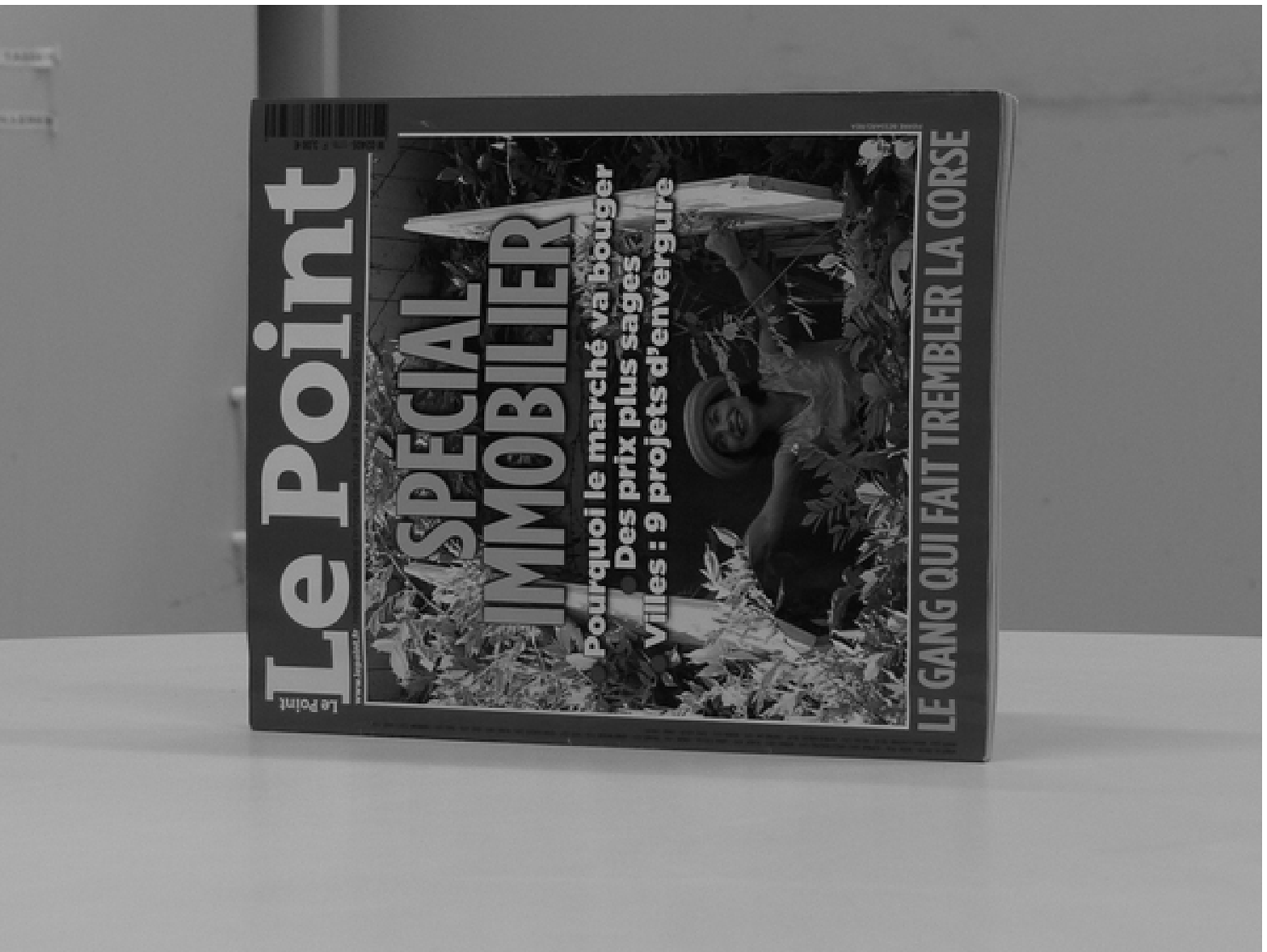}

\end{minipage}
\begin{minipage}{0.24\linewidth}
\centering
\includegraphics[width=0.99\linewidth]{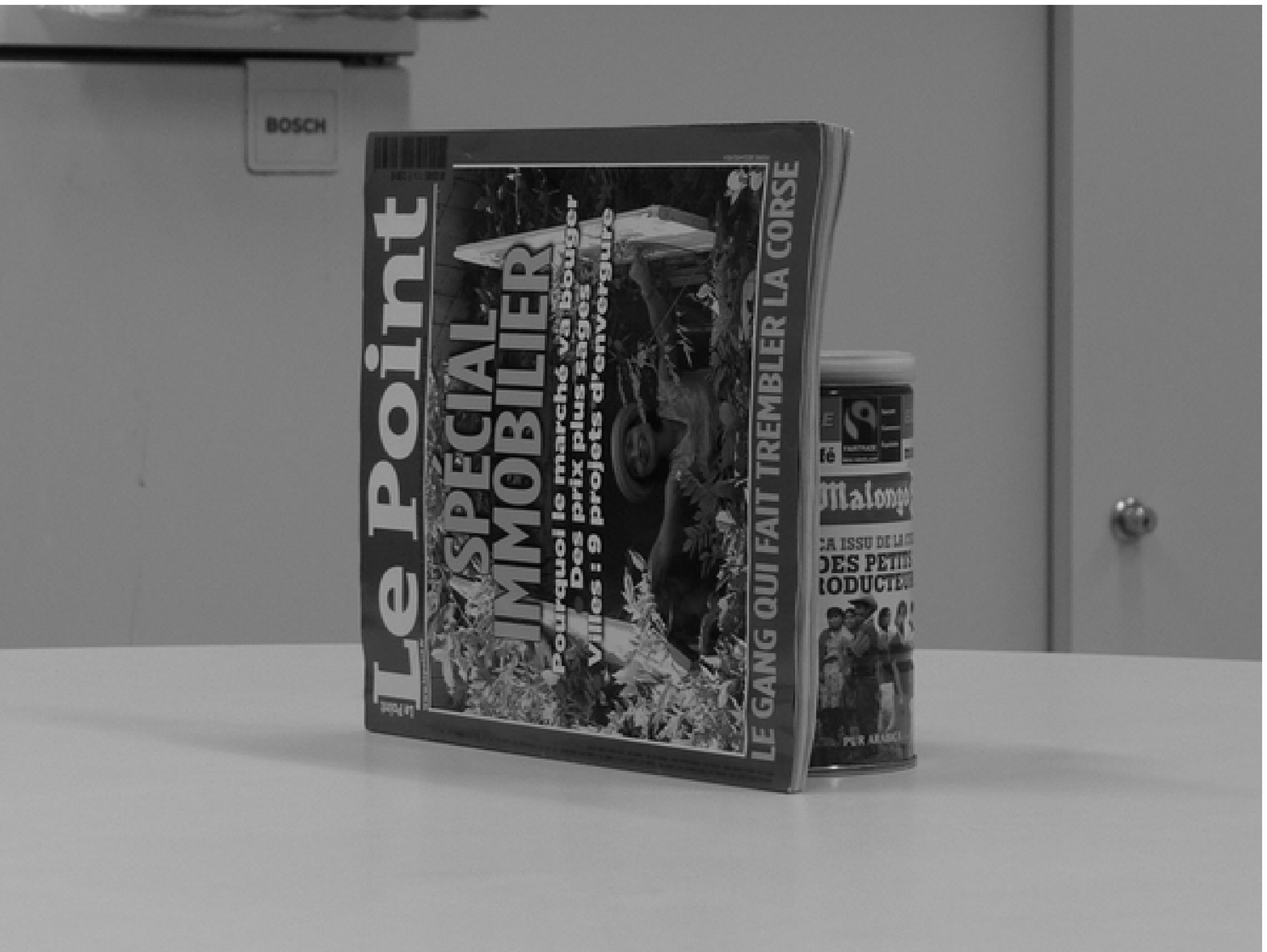}
\end{minipage}
\begin{minipage}{0.24\linewidth}\centering
\includegraphics[width=0.99\linewidth]{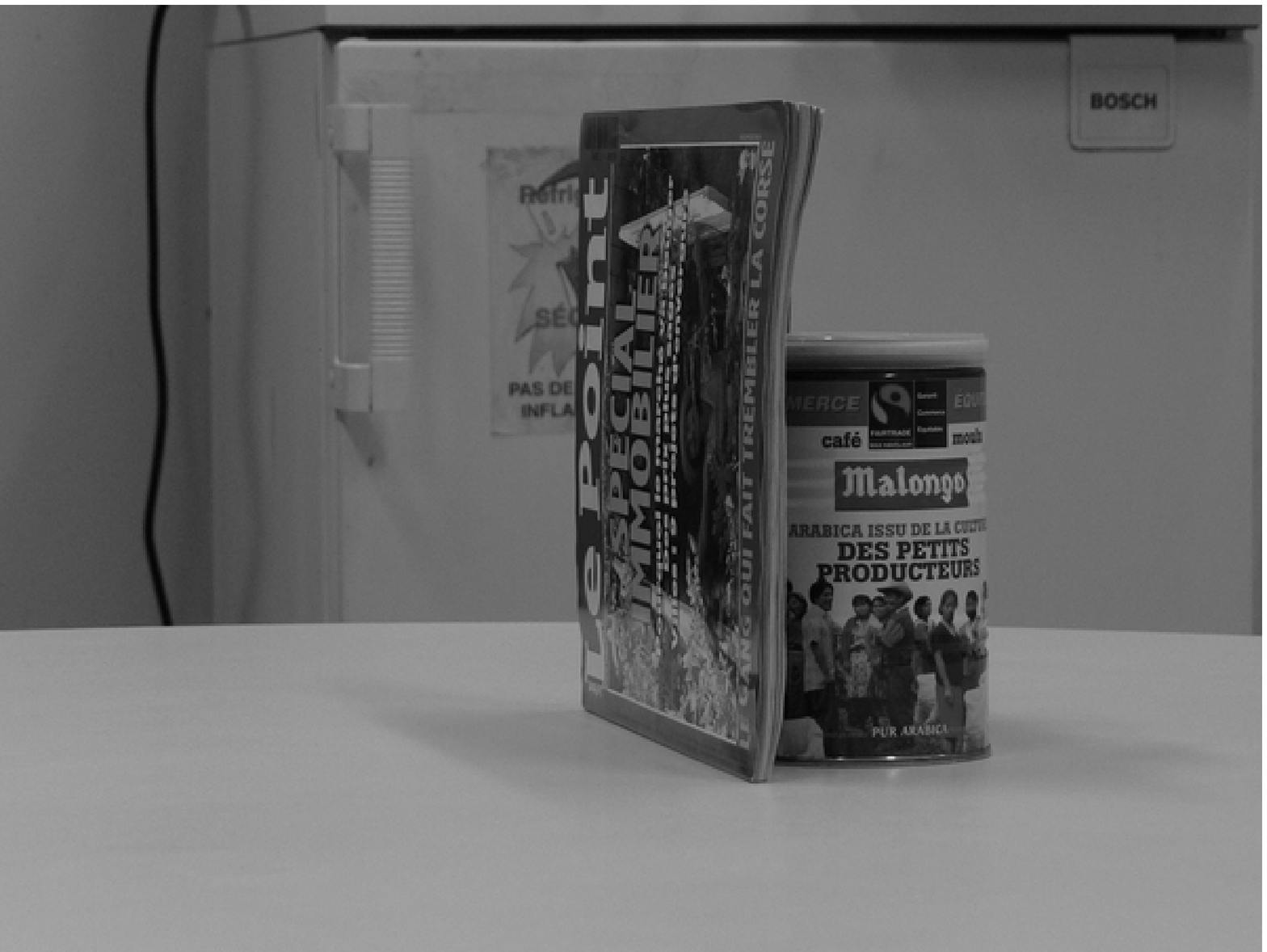}

\end{minipage}
\begin{minipage}{0.24\linewidth}
\centering
\includegraphics[width=0.99\linewidth]{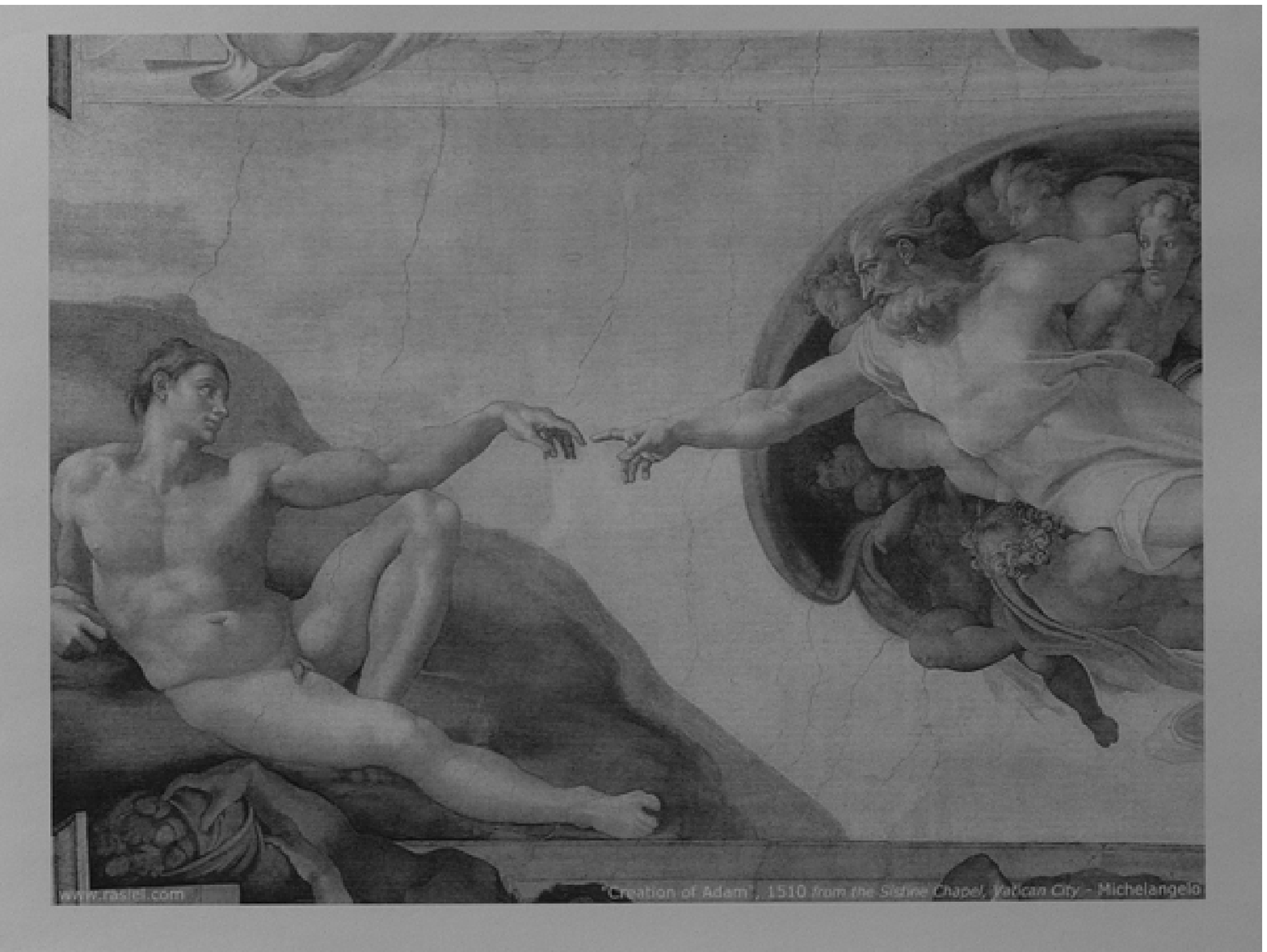}

\end{minipage}
\begin{minipage}{0.24\linewidth}
\centering
\includegraphics[width=0.99\linewidth]{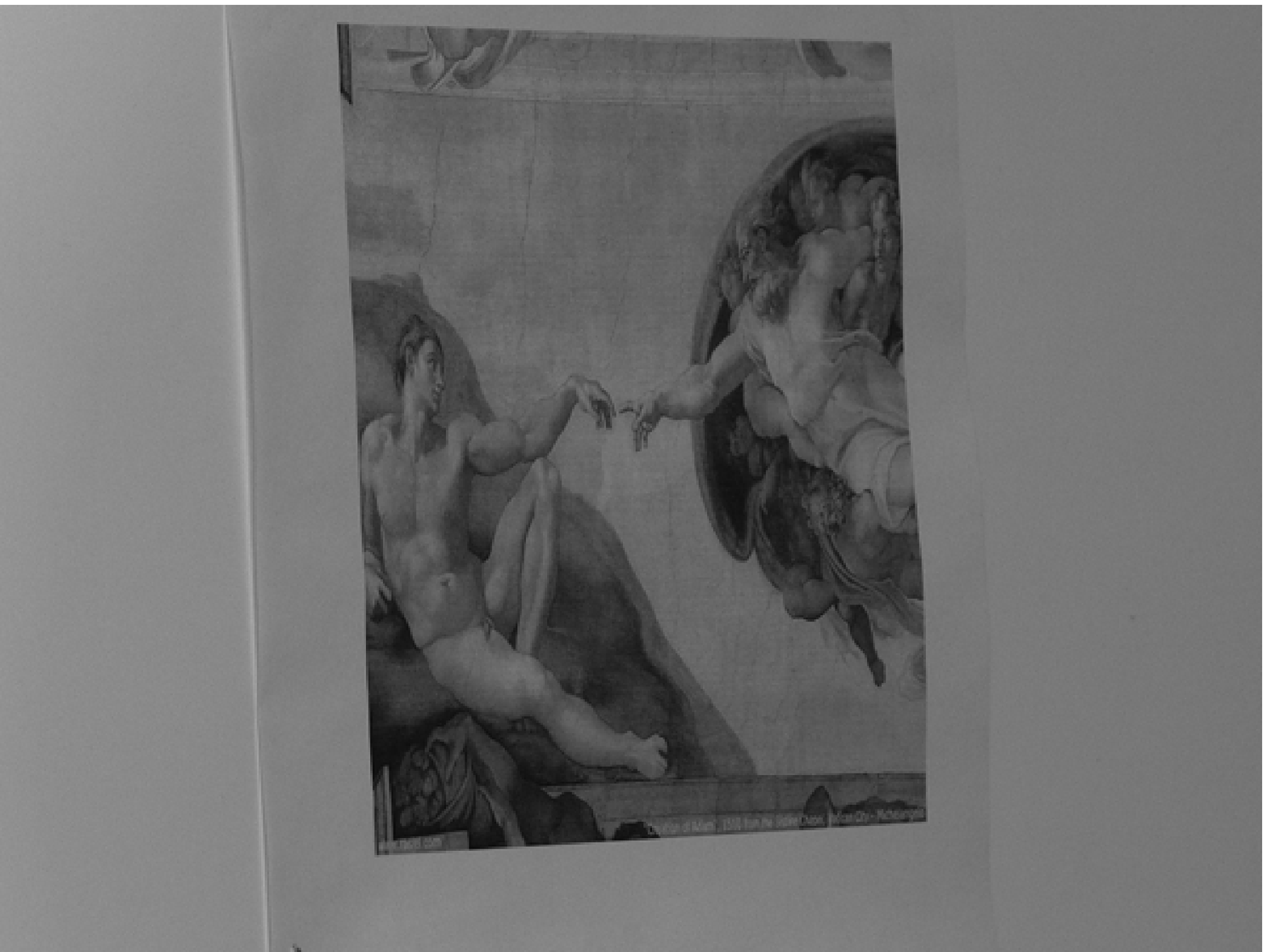}

\end{minipage}
\begin{minipage}{0.24\linewidth}
\centering
\includegraphics[width=0.99\linewidth]{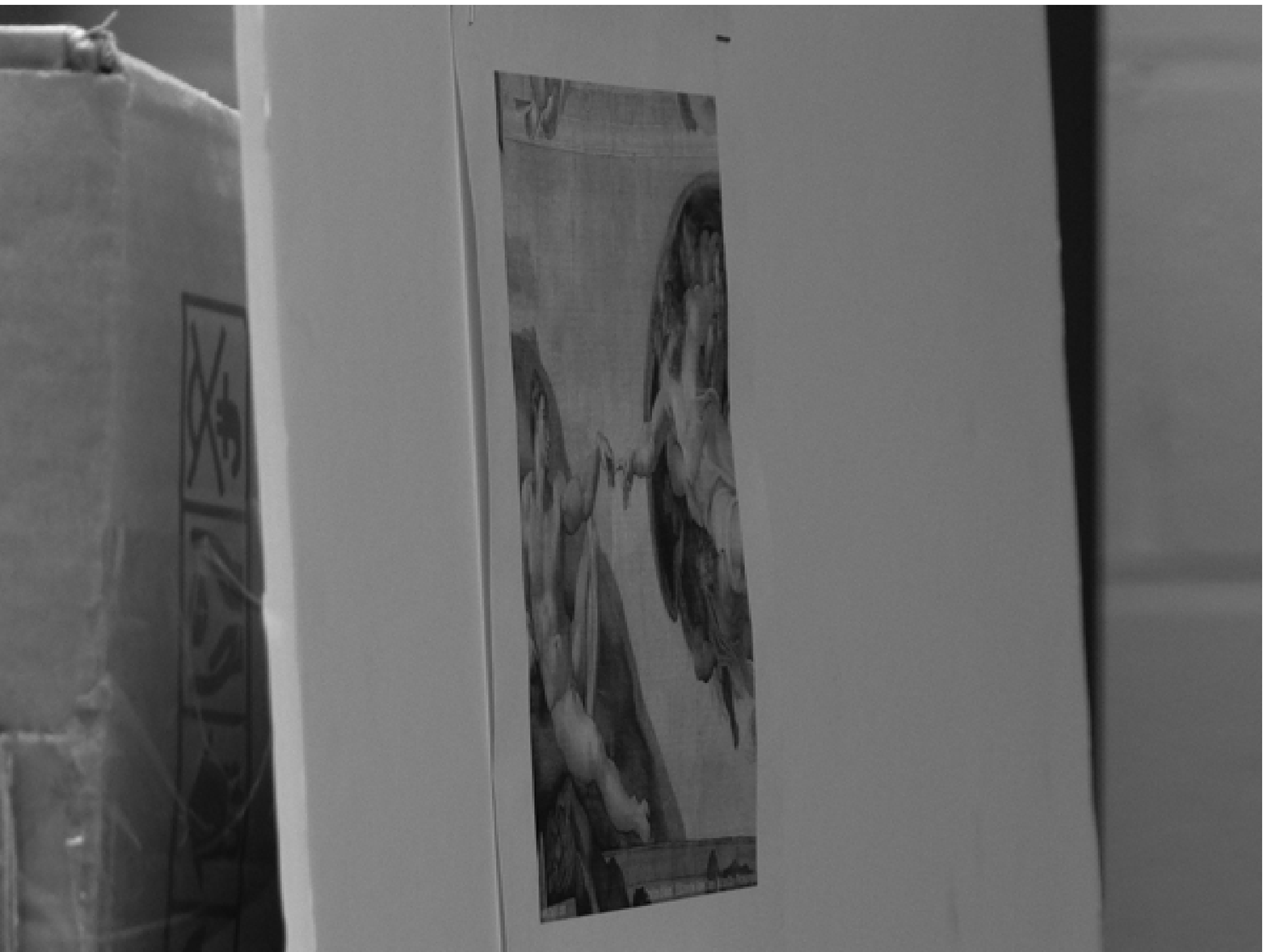}

\end{minipage}
\begin{minipage}{0.24\linewidth}\centering
\includegraphics[width=0.99\linewidth]{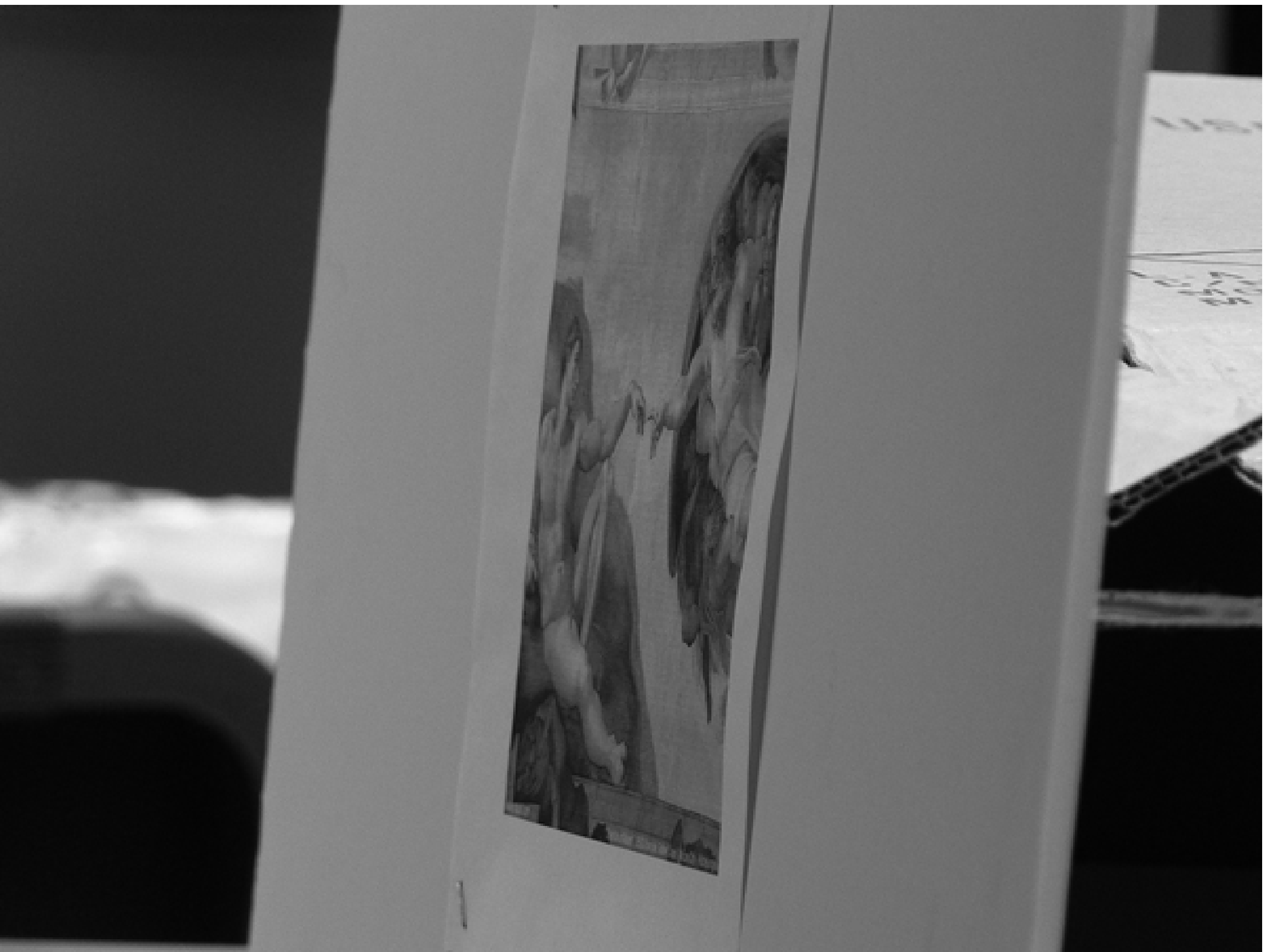}

\end{minipage}
\begin{minipage}{0.24\linewidth}
\centering
\includegraphics[width=0.99\linewidth]{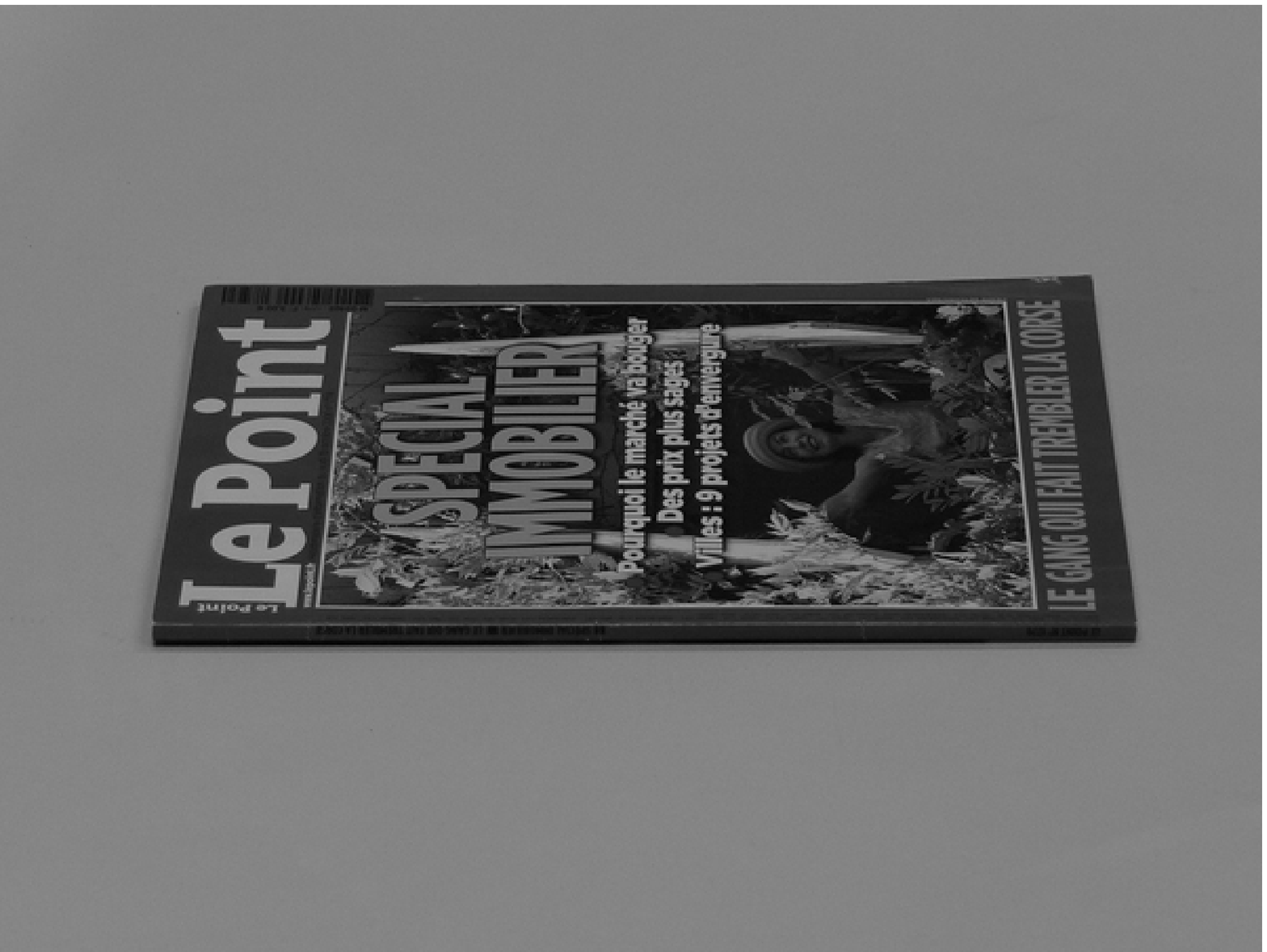}

\end{minipage}
\begin{minipage}{0.24\linewidth}
\centering
\includegraphics[width=0.99\linewidth]{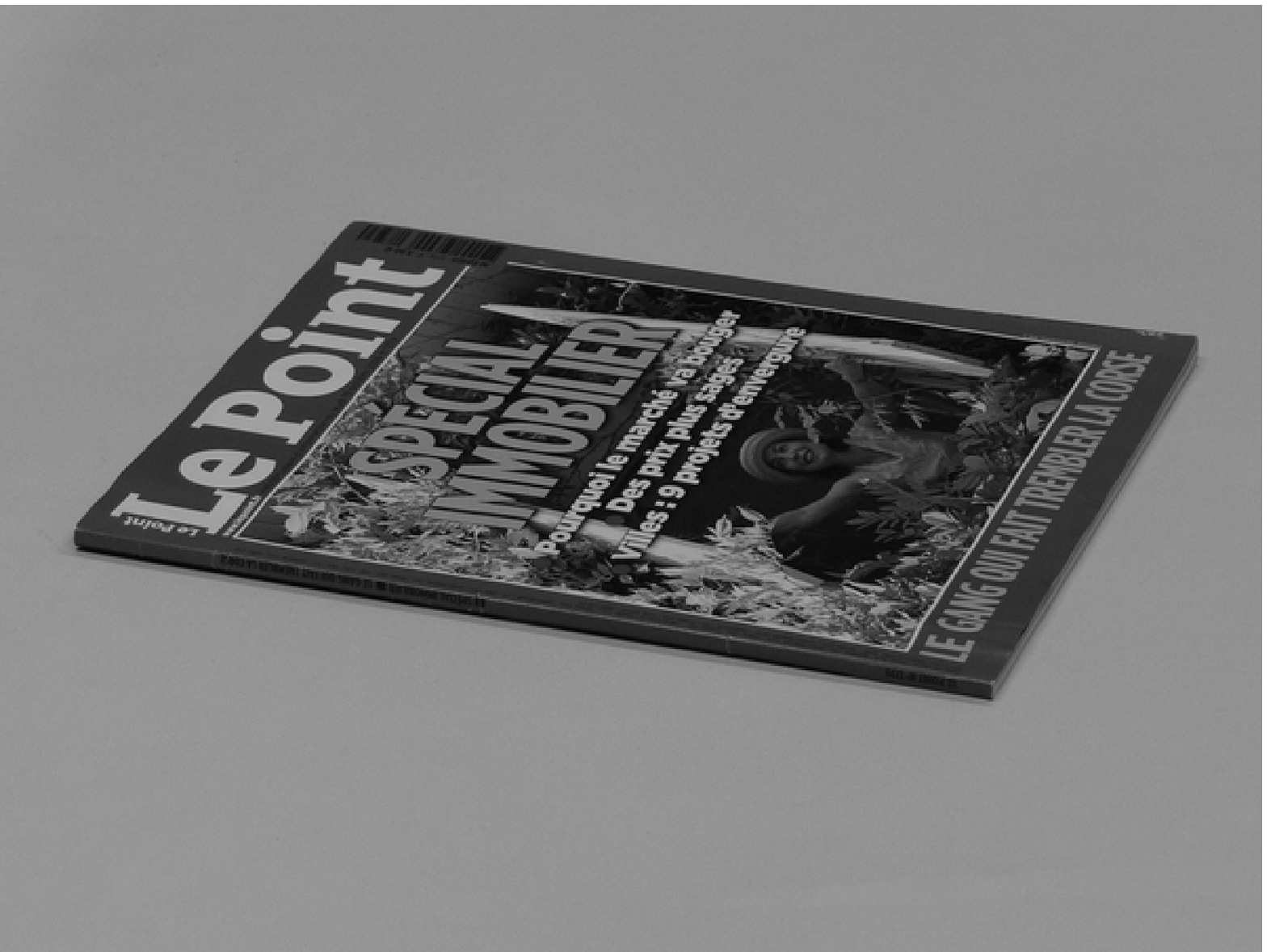}

\end{minipage}
\begin{minipage}{0.24\linewidth}
\centering
\includegraphics[width=0.99\linewidth]{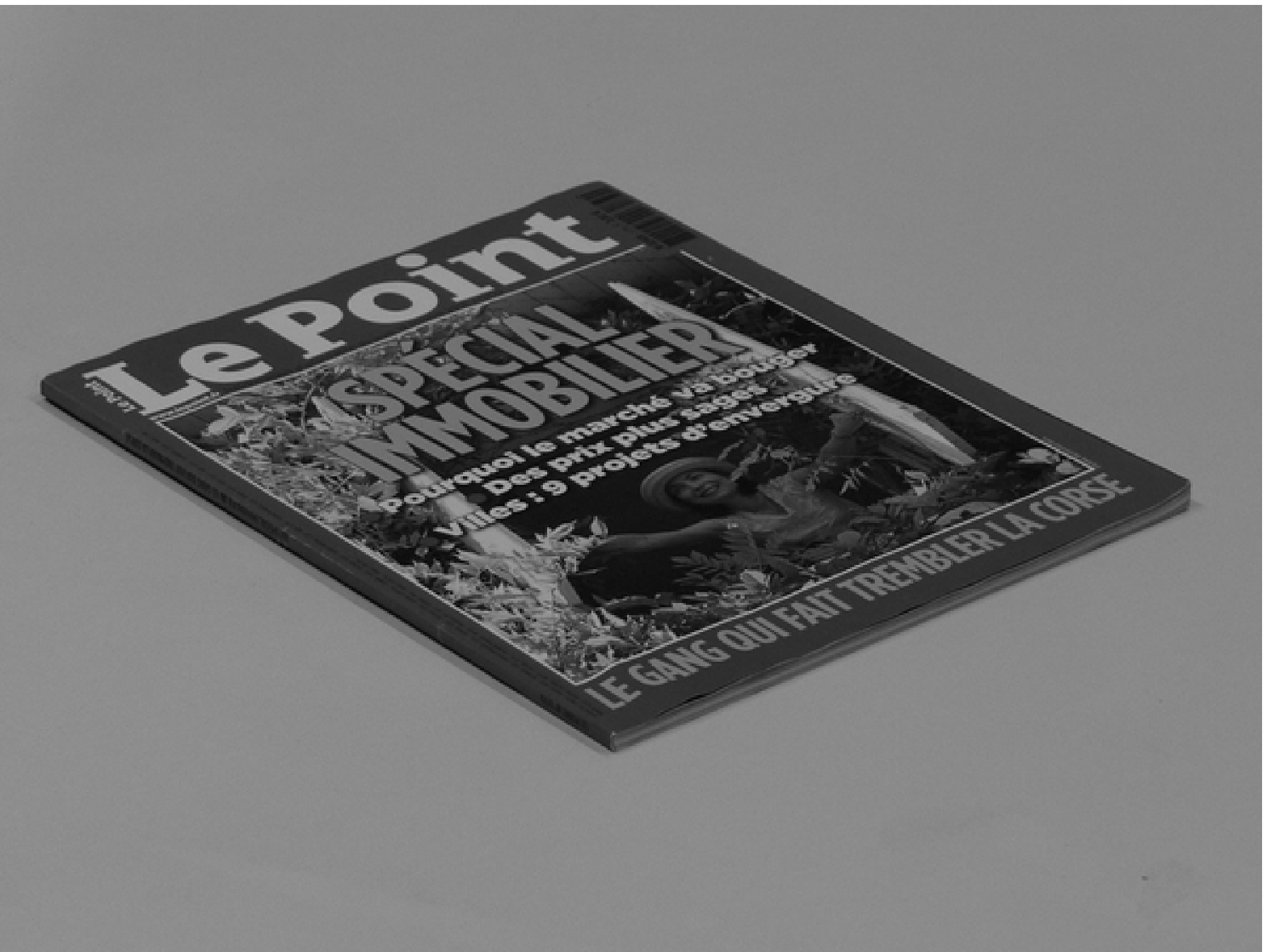}

\end{minipage}
\begin{minipage}{0.24\linewidth}\centering
\includegraphics[width=0.99\linewidth]{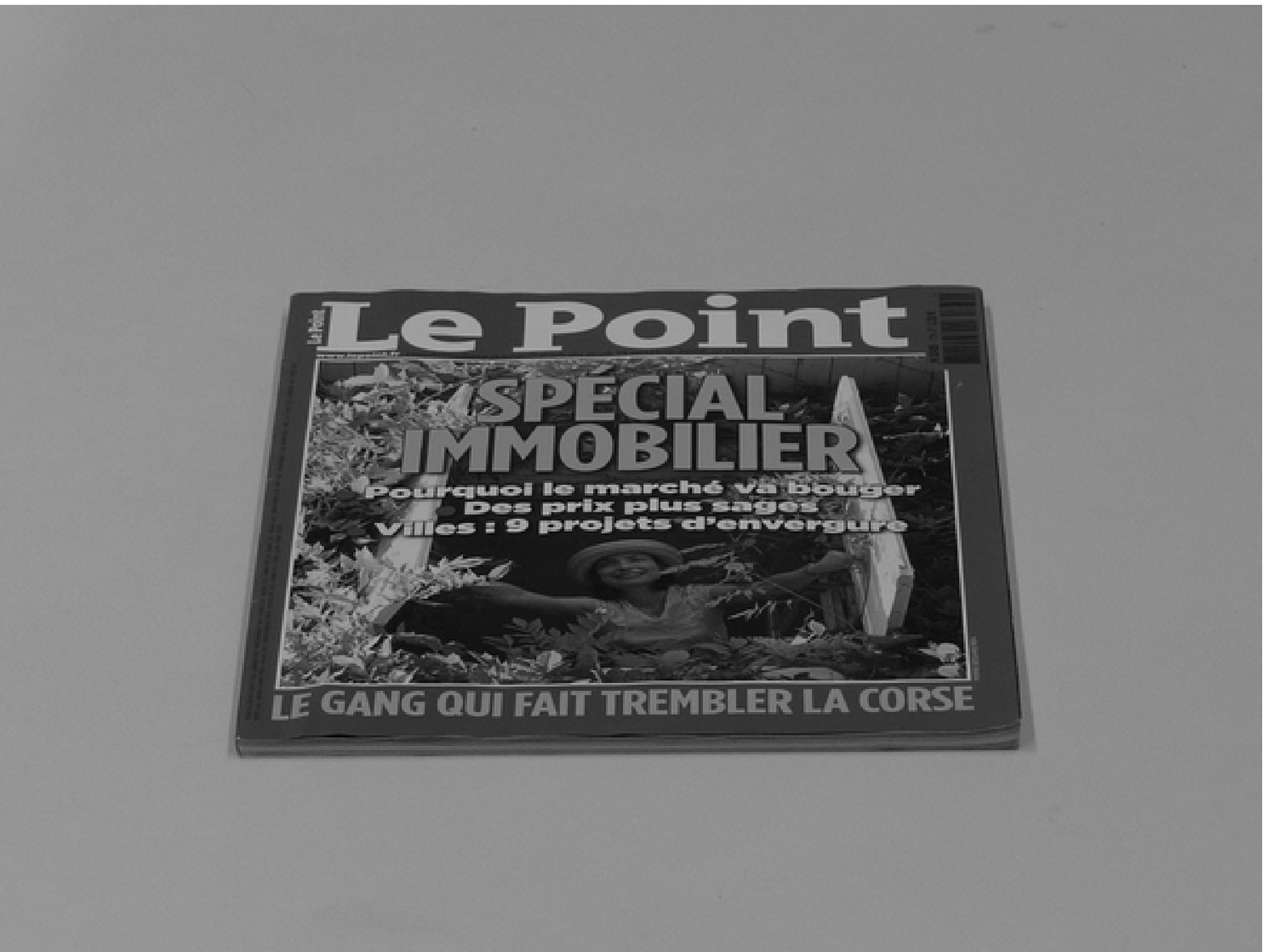}

\end{minipage}
\centering
\caption{ Some example images of \emph{dataset\_ MorelYu09}. It contains more images of extreme view points.}
\label{fig:morelyu09}
\end{figure*}

\begin{figure*}[!t]
\begin{minipage}{0.49\linewidth}
\centering
\includegraphics[width=0.89\linewidth]{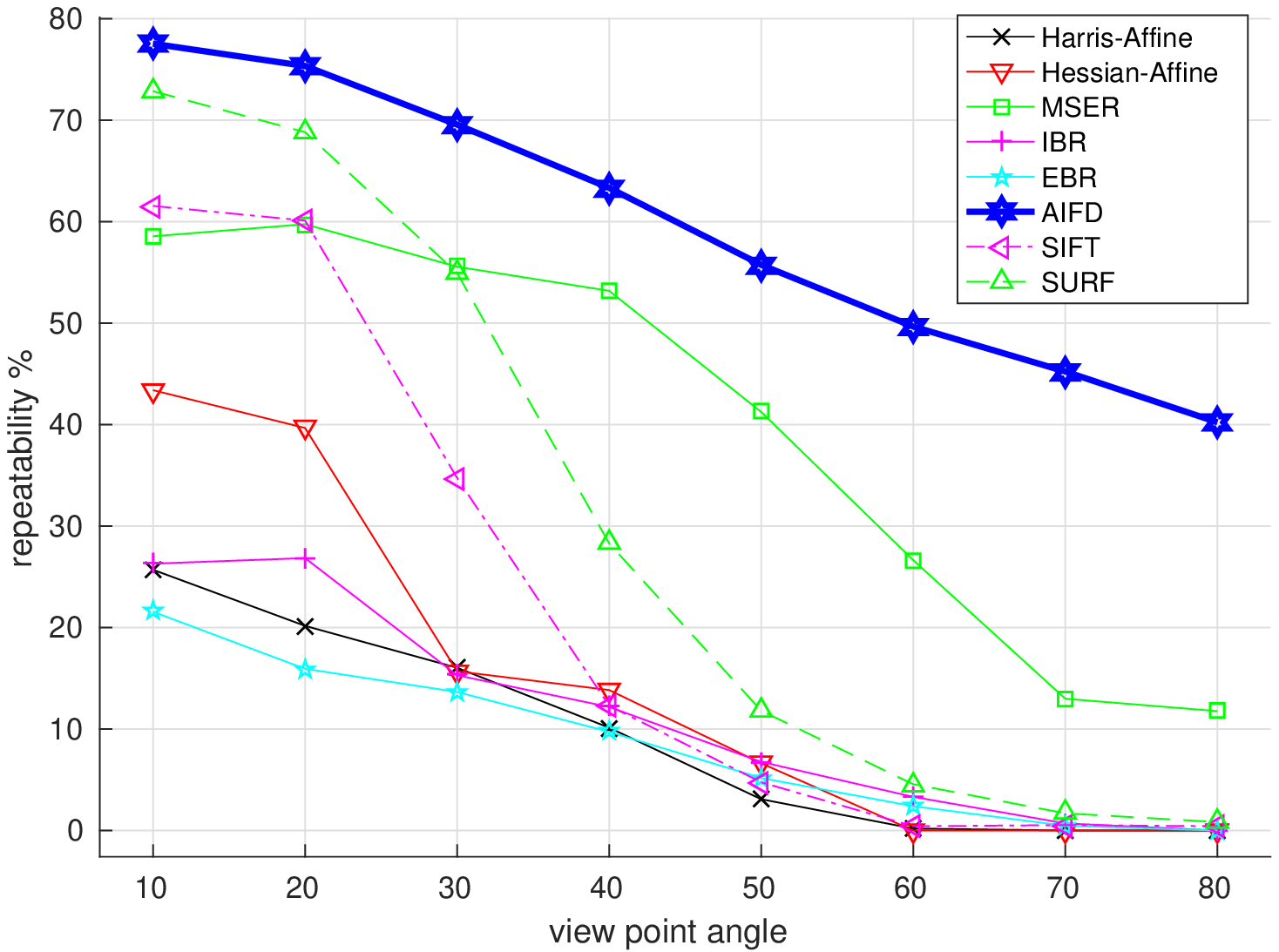}
\small\centerline{(a)}\medskip
\end{minipage}
\begin{minipage}{0.49\linewidth}
\centering
\includegraphics[width=0.89\linewidth]{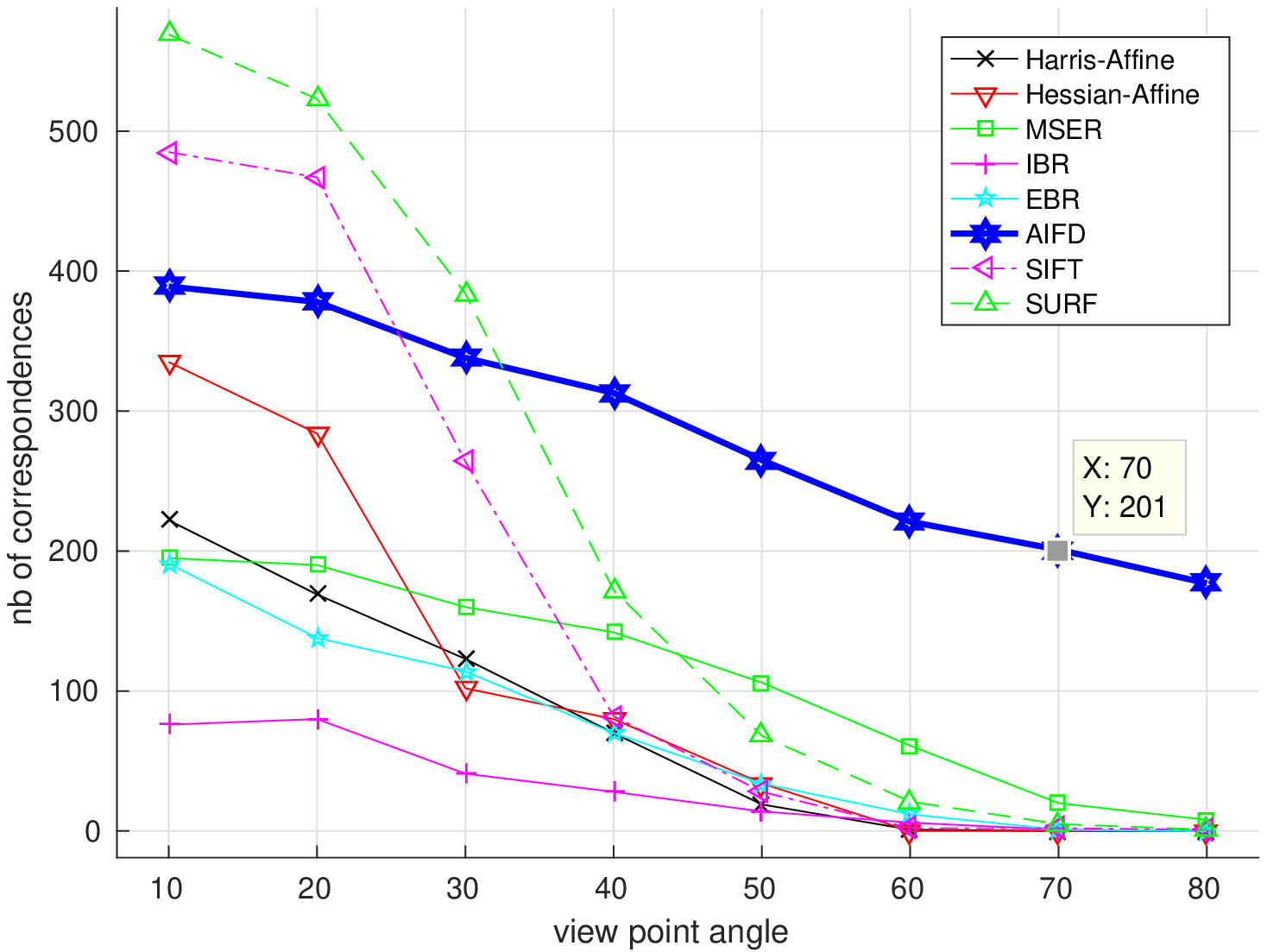}
\small\centerline{(b)}\medskip
\end{minipage}
\begin{minipage}{0.49\linewidth}
\centering
\includegraphics[width=0.89\linewidth]{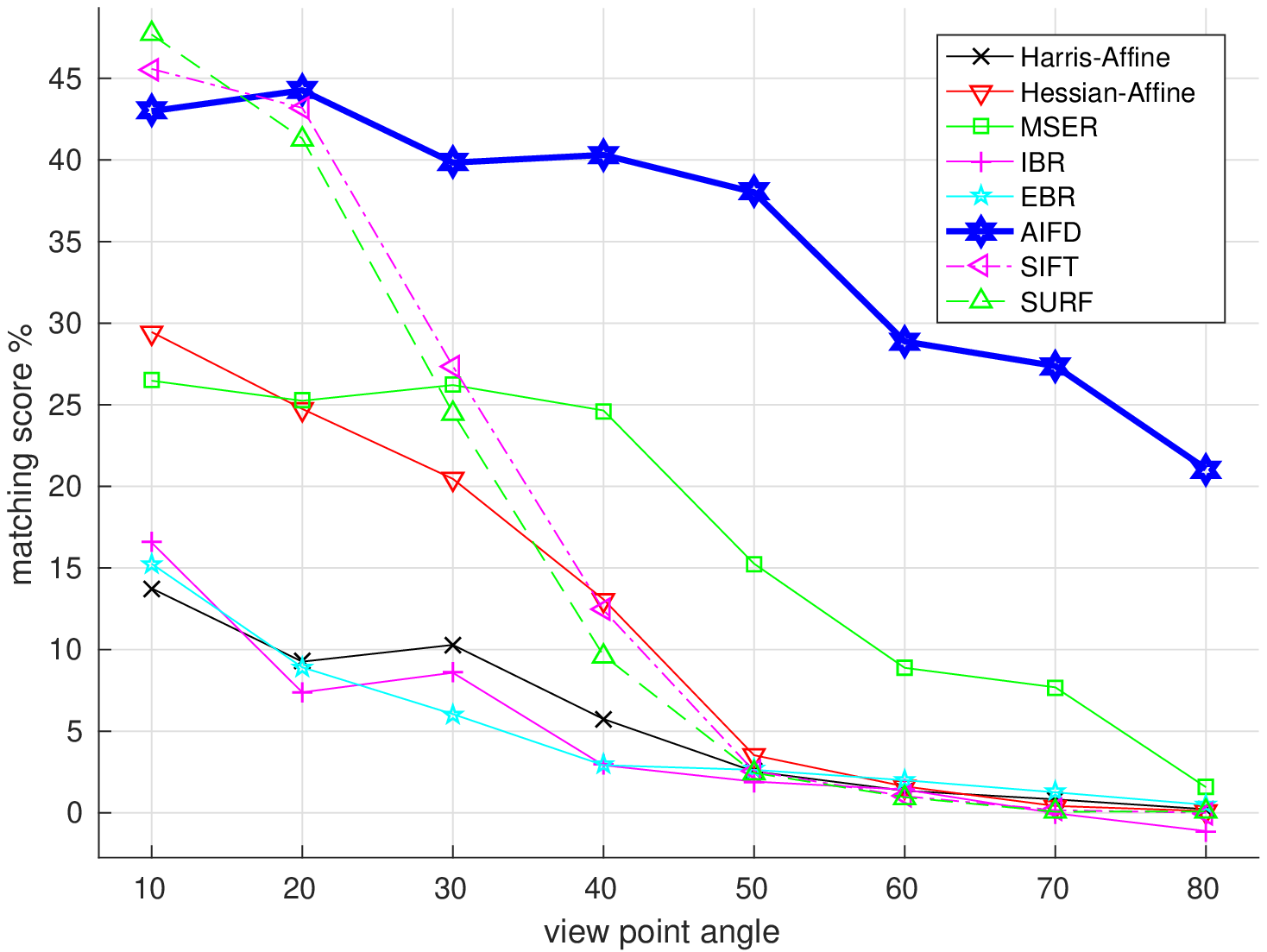}
\small\centerline{(c)}\medskip
\end{minipage}
\begin{minipage}{0.49\linewidth}\centering
\includegraphics[width=0.89\linewidth]{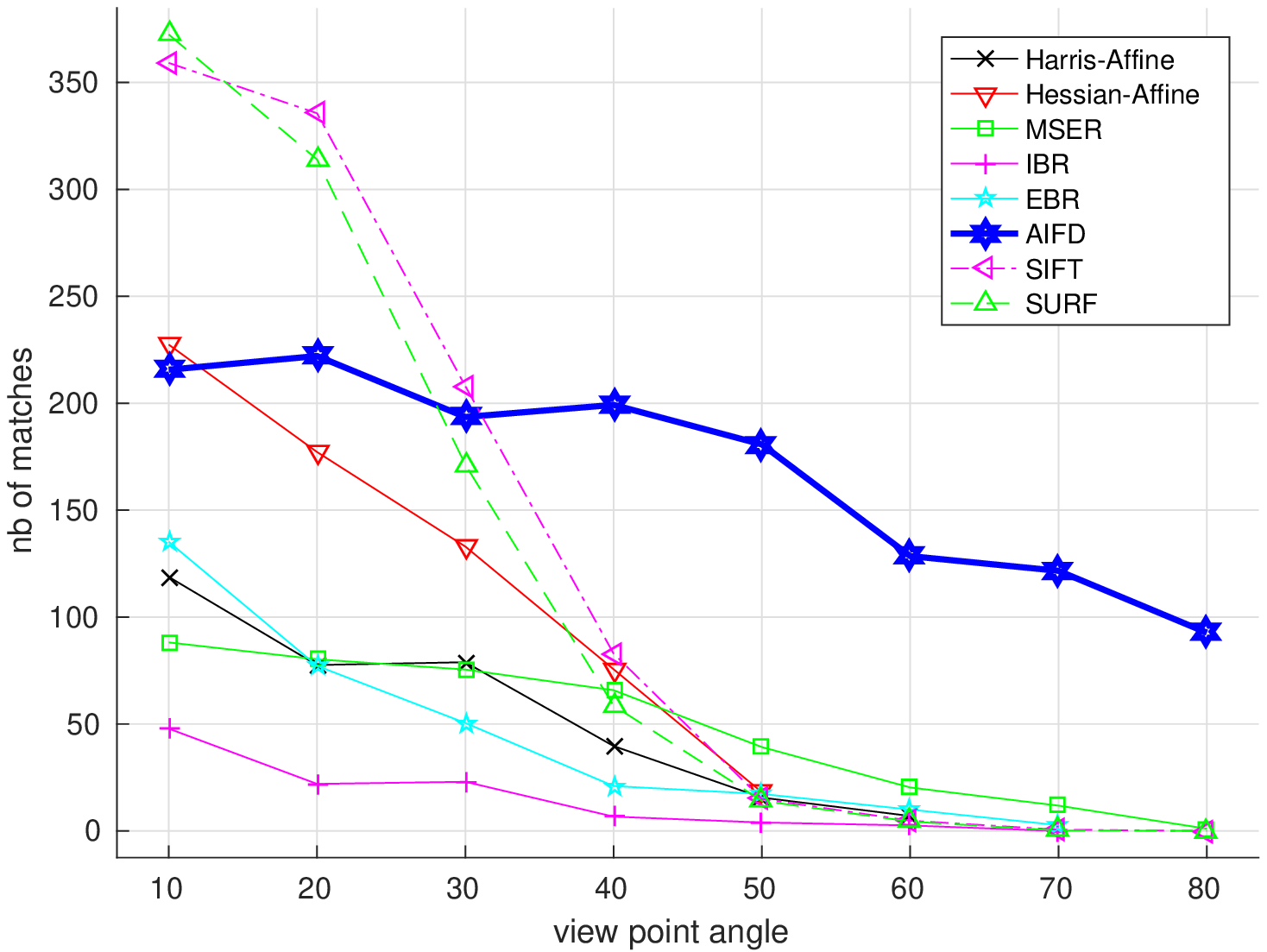}
\small\centerline{(d)}\medskip
\end{minipage}
\caption{The detector and descriptor performance on affine transformed images (front magazine sequence). (a) \emph{repeatability score}. (b) \emph{number of correspondences}. (c) \emph{matching scores}. (d) \emph{number of matches}. Generally speaking, our proposed AIFD has a much better performance on the affine transformed image sequence. \vspace{18pt}}
\label{fig:affine1_test}
\end{figure*}

\begin{figure*}[!h]
\begin{minipage}{0.49\linewidth}
\centering
\includegraphics[width=0.89\linewidth]{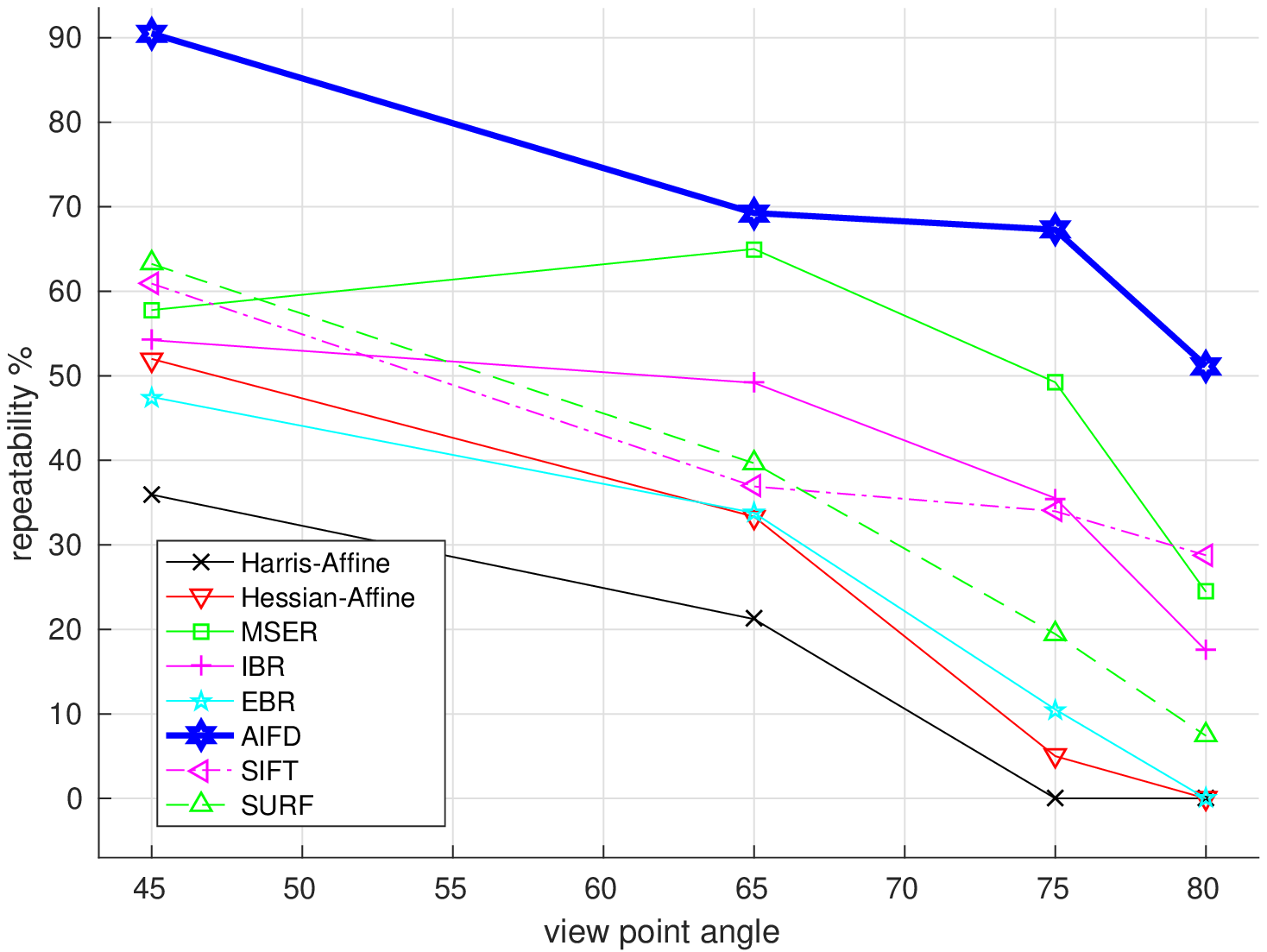}
\small\centerline{(a)}\medskip
\end{minipage}
\begin{minipage}{0.49\linewidth}
\centering
\includegraphics[width=0.89\linewidth]{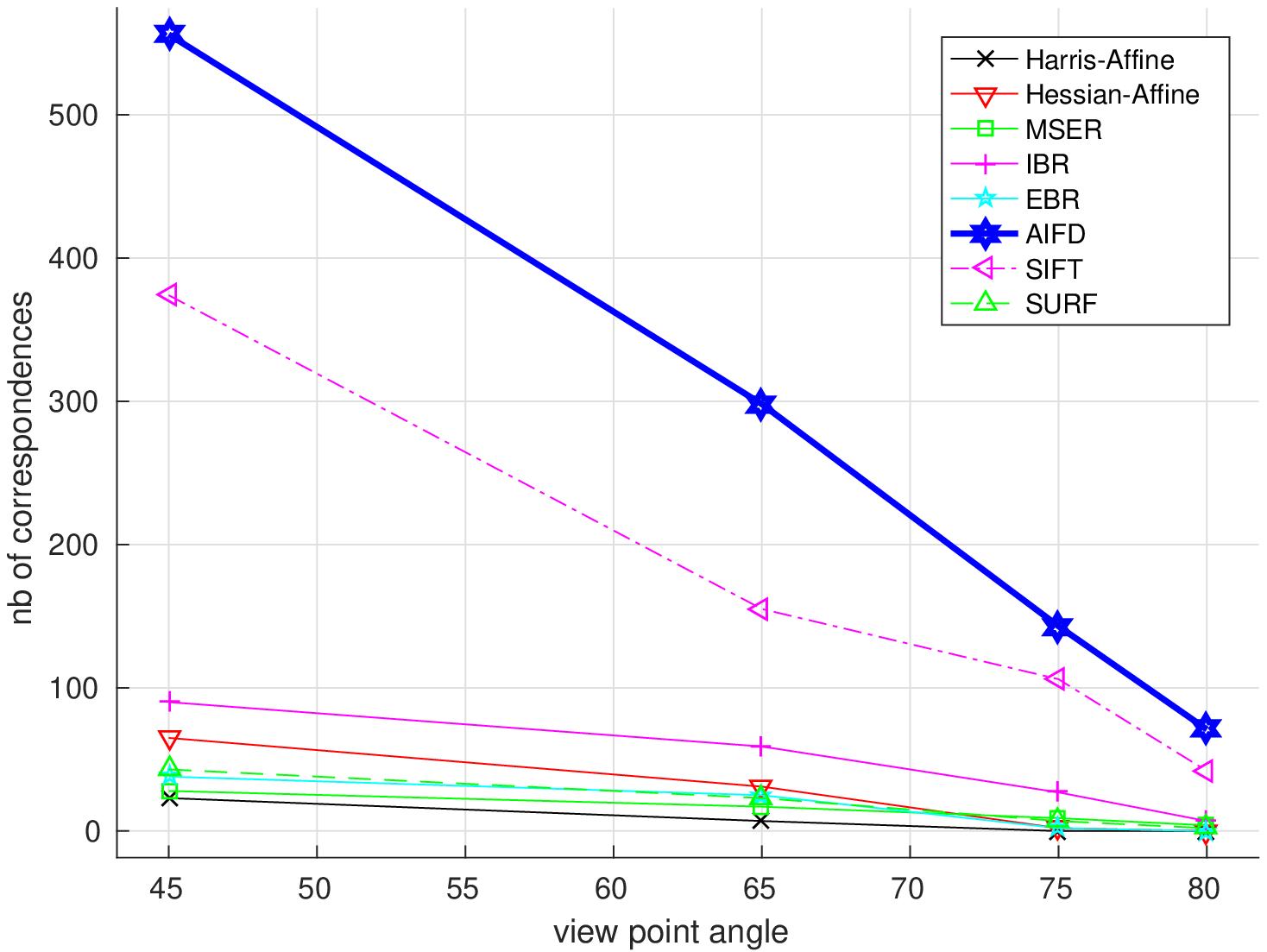}
\small\centerline{(b)}\medskip
\end{minipage}
\end{figure*}

\begin{figure*}[!t] \ContinuedFloat
\begin{minipage}{0.49\linewidth}
\centering
\includegraphics[width=0.89\linewidth]{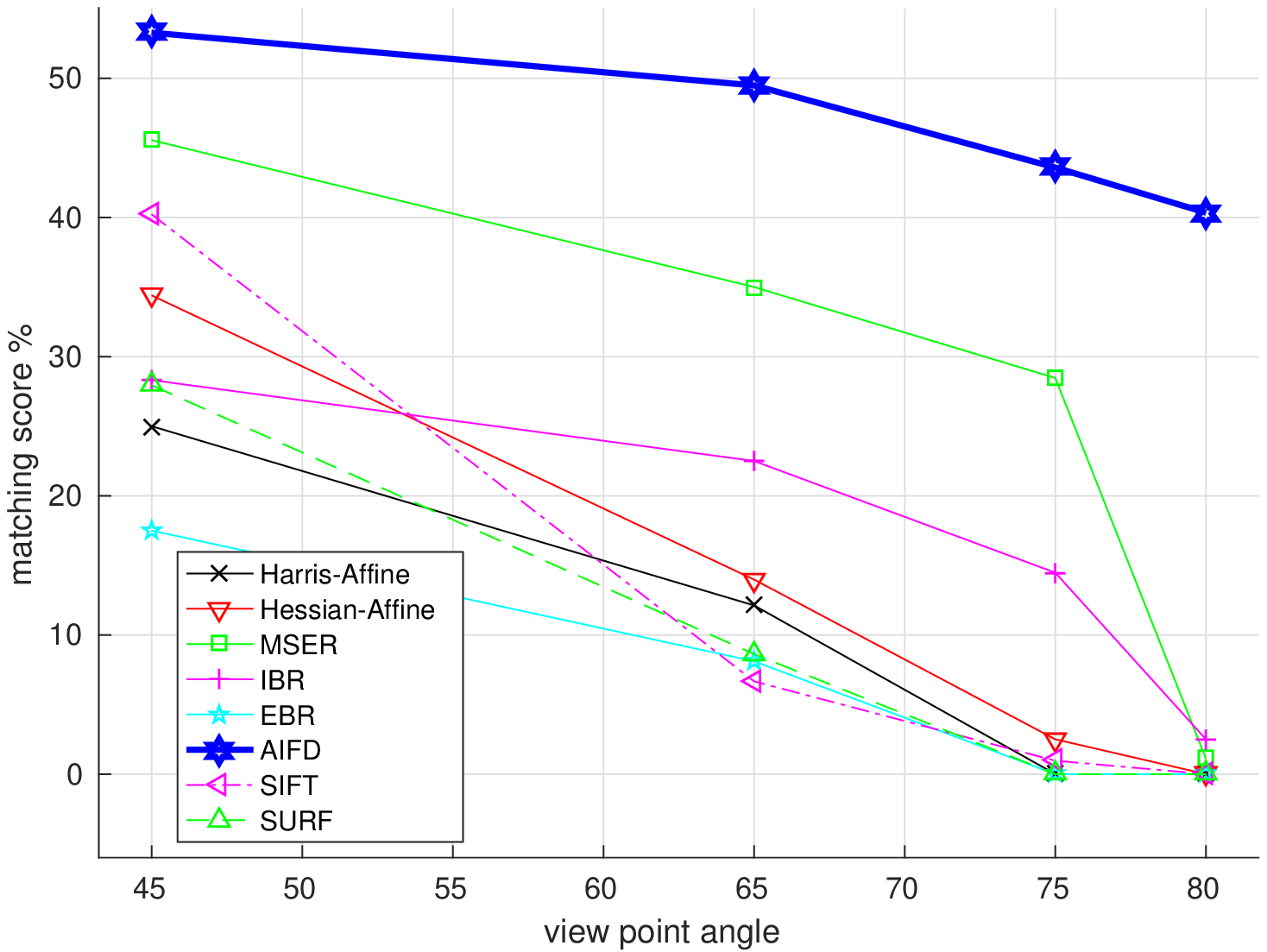}
\small\centerline{(c)}\medskip
\end{minipage}
\begin{minipage}{0.49\linewidth}\centering
\includegraphics[width=0.89\linewidth]{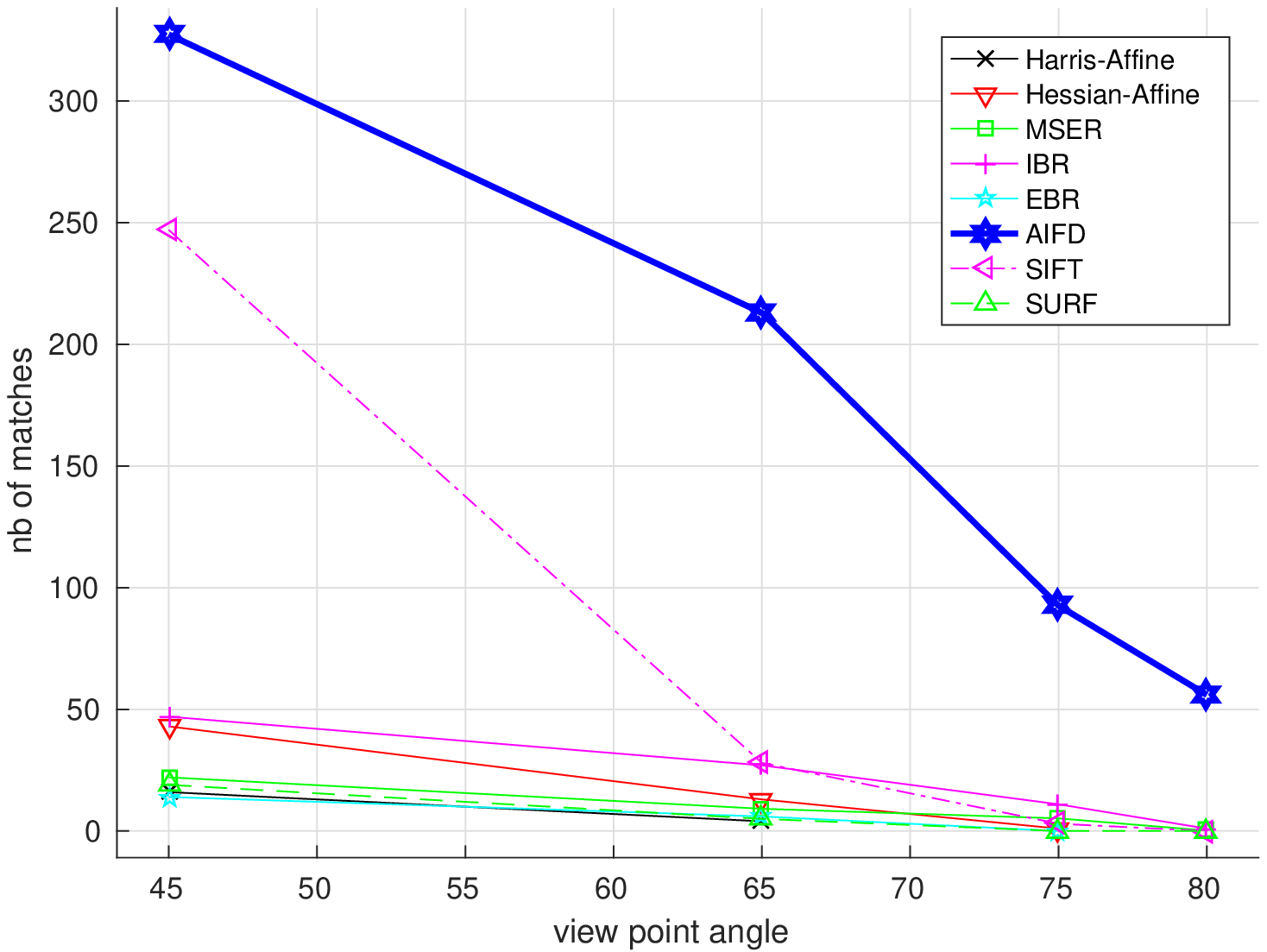}
\small\centerline{(d)}\medskip
\end{minipage}
\caption{The detector and descriptor performance on affine transformed images (printing\_left sequence). (a) \emph{repeatability score}. (b) \emph{number of correspondences}. (c) \emph{matching scores}. (d) \emph{number of matches}. Generally speaking, our proposed AIFD has a much better performance on the affine transformed image sequence.}
\label{fig:morelyu09_2}
\end{figure*}

To have a better comparison with other feature matching algorithms, we introduce the image sequences and software provided by  Mikolajczyk\footnote{http://www.robots.ox.ac.uk/$\sim$vgg/research/affine/} \cite{mikolajczyk2005performance} to test our proposed AIFD and some other state-of-the-art feature matching algorithms. These sequences are consisted by the images of different types and levels of geometric and photometric transformations, especially affine transformations. At the beginning of this section, the experiments concerning the affine resilience are specially stressed, complemented by some more experiments with images at extreme view points. Apart from the resilience to affine transformations, our proposed AIFD also perform well on different types of geometric and photometric transformations, like the change of zoom, rotation, blur, illumination and compression, etc. It can be proven by the experiments on some other Mikolajczyk provided image sequences.

\begin{figure*}[!t]
\begin{minipage}{0.49\linewidth}
\centering
\includegraphics[width=0.89\linewidth]{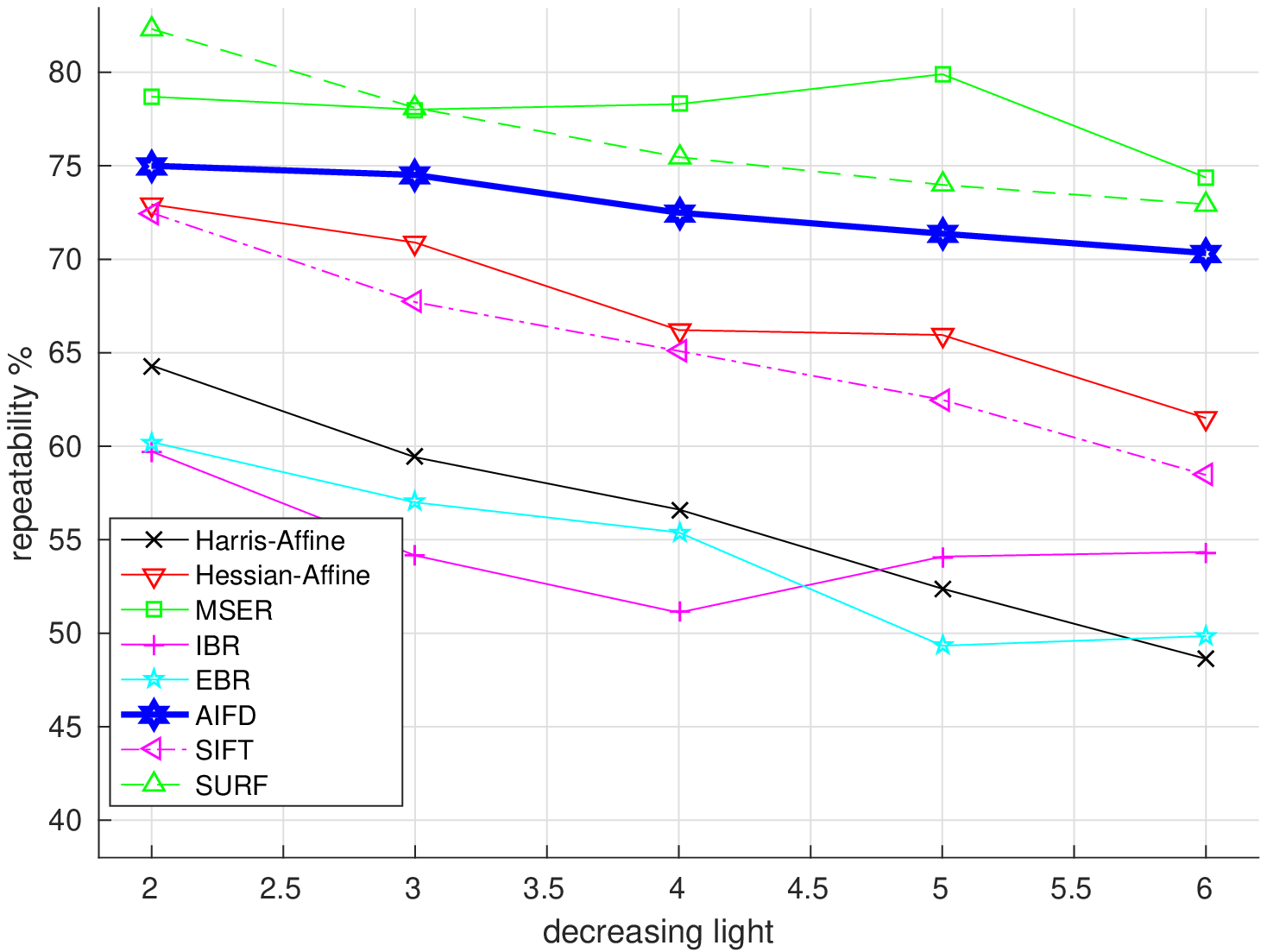}
\small\centerline{(a)}\medskip
\end{minipage}
\begin{minipage}{0.49\linewidth}
\centering
\includegraphics[width=0.89\linewidth]{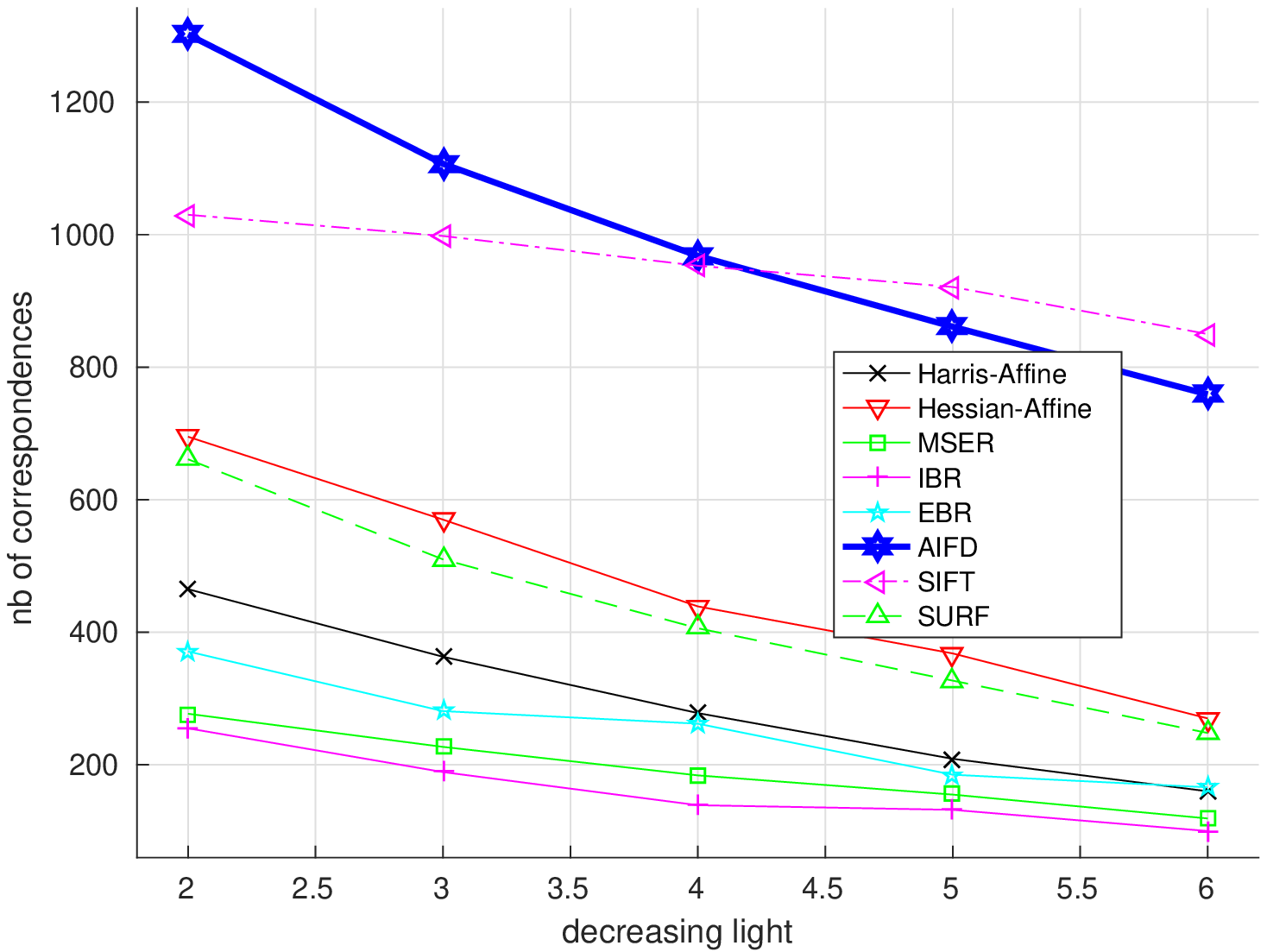}
\small\centerline{(b)}\medskip
\end{minipage}
\begin{minipage}{0.49\linewidth}
\centering
\includegraphics[width=0.89\linewidth]{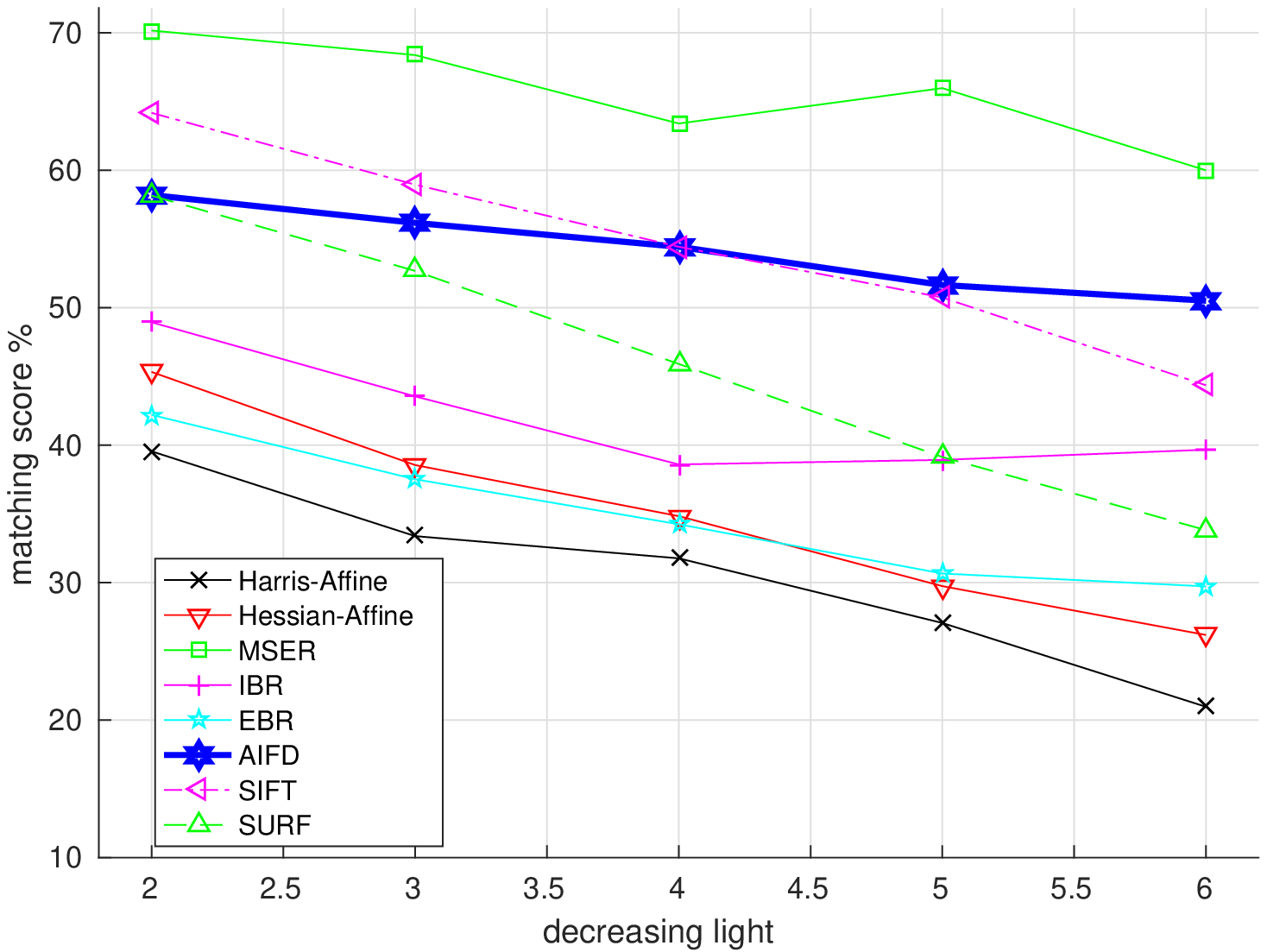}
\small\centerline{(c)}\medskip
\end{minipage}
\begin{minipage}{0.49\linewidth}\centering
\includegraphics[width=0.89\linewidth]{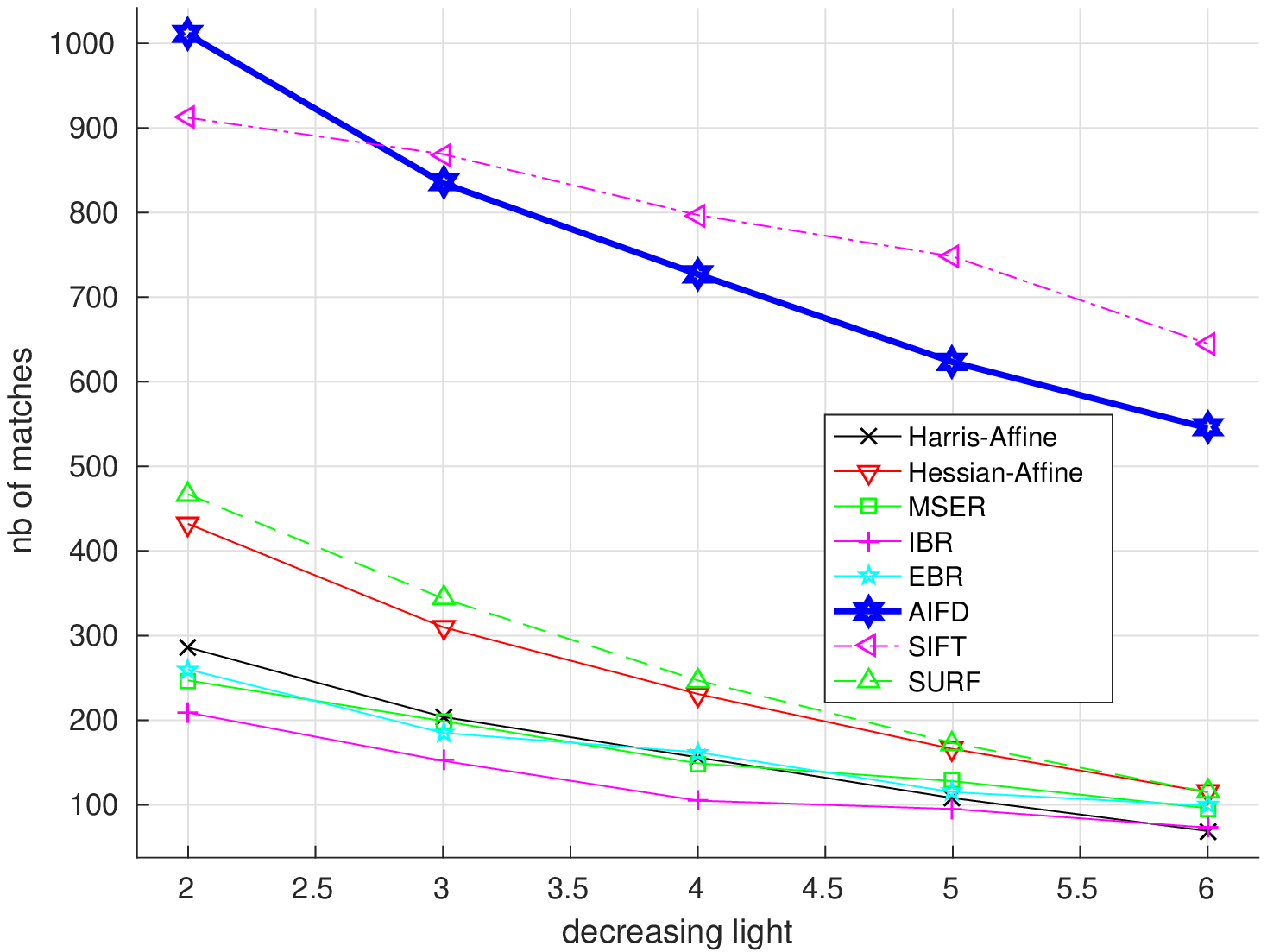}
\small\centerline{(d)}\medskip
\end{minipage}
\caption{The detector and descriptor performance on illumination changed images (Leuven sequence). (a) \emph{repeatability score}. (b)\emph{number of correspondences}. (c)\emph{matching scores}. (d) \emph{number of matches}. Generally speaking, our proposed AIFD has a comparable performance on the illumination changes. It gains a best and second best performance on the number of correspondence and number of matches. }
\label{fig:light_test}
\end{figure*}

\begin{figure*}[!t]
\begin{minipage}{0.49\linewidth}
\centering
\includegraphics[width=0.89\linewidth]{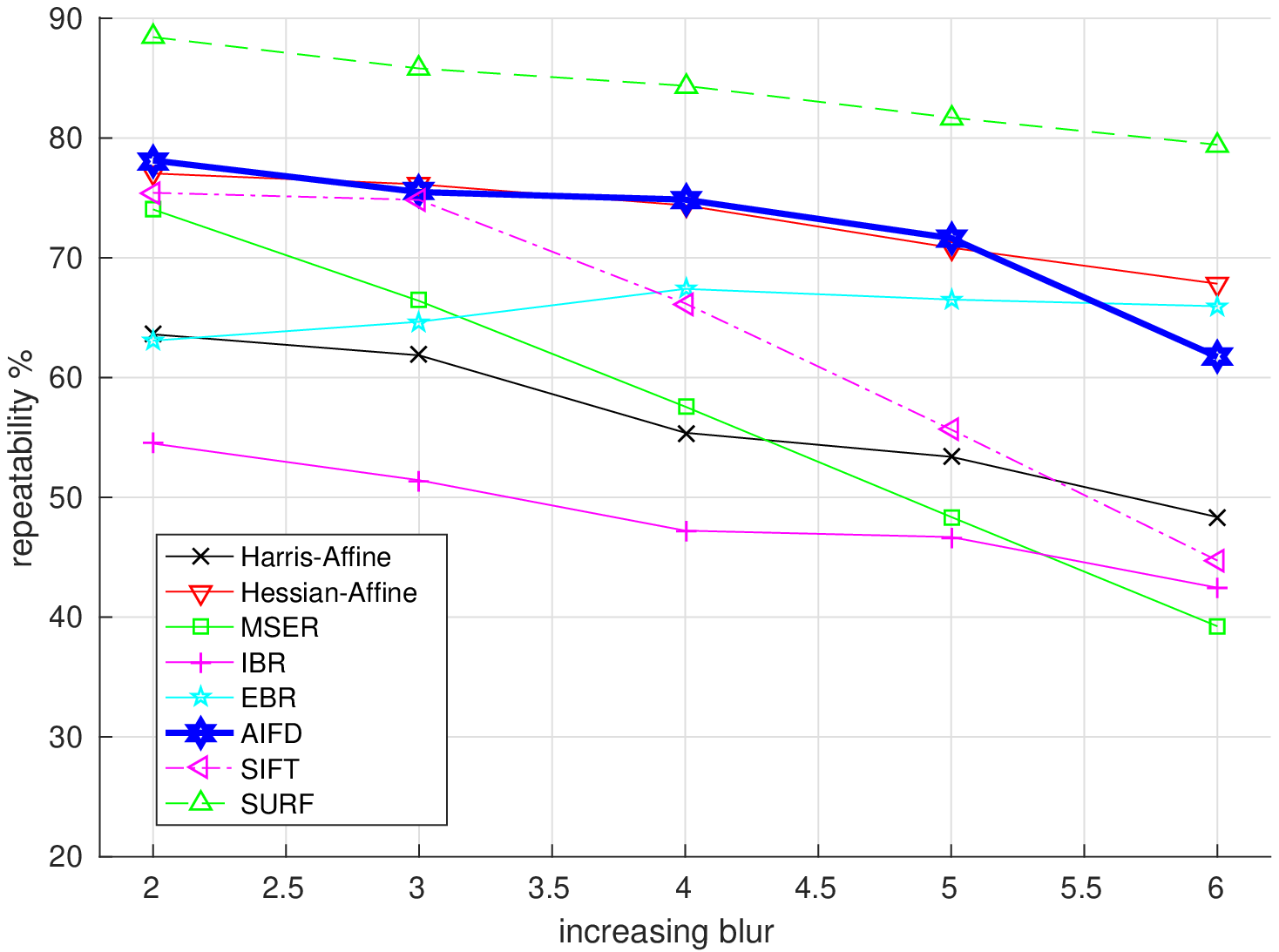}
\small\centerline{(a)}\medskip
\end{minipage}
\begin{minipage}{0.49\linewidth}
\centering
\includegraphics[width=0.89\linewidth]{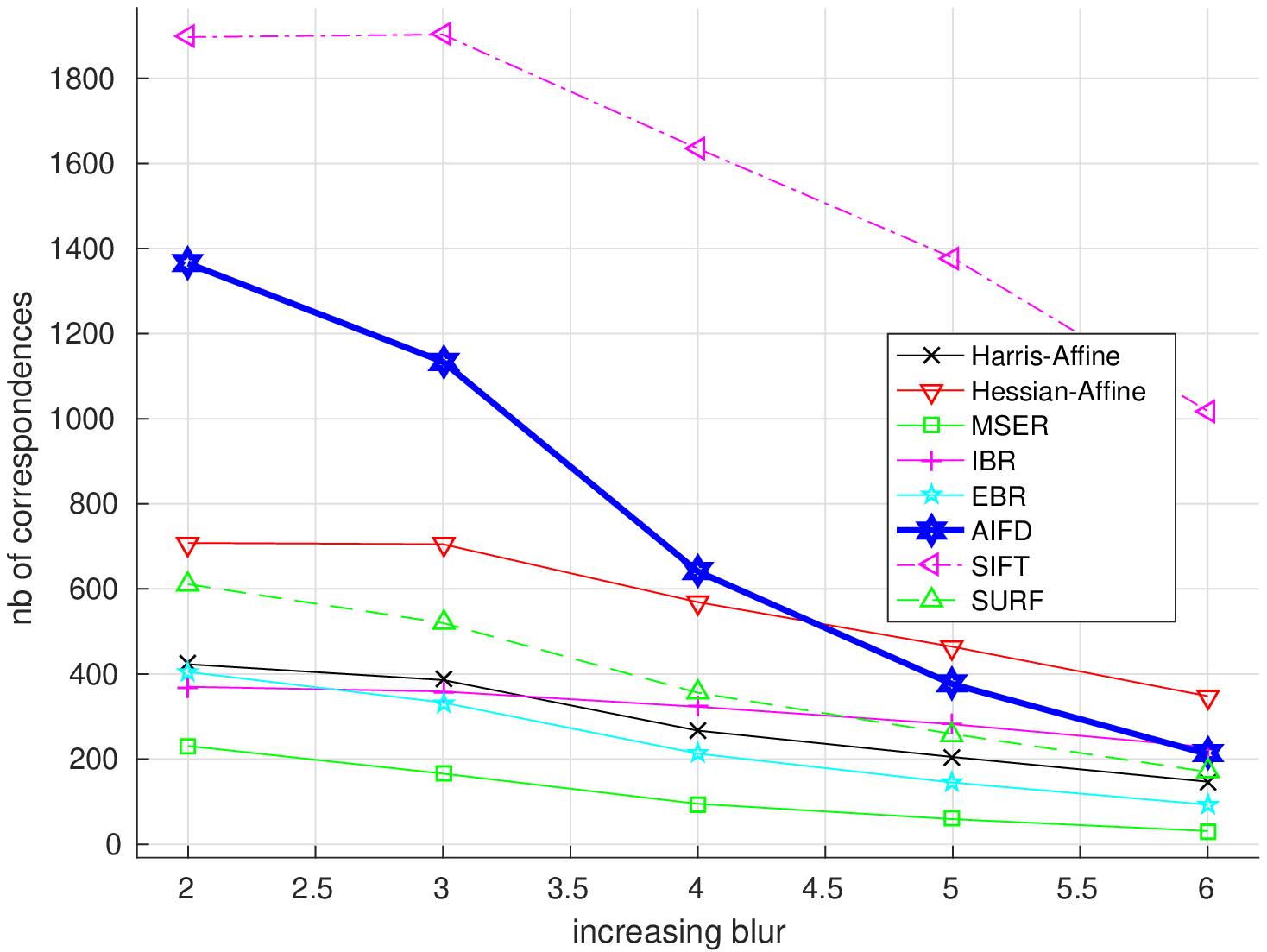}
\small\centerline{(b)}\medskip
\end{minipage}
\end{figure*}

\begin{figure*}[!t] \ContinuedFloat
\begin{minipage}{0.49\linewidth}
\centering
\includegraphics[width=0.89\linewidth]{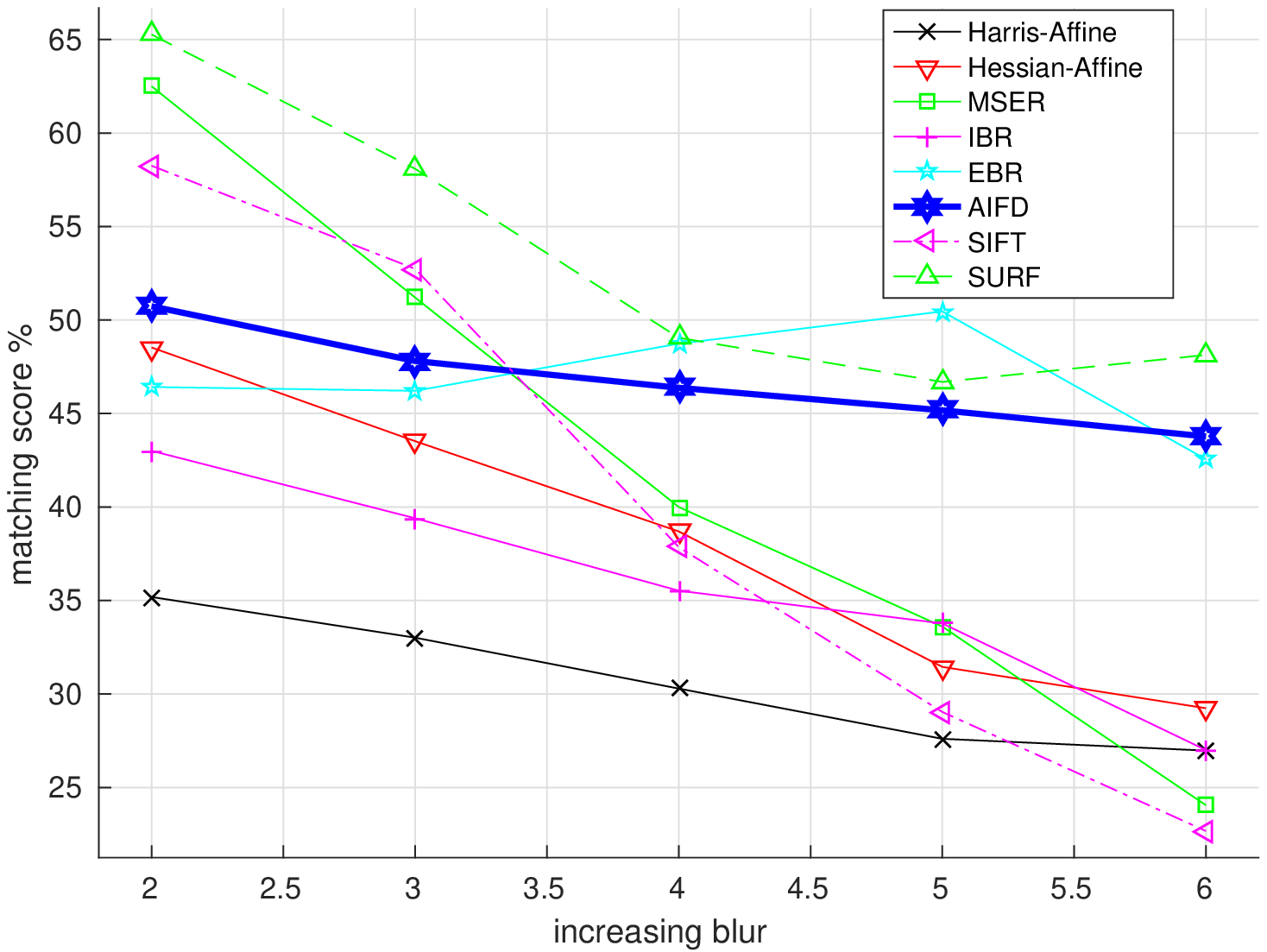}
\small\centerline{(c)}\medskip
\end{minipage}
\begin{minipage}{0.49\linewidth}\centering
\includegraphics[width=0.89\linewidth]{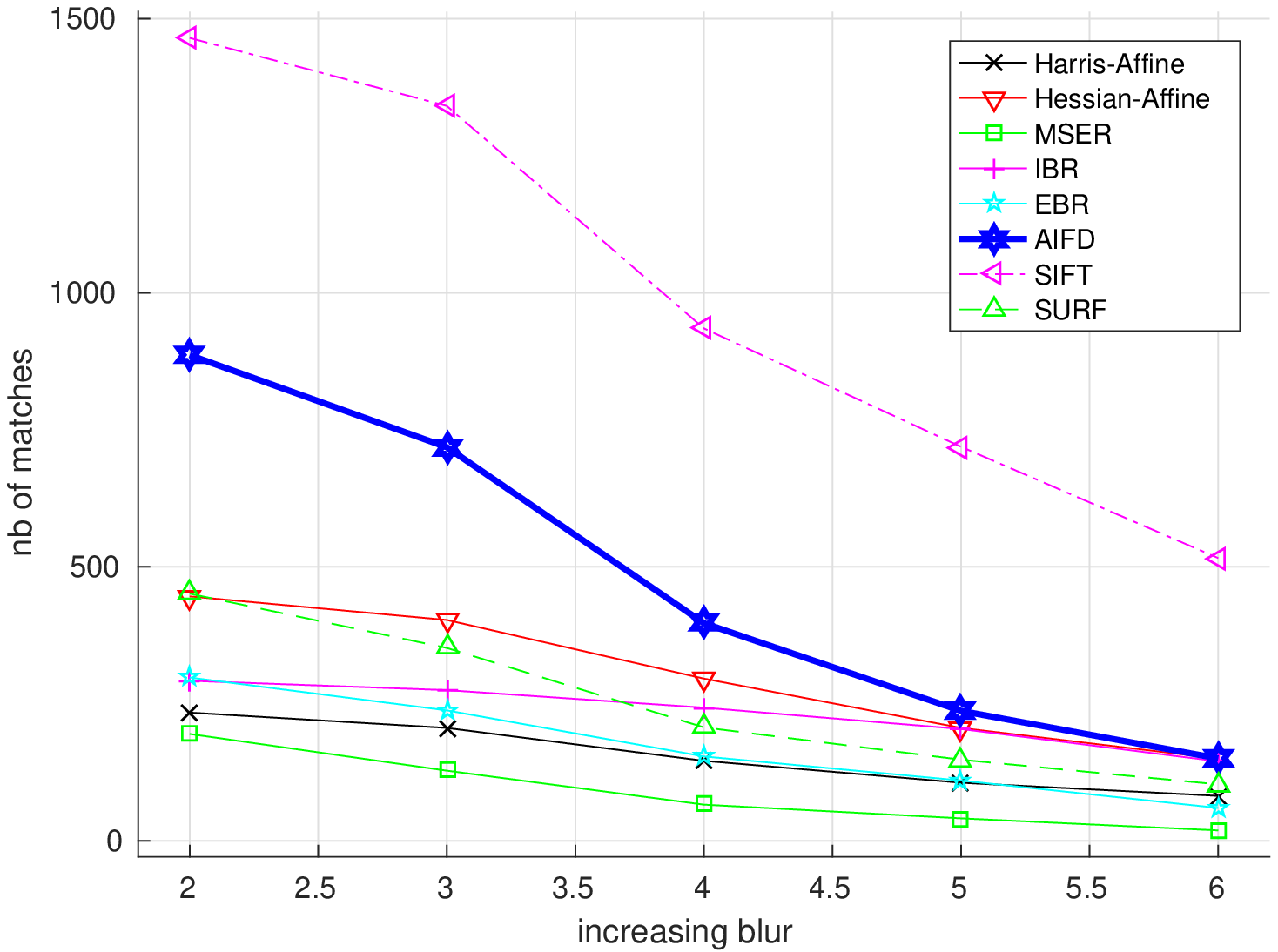}
\small\centerline{(d)}\medskip
\end{minipage}
\caption{The detector and descriptor performance on blur changed images (bikes sequence). (a) \emph{repeatability score}. (b) \emph{number of correspondences}. (c) \emph{matching scores}. (d) \emph{number of matches}. Generally speaking, our proposed AIFD has a comparable performance on the blur changes. }
\label{fig:blur_test}
\end{figure*}

\begin{figure*}[!h]
\begin{minipage}{0.49\linewidth}
\centering
\includegraphics[width=0.89\linewidth]{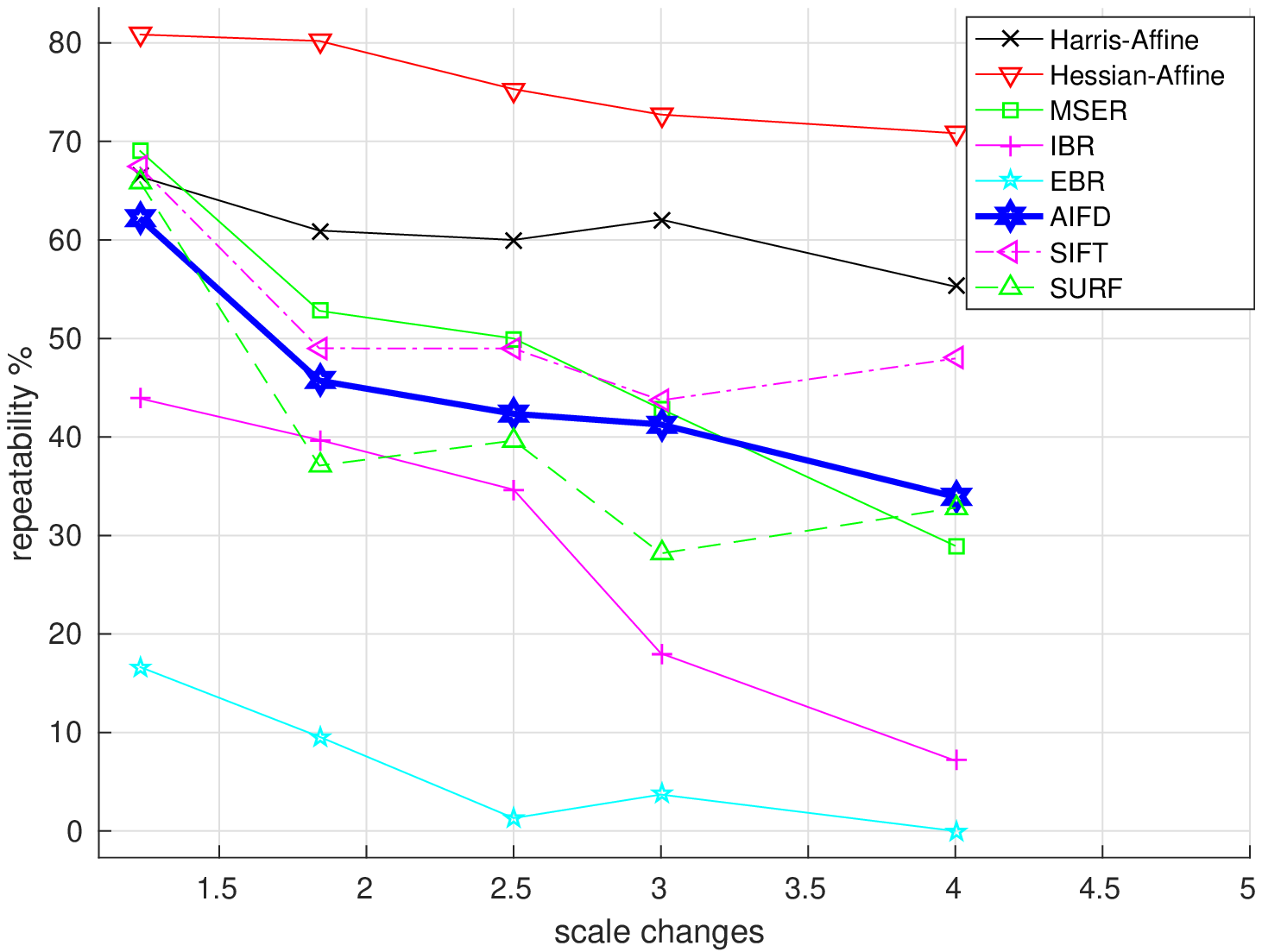}
\small\centerline{(a)}\medskip
\end{minipage}
\begin{minipage}{0.49\linewidth}
\centering
\includegraphics[width=0.89\linewidth]{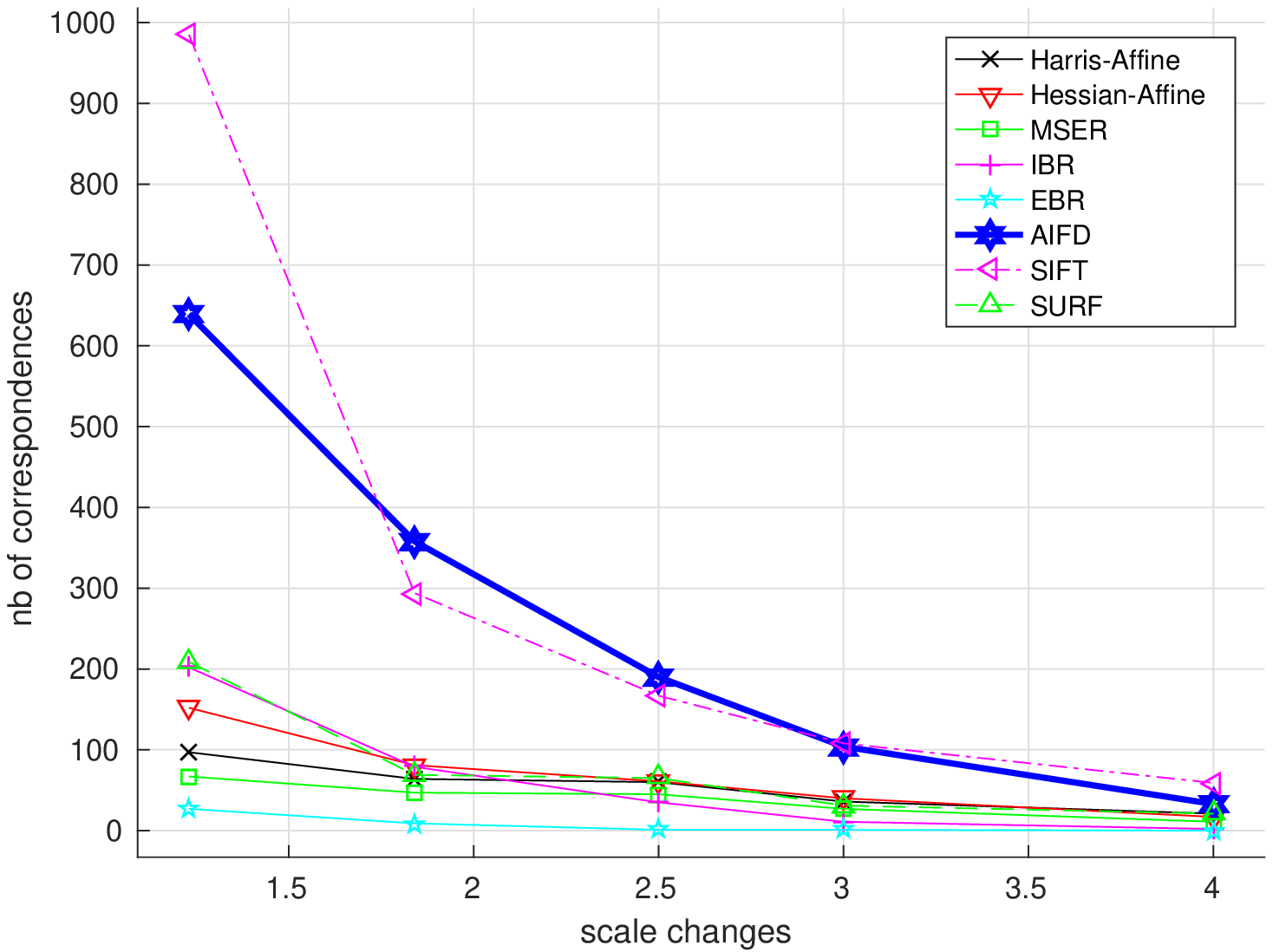}
\small\centerline{(b)}\medskip
\end{minipage}
\begin{minipage}{0.49\linewidth}
\centering
\includegraphics[width=0.89\linewidth]{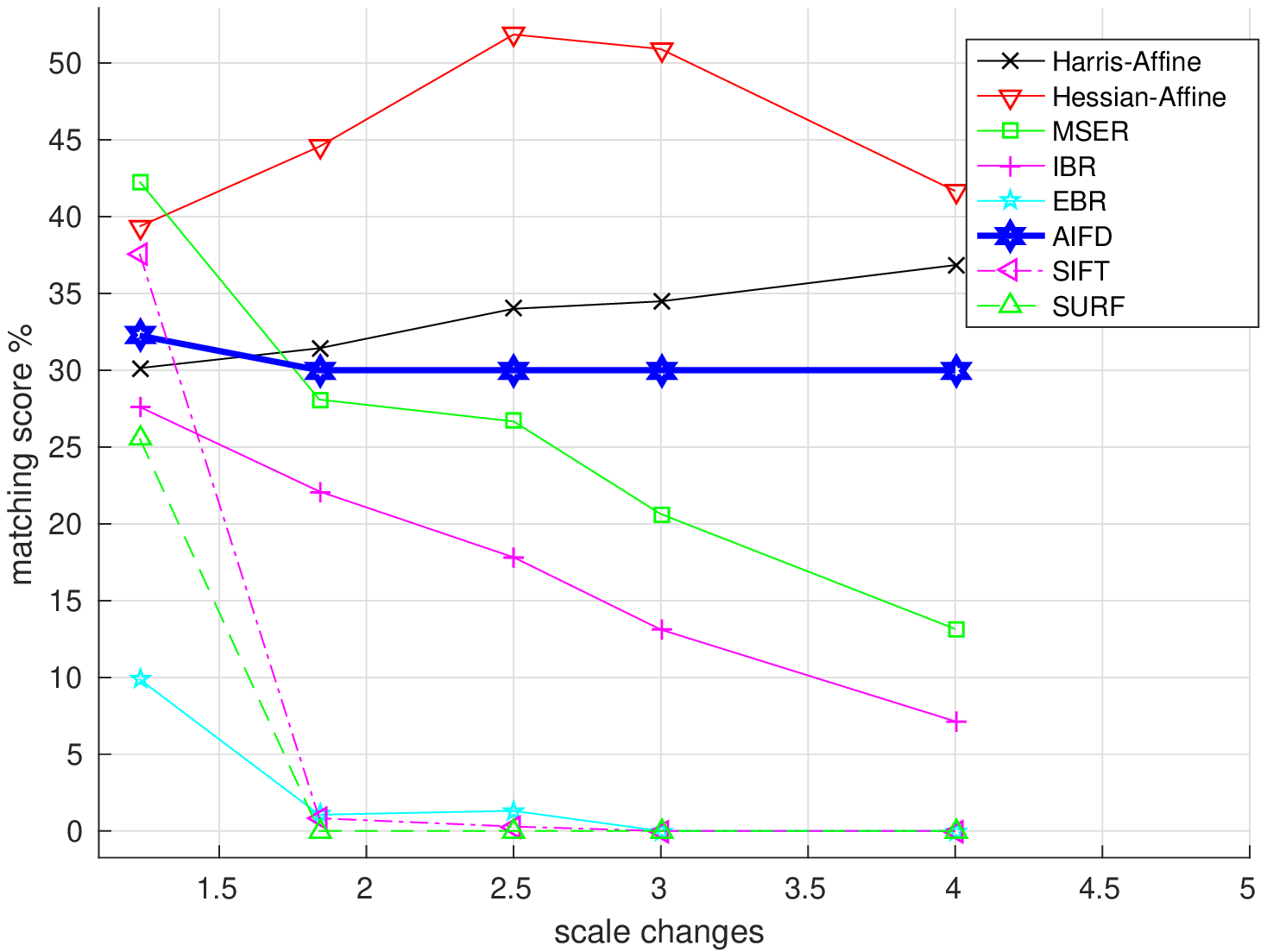}
\small\centerline{(c)}\medskip
\end{minipage}
\begin{minipage}{0.49\linewidth}\centering
\includegraphics[width=0.89\linewidth]{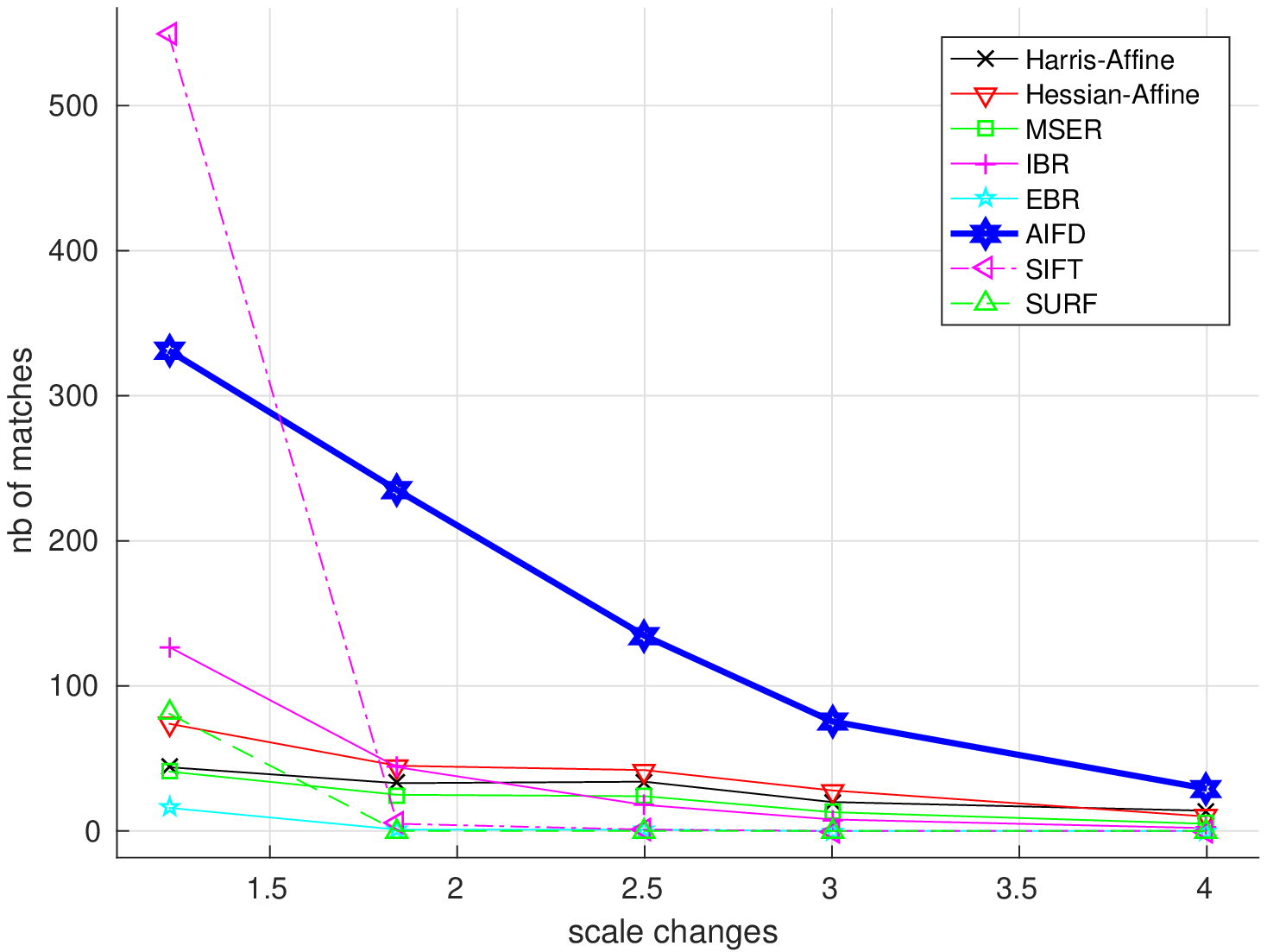}
\small\centerline{(d)}\medskip
\end{minipage}
\caption{The detector and descriptor performance on scale and rotation changed images (bark sequence). (a) \emph{repeatability score}. (b) \emph{number of correspondences}. (c) \emph{matching scores}. (d) \emph{number of matches}. Generally speaking, our proposed AIFD has a comparable performance on the scale and rotation changes. }
\label{fig:scal_test}
\end{figure*}

\begin{figure*}[!h]
\begin{minipage}{0.49\linewidth}
\centering
\includegraphics[width=0.89\linewidth]{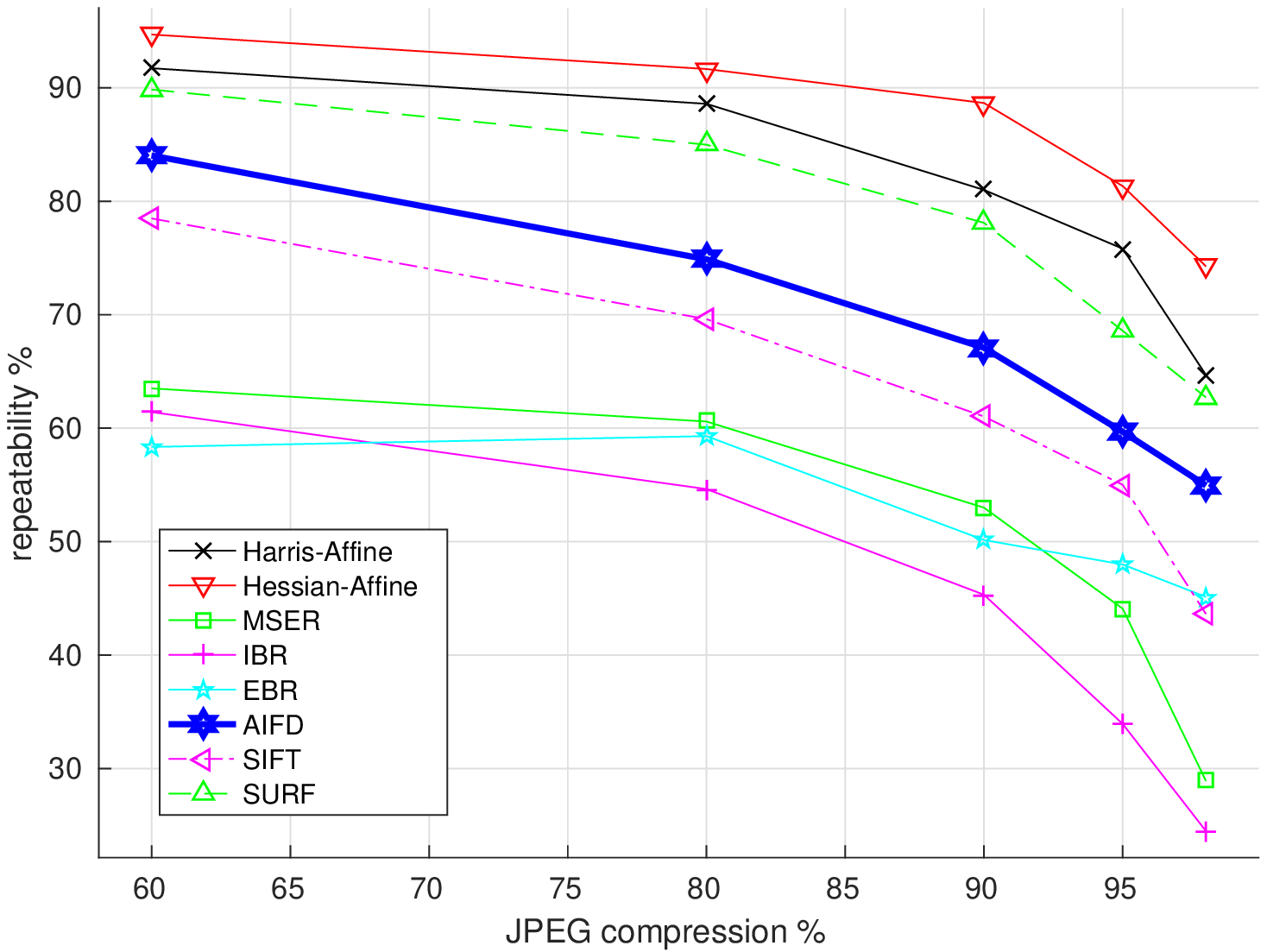}
\small\centerline{(a)}\medskip
\end{minipage}
\begin{minipage}{0.49\linewidth}
\centering
\includegraphics[width=0.89\linewidth]{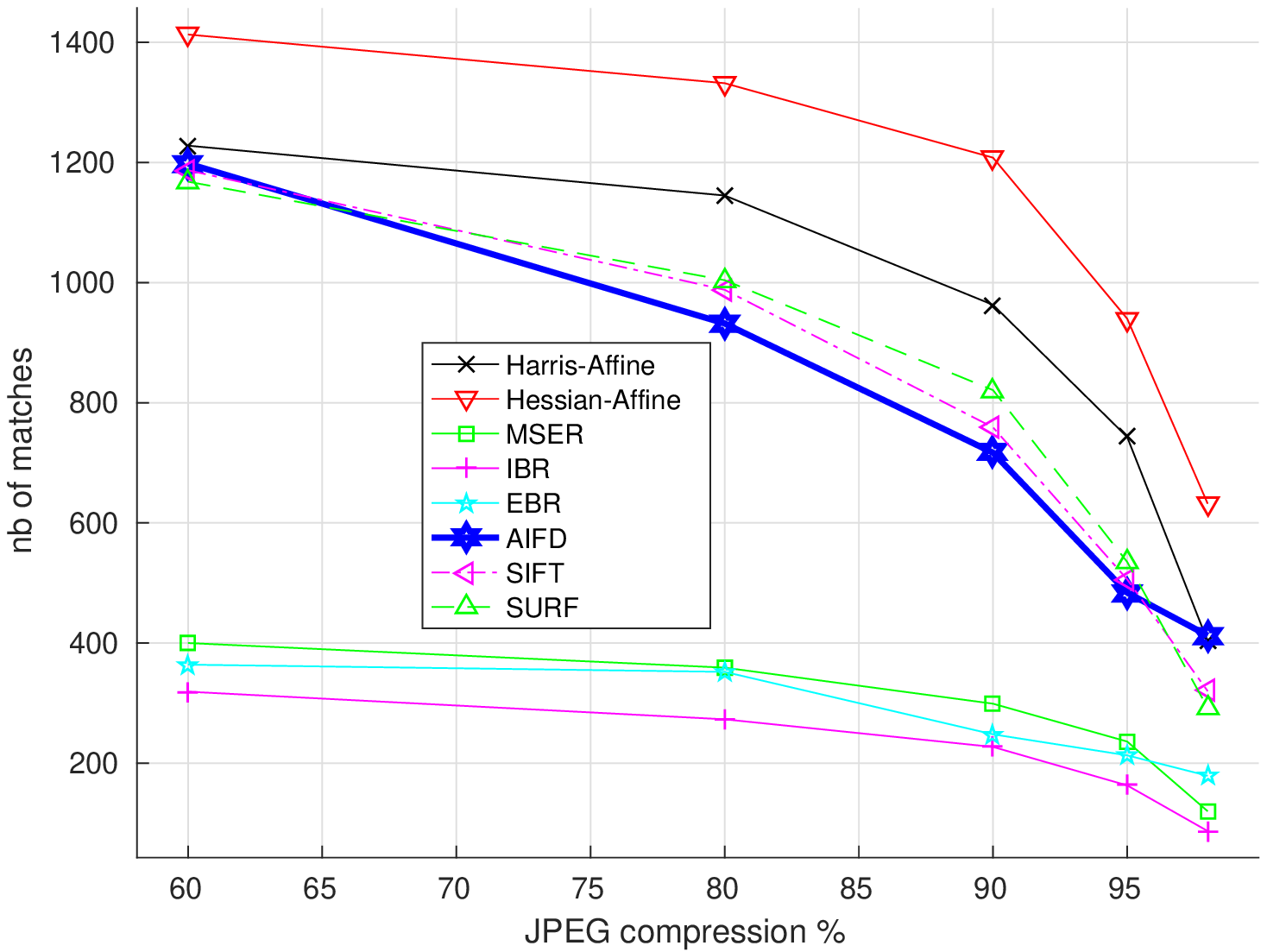}
\small\centerline{(b)}\medskip
\end{minipage}
\begin{minipage}{0.49\linewidth}
\centering
\includegraphics[width=0.89\linewidth]{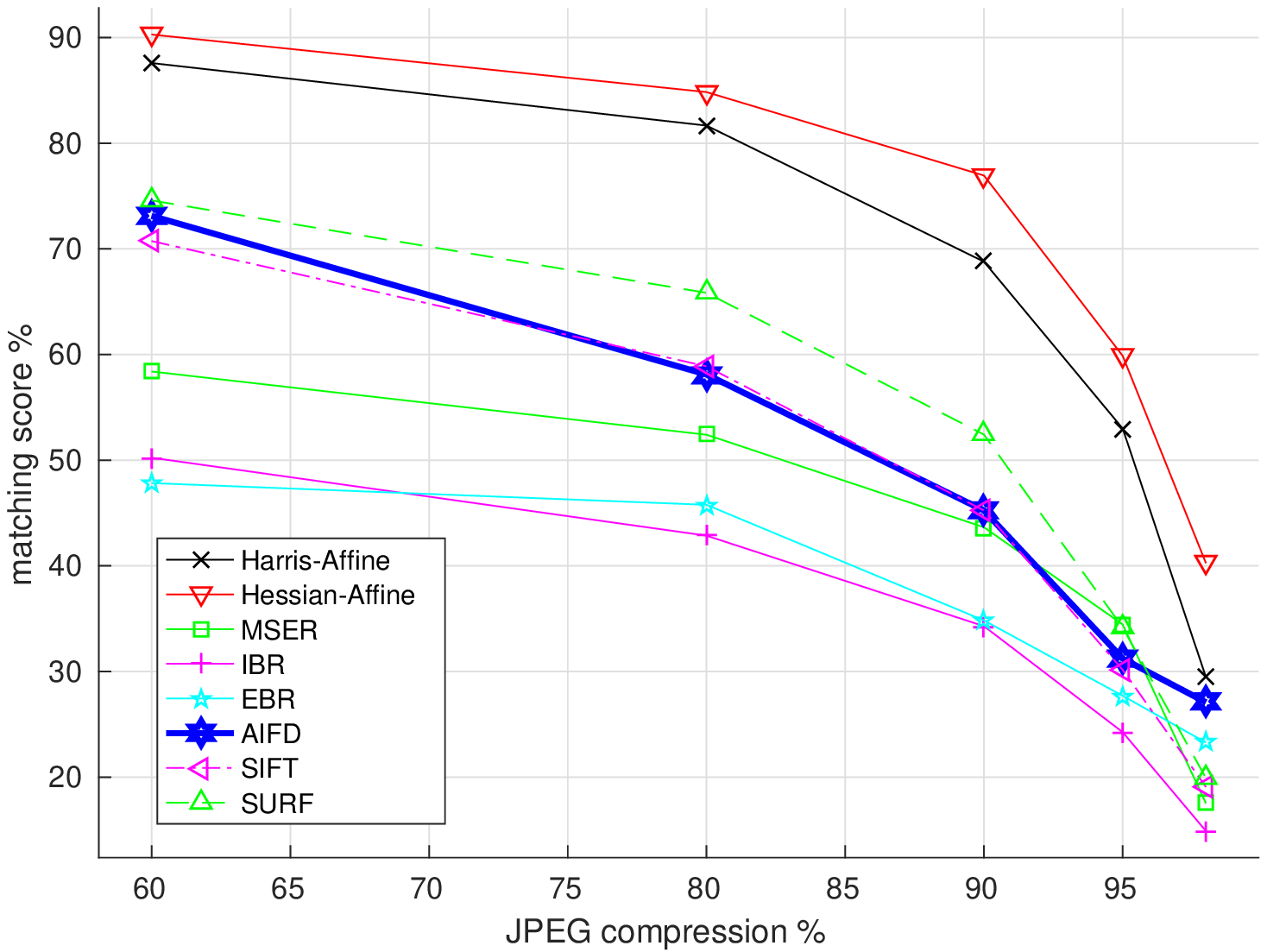}
\small\centerline{(c)}\medskip
\end{minipage}
\begin{minipage}{0.49\linewidth}\centering
\includegraphics[width=0.89\linewidth]{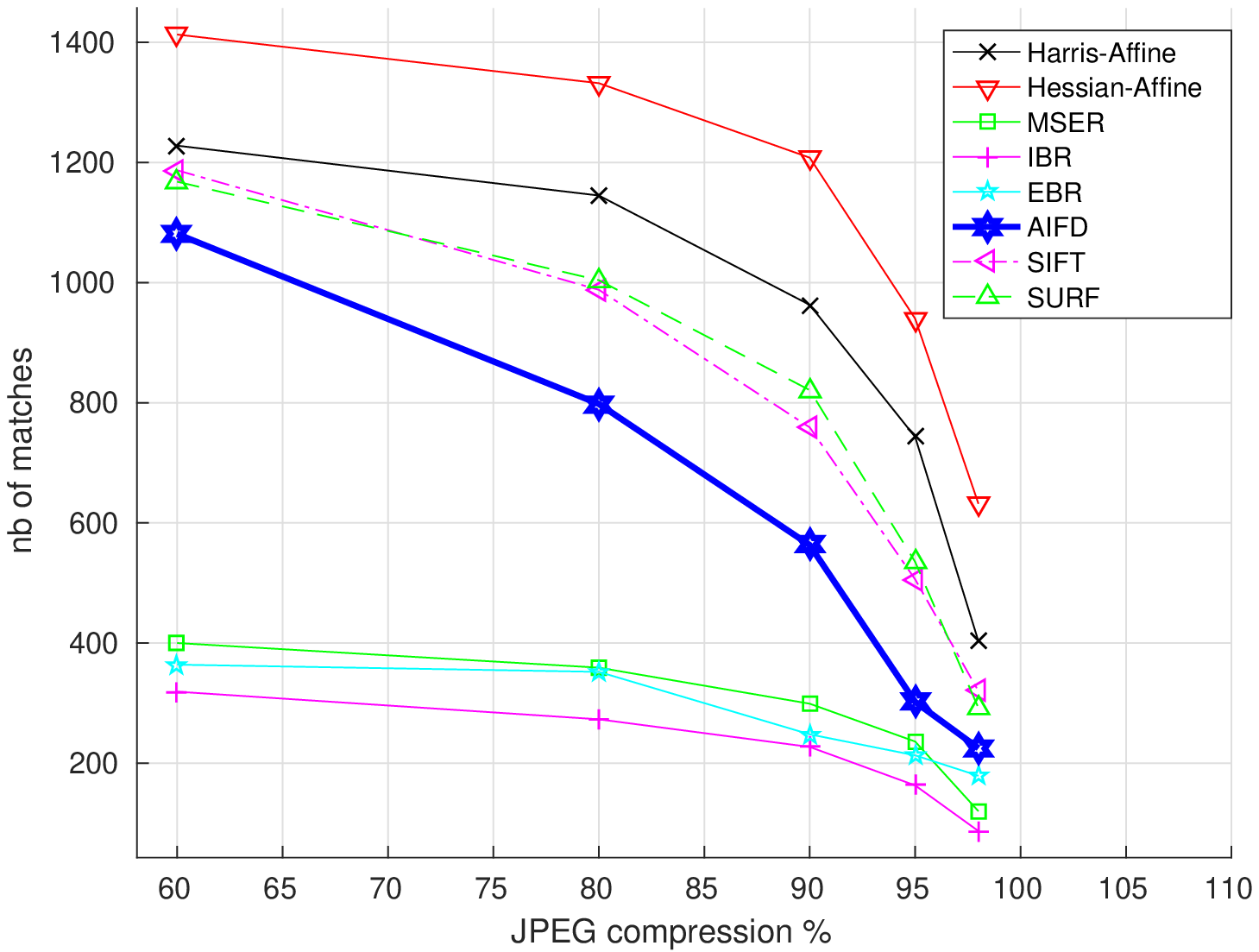}
\small\centerline{(d)}\medskip
\end{minipage}
\caption{The detector and descriptor performance on JPEG compressed images (UBC sequence). (a) \emph{repeatability score}. (b) \emph{number of correspondences}. (c) \emph{matching scores}. (d) \emph{number of matches}. Generally speaking, our proposed AIFD has a comparable performance on the JPEG compressed images. }
\label{fig:comp_test}
\end{figure*}

We take 4 tests upon each set of images to evaluate the algorithms robustness respectively on change of view points, scales, illumination conditions and compressions. These 4 tests include the evaluation criterion of \emph{repeatability score}, \emph{number of correspondences}, \emph{matching scores} and \emph{number of matches}. The \emph{repeatability score} for a given pair of images is computed as the ratio between the number of correspondences and the total numbers of extracted features. The total numbers of extracted features will only be accounted if appearing on both images. The \emph{number of correspondences} and the \emph{number of matches}, as their name implied, account the total number of correspondences and the total numbers of the matched features. Similar to the \emph{repeatability score}, the \emph{matching scores} refers to the ratio between matched features to total extracted features. Among these 4 criterion, \emph{repeatability score} and \emph{number of correspondences} can be used to evaluate the performance of detector; \emph{matching scores} and \emph{number of matches} can be used to evaluate the performance of descriptor.

More image sequences concerning about the change of blur, rotation and scales have also be provided by Mikolajczyk (bark, trees, etc.) and the results of these image sequences are similar to the presented results. On the change of scales, illumination conditions and compressions, our proposed AIFD has a performance comparable to the state-of-the-art algorithms, including SIFT \cite{lowe2004distinctive} \cite{lowe1999object}, SURF \cite{bay2006surf} \cite{bay2008speeded}, MSER \cite{matas2004robust}, Harris-Affine \cite{mikolajczyk2002affine}, Hessian-Affine \cite{mikolajczyk2004scale}, IBR \cite{tuytelaars2000wide}, EBR \cite{mikolajczyk2003shape}. 

The main purpose of this paper is to propose an image featuredescriptor more capable of detecting and matching the images at extreme view points. Except the image sequences provided by Mikolajczyk, we take more tests on the change of view points by introducing \emph{dataset$\_$MorelYu09}, a specialized dataset to evaluate the performance of feature matching algorithm on affine transformations. 

As it depicted in Figure~\ref{fig:morelyu09}, \emph{dataset\_ MorelYu09} contains more images of extreme view points and can be better used to assess the performance of image detector and descriptor on the affine robustness \cite{yu2011asift}.

Generally speaking, our proposed feature matching algorithm AIFD, as proven by the experiments, has a great advantage on the affine robustness compared to some other most frequently used pairwise matching algorithms. Meanwhile, its performance to some other geometric and photometric transformations, like the change of zoom, rotation, blur, illumination and compression, is also comparable to other state-of-the-art feature matching algorithms. It can be noticed that Hessian-Affine, specially to JPEG compression, has a quite good performance. MSER is the second best feature matching algorithm to the affine robustness.

\section{Conclusion}

In this paper, we have proposed an affine invariant feature descriptor, as an extension to our previous proposed affine invariant feature detector. Based on the theory of affine scale space, we have introduced the affine gradient filters and its polynomial expression to create the local affine gradient patch of the specified scale. A histogram regarding the local gradient of the detected features can then be generated with some modification to the affine transformations. With the affine gradient accounting for the local affine transformation, this gradient histogram can then be used to identify the feature\rq{}s local geometry and scale informations, and can be further applied as the feature\rq{}s descriptor, similar to the SIFT descriptor. Thanks to this affine invariant feature descriptor, the features detected by the affine invariant feature detector can then be labelled and matched in an affine transformed environment.
  
Affine scale space is a forward model providing a more general approach to the scale and affine invariant image representation, allowing to predict what will happen after the affine transformation. Based on this model, we have proposed affine invariant feature detector and descriptor, a pairwise matching algorithm more capable to detect and match the images under an affine transformed environment. Proved by the experiments, it has a better performance than other pairwise matching algorithms especially under an extreme different view points. To the images after affine transformations, it can detain a higher repeatability, and matching score; more pair of correspondences and matched points.

\section{Acknowledgements}
We would like to express our gratitude to the projects EU FP7 LIVECODE(295151), EU FP7 HAZCEPT(318907), EUHorizon 2020 STEP2DYNA(691154), EUHorizon 2020 ENRICHME(643691) for financial support.






\bibliographystyle{unsrt} 
\bibliography{topref}
\end{document}